\documentclass[journal,10pt,twocolumn,twoside]{IEEEtran}
\usepackage[table]{xcolor}

\usepackage{ifpdf}
\usepackage{cite}
\usepackage{amsmath,amssymb,amsfonts,bm}
\usepackage{algorithmic}
\usepackage{graphicx, adjustbox}
\usepackage{array}
\usepackage{enumitem}
\usepackage{breqn}
\usepackage{bm}
\usepackage{ dsfont }
\usepackage{ upgreek }
\usepackage{textcomp}
\usepackage{multirow}
\usepackage{algorithm}
\usepackage{algorithmic}
\usepackage{caption}
\usepackage{pgfplots}
\usepackage{tkz-euclide}
\usepackage{subcaption}
\usepackage{tikz}{\tiny }
\usepackage{hyperref}
\usepackage{balance}
\usepackage{ tipa }

\usepackage{ upgreek }
\usepackage{soul}
\usepackage{breqn}
\usepackage{makecell}
\usepackage{tabularray}
\usepackage[english]{babel}

\usepackage{array}
\usepackage{multirow}
\usepackage{caption}

\addto\captionsenglish{}
\allowdisplaybreaks

\usepackage{amsthm}

\usepackage{fancyhdr}

\usepackage{ upgreek }
\usepackage{soul}
\makeatletter
\newcommand{\vast}{\bBigg@{4}}

\newcommand{\Vast}{\bBigg@{5}}
\makeatother
\pgfplotsset{compat=1.18}

\usepackage{amssymb}

\usepackage{tikz}

\begin{document}

\title{\huge Federated Learning for Cyber Physical Systems: \\ A Comprehensive Survey}
	
	\author{Minh K. Quan, Pubudu N. Pathirana, Mayuri Wijayasundara, Sujeeva Setunge, Dinh C. Nguyen, \\
 Christopher G. Brinton, David J. Love,~\IEEEmembership{Fellow,~IEEE,} 
and H. Vincent Poor,~\IEEEmembership{Life Fellow,~IEEE}
	\\
  \thanks{Minh K. Quan, Pubudu N. Pathirana, and Mayuri Wijayasundara are with the School of Engineering, Deakin University, Waurn Ponds, VIC 3216, Australia (e-mails: \{m.quan, pubudu.pathirana, mayuri\}@deakin.edu.au).}
   \thanks{Dinh C. Nguyen is with the Department of Electrical and Computer Engineering, University of Alabama in Huntsville, USA (e-mail: dinh.nguyen@uah.edu).}
   \thanks{Sujeeva Setunge is with The School of Engineering, Royal Melbourne Institute of Technology University, Melbourne, VIC 3000, Australia (e-mail: sujeeva.setunge@rmit.edu.au).}
 \thanks{Christopher G. Brinton, and David J. Love are with the Elmore Family School of Electrical and Computer Engineering, Purdue University, USA (e-mails: \{cgb, djlove\}@purdue.edu).}
  \thanks{H. Vincent Poor is with the Department of Electrical and Computer
Engineering, Princeton University, Princeton, NJ 08544 USA (e-mail: poor@princeton.edu).}
}
	\markboth{IEEE Communications Surveys \& Tutorials}%
	{}
	
	\maketitle
	
	\begin{abstract}
 	The integration of machine learning (ML) in cyber physical systems (CPS) is a complex task due to the challenges that arise in terms of real-time decision making, safety, reliability, device heterogeneity, and data privacy. There are also open research questions that must be addressed in order to fully realize the potential of ML in CPS. Federated learning (FL), a distributed approach to ML, has become increasingly popular in recent years. It allows models to be trained using data from decentralized sources. This approach has been gaining popularity in the CPS field, as it integrates computer, communication, and physical processes. Therefore, the purpose of this work is to provide a comprehensive analysis of the most recent developments of FL-CPS, including the numerous application areas, system topologies, and algorithms developed in recent years. The paper starts by discussing recent advances in both FL and CPS, followed by their integration. Then, the paper compares the application of FL in CPS with its applications in the internet of things (IoT) in further depth to show their connections and distinctions. Furthermore, the article scrutinizes how FL is utilized in critical CPS applications, e.g., intelligent transportation systems, cybersecurity services, smart cities, and smart healthcare solutions. The study also includes critical insights and lessons learned from various FL-CPS implementations. The paper's concluding section delves into significant concerns and suggests avenues for further research in this fast-paced and dynamic era.
	\end{abstract}
	
	
	\IEEEpeerreviewmaketitle

\section{Introduction}
\label{introduce}
Cyber physical systems (CPS) are gaining prominence across several industries, resulting in a fast growth in their prevalence \cite{ikuabe2020cyber}. CPS are characterized by the integration of computational and physical components to perform a certain function or task \cite{bagheri2015cyber}. Several industries, e.g., manufacturing, transportation, healthcare, agriculture, and energy, have the potential to be transformed by CPS technology \cite{song2017security}. Integrating artificial intelligence (AI) into these systems is one of the most significant factors that affect CPS productivity. AI algorithms may be used to assess data received by CPS sensors and make decisions or take actions based on this evaluation \cite{lv2021artificial}, allowing CPS to adapt in real time to changing environmental conditions. This can improve the accuracy, effectiveness, and safety of CPS in several applications. As the number of CPSs continues to rise and their capabilities continue to advance, it is projected that AI will become increasingly integrated into CPS \cite{sakhnini2020ai}. Therefore, CPS must be operated with proper privacy and security safeguards to reduce unintended consequences on applications \cite{keshk2018privacy}. According to \cite{darwish2018cyber}, it is expected that this trend will encourage further research and development in the CPS field, as well as the adoption of comparable technologies, e.g., IoT, sensor networks, cloud computing, and multi-agent systems in other industries.

Recent studies have suggested utilizing FL \cite{nguyen2021federated1} to construct an intelligent, private CPS. FL is a distributed ML approach for classification models that works by coordinating several sensors with a central server without providing entire data sets \cite{konevcny2016federated}. Intelligent CPS networks employ a system where several CPS devices function as ``workers" that collaborate with an ``aggregator" (e.g., a server) to train the overarching NN model. Specifically, the main responsibility of the aggregation server is to generate a universal model using predetermined learning parameters. Afterwards, every ``worker" acquires the most recent model from the aggregator, improves upon the model using its local data set (e.g., by running some form of stochastic gradient descent), and finally sends this updated model to the aggregator. The aggregation server is capable of generating an up-to-date global framework by gathering all regional alterations. Thanks to the distributed computing power of worker devices, training quality may be boosted while user privacy can be enhanced. When the global model converges, local devices will download the most recent version from aggregators and begin computing their own local adjustments. 

\subsection{Background and Motivation}
FL can deliver the following substantial improvements for CPS applications as a result of its innovative operating philosophy:
\begin{itemize}
  \item Improved Privacy: FL enables devices to engage in model training without disclosing their information to a central server or other devices. This is especially critical in CPS applications where sensitive personal or proprietary data may be included. As data privacy laws, e.g., GDPR (General Data Protection Regulation) \cite{voigt2017eu}, become increasingly stringent, FL's potential to enhance privacy protection makes it an appealing alternative for constructing secure and intelligent CPS systems. This is due to the fact that FL enables data to be analyzed locally on personal devices, reducing the risk of sensitive information being transmitted to a central server.

  \item Data Ownership: In FL, each device retains ownership and control over its own data, as it is not required to exchange raw data with a central server or other devices. Each device trains its own model using its own data, starting with a global model as the initial point. While the training process is localized, the global model serves as a foundation for further refinement on each device. After updating the model, the device transmits the new model parameters to the central server or to other devices. In the case of CPS applications, where devices may be owned by multiple parties and subject to different legal or regulatory constraints, this might be of particular importance \cite{song2019profit}.
  
  \item Reduced Communication and Computation Requirements: FL decreases communication and computation compared to centralized learning techniques \cite{yang2020energy}. Traditional centralized learning transmits all training data to a single server, which is resource-intensive for data transmission and processing. FL enables each device to train models locally, which can potentially reduce data transit compared to centralized training where all data is sent to a central server. However, the iterative nature of FL, with multiple rounds of model exchange, can lead to a higher overall data transmission in some cases. Similarly, while local processing on each device may be less intensive, the aggregate processing across all devices and the server could potentially be more than in centralized training. This can be especially beneficial in CPS applications where devices may have limited communication or processing capacity.
  
  \item Improved Model Performance: FL allows each device to train on a broader, more diversified dataset, with the only expense being that each deice must contribute a processed version of its own data. This is important for CPS applications with heterogeneous and dispersed data. Each sensor or camera in a smart city application may capture data from a unique location and/or environment, resulting in a more comprehensive and representative training set. This can improve model performance \cite{nguyen2021federated3} compared to training on a single device or centralized dataset, because the model can absorb a wider variety of information and experiences. 

  \item Improved scalability, robustness, and adaptability: With FL, a greater number of devices may participate in the model-training process in a decentralized way, making it simpler to scale the training process. In contrast to training the model on a single dataset, training it on a varied array of data can increase its resilience. This allows the model to adjust to variations in the distribution of data over time in CPS applications where data may change frequently \cite{li2021ditto}.
\end{itemize}

FL has been proposed as a solution for several CPS applications owing to its inherent advantages in domains e.g., unmanned aerial vehicles (UAVs), smart transportation, and smart healthcare. By working together with roadside units for collaborative data learning, FL has successfully provided intelligent vehicular services \cite{zhang2020fenghuolun} e.g., road safety prediction and autonomous driving. This has resulted in improved accuracy during training and increased privacy for vehicle recognition. Intelligent healthcare services powered by ML models can be offered through FL, without the need for patient data to be shared among several medical institutions \cite{liu2022blockchain}. FL allows healthcare organizations, e.g., hospitals, to develop ML models using their own data and transmit only the learned parameters to an aggregator for further processing. This improved level of inter-hospital communication, which also prioritizes user privacy, provides several advantages, including expedited clinical diagnosis and treatment. For CPS applications, FL is a crucial resource due to its ability to protect sensitive information while also allowing for data sovereignty and decentralization. In light of the major achievements with FL-CPS applications, it is appropriate to highlight this important field of study.

\subsection{Our Contributions}
Despite the fact that FL has been studied extensively, to our knowledge, no recent study has examined into FL's possible use in CPS networks. The viability of utilizing FL in CPS services, including data offloading, CPS connectivity, and localization, has not been examined in any of the surveys from \cite{zhang2021survey} to \cite{abdulrahman2020survey}. Although there are several research works on the use of FL for IoT, IIoT \cite{nguyen2021federated}\cite{imteaj2021survey}, and UAV-enabled systems \cite{brik2020federated}, the potential of FL for CPS remains largely unexplored. Moreover, the whole range of FL-CPS applications, from smart city to smart transportation, has yet to be discussed in any detail. Given these constraints, we feel a full investigation of FL's inclusion into CPS systems is necessary. We provide a cutting-edge examination of FL's use in a variety of critical CPS applications such as data-sharing between sensors, caching, attack detection, localization, and CPS privacy. This paper's key contribution is to explore and evaluate the applicability of FL in several CPS domains, e.g., smart transportation, smart cities, smart healthcare, and smart attack detection. Moreover, the most remarkable findings from the survey are presented. Furthermore, this work identifies significant research gaps in the FL-CPS field, and Table \ref{table:related_works} highlights the innovative findings of this work compared to previous publications. In conclusion, we discuss a range of interesting future possibilities and critical research concerns for FL-CPS. To this end, our paper offers several distinct contributions to the emerging field of FL in FL-CPS:
\begin{itemize}
    \item We present the first comprehensive and systematic analysis of FL integration within CPS networks, synthesizing diverse research efforts to highlight synergies and opportunities between FL and CPS.
    
    \item We advance the current understanding by providing an in-depth comparison of the Internet of Things (IoT) and CPS, two closely related yet distinct paradigms, clarifying their differences and similarities to better evaluate FL's applicability in each.
    
    \item We explore FL applications in critical CPS domains, including smart cities, intelligent transportation systems, healthcare, and threat detection, and present detailed taxonomy tables that classify the technical components, contributions, and limitations of each.
    
    \item We go beyond summarizing existing research by critically evaluating the FL-CPS landscape, identifying key challenges and research gaps, thus offering a roadmap for future research in this field.
    
    \item Building on these gaps, we propose promising future research directions for FL-CPS, providing insights that promote ongoing development and innovation.
\end{itemize}

These distinctive contributions position our paper as an essential reference for both scholars and practitioners involved in FL-CPS research. Through its comprehensive analysis, thorough comparisons, and forward-looking insights, our work significantly enhances both the theoretical framework and practical understanding of FL-CPS, thereby supporting its ongoing evolution and advancement. 

\begin{table*}[h!]
\caption{Comparison with related works and our paper's key contributions.}
\label{table:related_works}
\renewcommand{\arraystretch}{1.1} 
\begin{tabular}{|m{2cm}|m{3.5cm}|m{6cm}|m{4.5cm}|}
\hline
\makecell[c]{\textbf{Related works}} & \makecell[c]{\textbf{Topic}} & \makecell[c]{\textbf{Key contributions}} & \makecell[c]{\textbf{Limitations}}\\ \hline

\centering \cite{zhang2021survey} & \centering FL for current applications & Examination of FL, including data partitioning, privacy mechanisms, ML models, communication architecture, and systems heterogeneity. & Primarily focuses on general FL applications, limited discussion on CPS-specific challenges. \\ \hline

\centering \cite{li2021survey} & \centering FL for data privacy and protection & Six-dimensional FL system taxonomy covering ML models, communication architectures, data dissemination, privacy methods, and federation scale. & Emphasis on privacy, less focus on other CPS integration aspects. \\ \hline

\centering \cite{aledhari2020federated} & \centering FL for enabling technologies & Investigation of FL focusing on hardware and software platforms, real-world applications, protocols, and practical use cases. & Broad overview, may lack depth in analyzing specific CPS challenges. \\ \hline

\centering \cite{abdulrahman2020survey} & \centering A new classification of FL fields & Novel categorization of FL subjects and research domains derived from a comprehensive analysis of technical obstacles and relevant ongoing research. & Focus on classifying FL fields, less on CPS integration specifics. \\ \hline

\centering \cite{lim2020federated} & \centering Mobile edge networks in FL & Application of FL to improve mobile edge networks, significant challenges and possible directions for further FL research. & Specific to mobile edge networks, may not generalize to other CPS. \\ \hline

\centering \cite{nguyen2022federated} & \centering FL for smart healthcare & Utilization of FL in intelligent healthcare, covering incentives, prerequisites, FL models, and implementations across diverse healthcare fields. & Domain-specific to healthcare, limited applicability to other CPS. \\ \hline

\centering \cite{banabilah2022federated} & \centering FL for current applications & Organization of FL literature around AI, NLP, blockchain technology, IoT, resource allocation, and autonomous vehicles. & Broad application focus, may lack in-depth CPS integration analysis. \\ \hline

\centering \cite{imteaj2021survey} & \centering FL for limited-resource IoT devices & Exploration of FL issues and emerging obstacles, particularly difficulties in implementing FL in diverse IoT environments. & Focus on resource-constrained IoT, may not cover broader CPS. \\ \hline

\centering \cite{kholod2020open} & \centering FL for IoT & Examination of open-source FL architectures and their use in IoT systems. & Primarily on open-source architectures, less on general FL-CPS integration. \\ \hline

\centering \cite{pham2021fusion} & \centering FL for IIoT & Examination of FL and IIoT integration, focusing on management of resources, data, and privacy for large and diverse datasets. & Specific to IIoT, limited generalization to other CPS domains. \\ \hline

\centering \cite{wang2021blockchain} & \centering Blockchained FL & Overview of blockchain technology deployment in FL, with focus on mechanism design and system composition. & Focuses on blockchain integration, not general FL-CPS challenges. \\ \hline

\centering \cite{aceto2019survey} & \centering ICT for Industry 4.0 & Analysis of complexity and multifaceted nature of Industry 4.0 paradigm in scientific literature. & Broader scope on Industry 4.0, less emphasis on specific FL-CPS issues. \\ \hline

\centering \cite{dafflon2021challenges} & \centering CPS for industrial manufacturing & Comprehension of obstacles, methods, and strategies related to CPS for manufacturing in Industry 4.0. & Domain-specific to industrial manufacturing, limited generalizability. \\ \hline

\centering \cite{ahmadi2017review} & \centering CPS standard architecture for manufacturing & Overview of CPS for manufacturing’s five-component design and each component’s manufacturing problems. & Focuses on architecture for manufacturing CPS, not general FL. \\ \hline

\centering Our paper & \centering A comprehensive study of FL-CPSs & 
\textbf{We present a comprehensive survey on FL-CPSs with key contributions:}
\begin{itemize}
    \item We examine FL research in CPS networks, comparing integration ideas and exploring achievements.
    \item We also compare IoT and CPS to show their similarities and differences.
    \item We then examine FL’s involvement in CPS applications such as smart cities, transportation, healthcare, and threat detection.
    \item We outline several major research challenges and discuss promising research solutions for future FL-CPSs.
\end{itemize}
&  \\ \hline

\end{tabular}
\end{table*}

\begin{table}[h!t]
\centering
	\caption{List of key acronyms.}
	\color{black}
		\label{Table:TableAcronyms}
    \renewcommand{\arraystretch}{1.1}
\begin{tabular}{|c|c|}
\hline
\textbf{Acronyms} & \textbf{Definitions}          \\ \hline
AFL & Asynchronous federated learning \\ \hline
AI & Artificial intelligence \\ \hline
AL & Active learning \\ \hline
BFL & Blockchain-based federated learning \\ \hline
C-FL & Consensus-driven FL \\ \hline
CFL & Centralized federated learning \\ \hline
CNN & Convolutional neural network \\ \hline
CPS & Cyber physical systems \\ \hline
DFL & Decentralized federated learning \\ \hline
DL & Deep learning \\ \hline
DNN & Deep neural networks \\ \hline
DP & Differential privacy \\ \hline
EHRs & Electronic health records \\ \hline
FL & Federated learning \\ \hline
FTL & Federated transfer learning \\ \hline
FedAvg & Federated averaging \\ \hline
GAN & Generative adversarial network \\ \hline
HCPS & Healthcare cyber physical systems \\ \hline
HE & Homomorphic encryption \\ \hline
HFL & Horizontal federated learning \\ \hline
IIoT & Industrial internet of things \\ \hline
ITS & Intelligent transportation systems \\ \hline
IoT & Internet of things \\ \hline
MDI & Medical data imaging \\ \hline
ML & Machine learning \\ \hline
NLP & Natural language processing \\ \hline
NN & Neural network \\ \hline
Non-IID & Non-independent and identically distributed \\ \hline
PFL & Personalized federated learning \\ \hline
QFL & Quantum-based federated learning \\ \hline
RNN & Recurrent neural network \\ \hline
SFL & Split federated learning \\ \hline
SG & Smart grid \\ \hline
SGD & Stochastic gradient descent \\ \hline
SL & Split learning \\ \hline
SVM & Support vector machine \\ \hline
TL & Transfer learning \\ \hline
UAVs & Unmanned aerial vehicles \\ \hline
V2I & Vehicle-to-Infrastructure \\ \hline
V2V & Vehicle-to-Vehicle \\ \hline
VCPS & Vehicular cyber physical systems \\ \hline
VFL & Vertical federated learning \\ \hline
\end{tabular}
\end{table}
\subsection{Structure of the Survey}
\begin{figure*}[htbp]
\centerline{\includegraphics[width=0.95\linewidth]{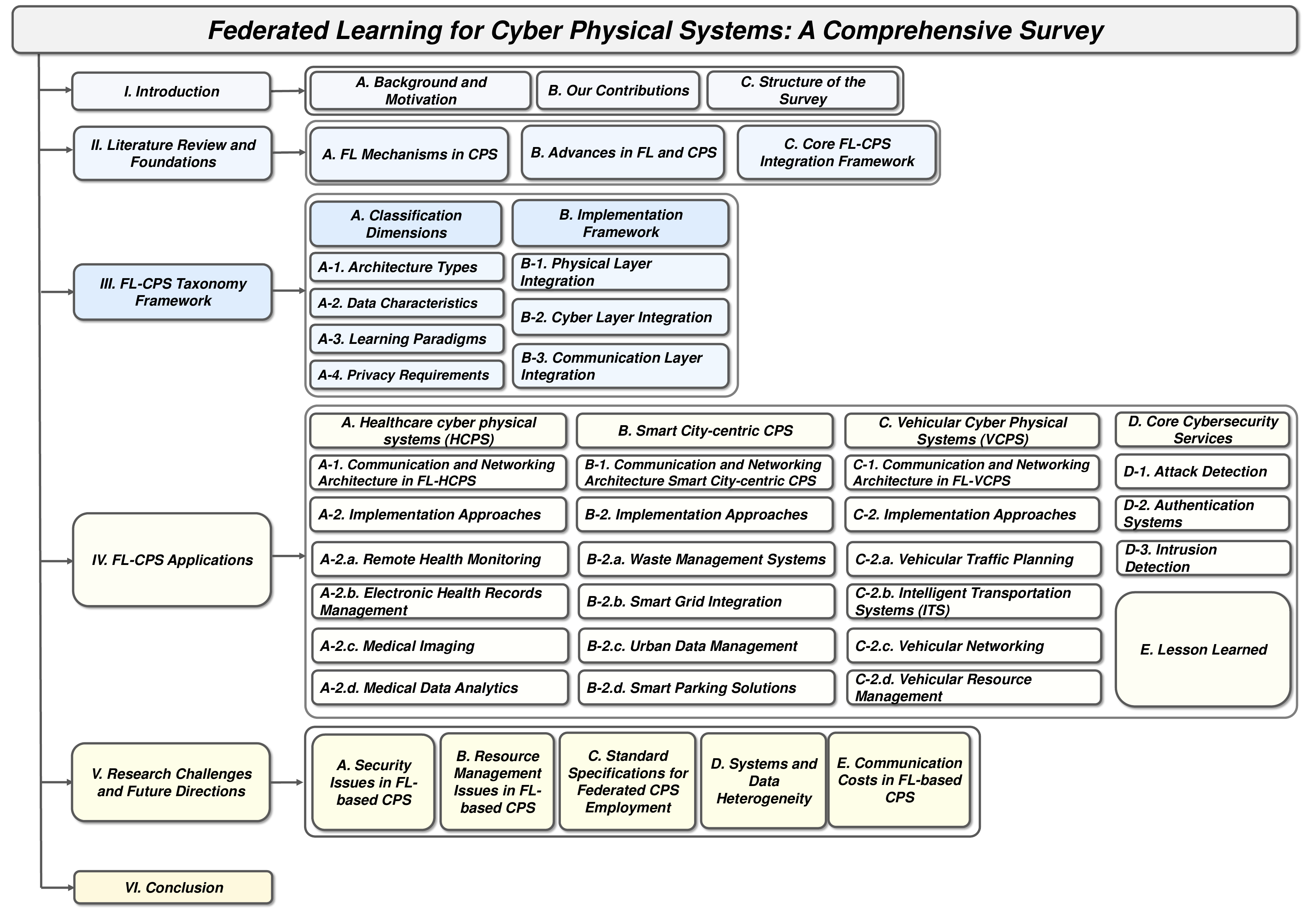}}
\caption{The structure of the survey.}
\label{fig0}
\end{figure*}

This work's remaining portions are organized as in Fig.~\ref{fig0}.
The second section offers an overview of FL and CPS. The key applications of FL on healthcare CPS are addressed in Section III. Section IV discusses FL for smart city-focused CPS. The majority of Section V is devoted to the FL of vehicular CPS. The deployment FL of cybersecurity and privacy services in CPS is then analyzed in Section VI. The seventh section highlights remaining concerns and future research areas. Section VIII is the final part of the paper. Moreover, Table \ref{Table:TableAcronyms} provides a list of the abbreviations and acronyms used throughout this study.

\section{Literature Review and Foundations}
\subsection{FL Mechanisms in CPS}
\subsubsection{What is CPS}
\paragraph{Typical CPS parts and features}
CPS is a combination of the physical and cyber worlds. Rather of emphasizing how they merge, the focus should be on where they overlap. Practically every discipline of engineering has been consulted in the development of its unique concepts and functioning models. However, because the prototypes did not fall within existing paradigms, CPS developed its own set of standards and procedures \cite{ahmadi2017review}. Information on each component of CPS must be collected jointly; otherwise, it would be fairly imbalanced. Consequently, an exhaustive analysis of the two together is required. To completely appreciate this CPS road plan, one needs examine the interconnection of digital and analogue systems, as well as software and hardware. Comprising control systems, sensors, and central processing units, these CPS are intricate and secure networked systems. It is intended for accurate sensing and physical system interaction. CPS has provided a remedy in the form of a shift away from hardware-based services and toward software-enabled services. It can think and communicate with humans and machines. The following are typical CPS parts \cite{singh2021emergence}:
\begin{itemize}
  \item \textit{Physical Parts}: This part addresses the system's logic, which consists of input-dependent algorithms and decision-making systems.
  \item \textit{Software Parts}: This part is responsible for the system's logic, which contains algorithms and techniques for making decisions based on the information provided into the system.
  \item \textit{Communication Parts}: This part supports data transfer over a network by linking various software and hardware components.
\end{itemize}

Regarding identifying aspects of CPS, the following are its four distinguishing features:
\begin{itemize}
  \item \textit{Adaptive System}: This demonstrates the adaptability and flexibility of CPS. The notion it has maintained its interactions with the environment at a value determined by the ecological system is indicative of the above.
  \item \textit{Systemic Integration}: Multiple continuous and discrete or independent behaviours coexist in the CPS. Consequently, CPS can be viewed as both a digital and analogue system. This allows for a greater degree of system mobility.
  \item \textit{Dedication to Software}: This characteristic yet unexpected inclusion aids in diminishing the relevance of some elements. The user experience of CPS is expected to be as efficient as possible.
  \item \textit{Reliable Approaches}: CPS devices exhibited a degree of responsiveness to unforeseen changes in their environments. It is also capable of adjusting to changes in the aggregate weight of observed and replaced data. In addition, it can adjust to changing requirements in real time.
\end{itemize}

Furthermore, to provide excellent services to users, CPS must communicate with devices over wired or wireless media; hence, it utilizes both wired and wireless communication capabilities to develop cutting-edge hardware and software.

\begin{figure*}[htbp]
\centerline{\includegraphics[width=0.99\linewidth]{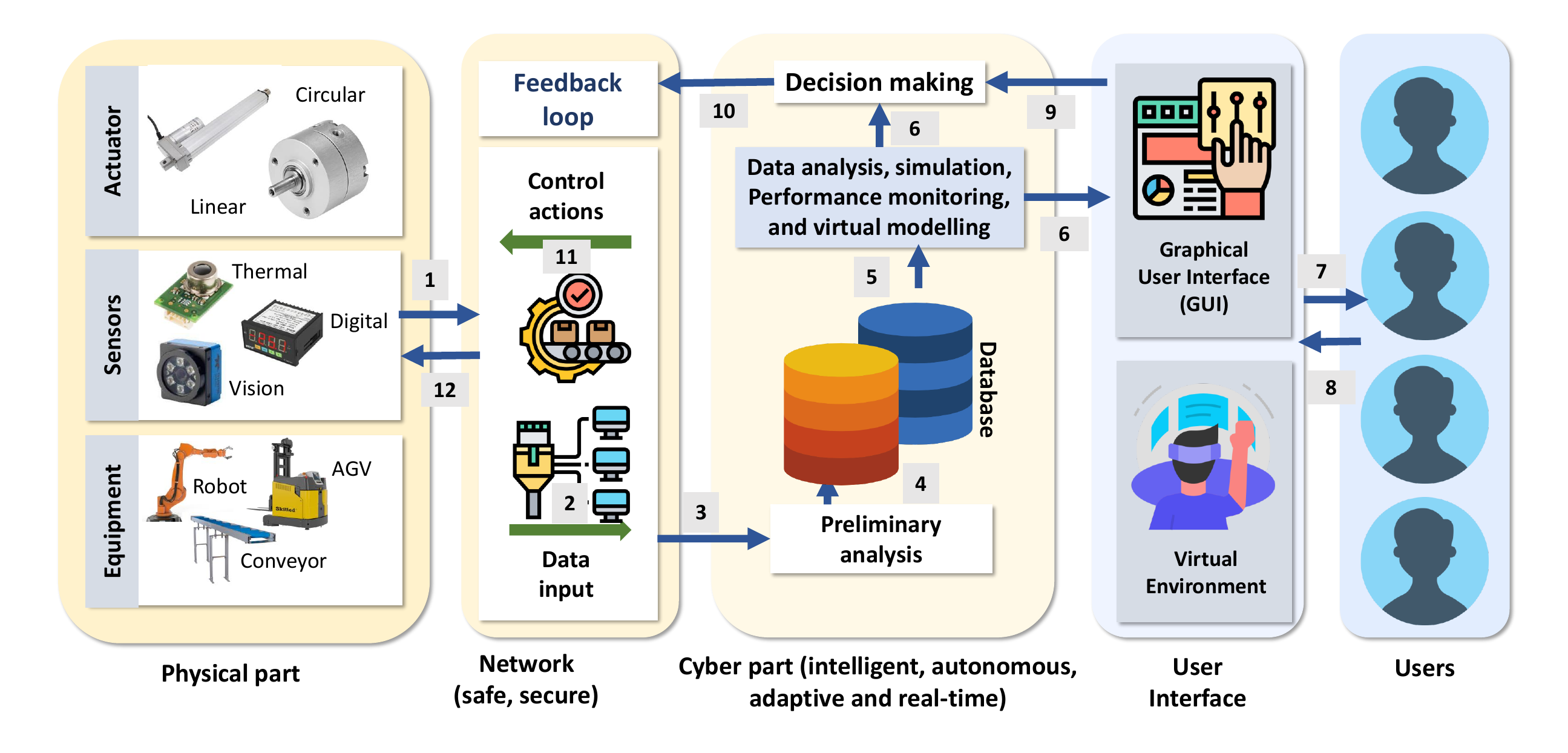}}
\caption{The overall architecture of a typical CPS.}
\label{fig9}
\end{figure*}

\paragraph{Infrastructure of CPS Architecture}
CPS architecture is the practice of designing and implementing the physical components of establishing a machine, a framework, or the proper type. A CPS has physical and cyber components, as shown in Fig. \ref{fig9}. This figure illustrates the dynamic interplay within a typical CPS. The process initiates at the Physical Part, where sensors gather real-time data from the environment. This data is then channeled through a secure network to the Cyber Part for analysis, simulation, and decision-making. The CPS, characterized by its intelligence, adaptability, and real-time responsiveness, utilizes this analyzed data to generate control actions that are fed back to the Physical Part through actuators, influencing the physical environment. The system also incorporates a User Interface, enabling human interaction and visualization of the CPS's operations. This continuous feedback loop between the physical and cyber domains, coupled with human oversight, allows the CPS to adapt and respond effectively to its environment. In particular, the CPS framework's architectural layers are introduced below to provide a high-level perspective:
\begin{enumerate}
  \item \textbf{Connection layer}: The connection layer is the key emphasis of a CPS network. We begin with sensor-collected real-time data. Sensors collect data about their surroundings and send it to other sensors via successive media. Transferring the data to the intelligent gateway for analysis. This layer enables plug-and-play functionality for information-receiving devices, e.g., a computer's mouse and keyboard or an integrated video card and hard drive. Additionally, CPS enables machine-to-machine connectivity between devices. The communication capabilities of CPS rely on these characteristics. In addition to system differences, sensors can also be differentiated by the data they collect.
  \item \textbf{Conversion layer}: Using analytics, analog data is converted to digital form. The data is gathered and organized using modern, regulated, and straightforward techniques. At this level, data are collected separately from other system layers. Analysis allows us to confirm a machine's prediction of health and anticipate its faults. Examining the statistical veracity of data combinations. The system makes it possible for CPS to possess consciousness.
  \item \textbf{Cyber layer}: The cyber layer of the CPS is employed for data backup and restoration. By storing previously completed tasks, it can function as a time machine. Previous versions of files can be backed up either automatically or manually. Before beginning data mining, it is necessary to gather from the entire system so that trends can be identified, duplicates can be removed, and the data can be arranged appropriately. At this level, data management and decision-making are handled.
  \item \textbf{Perception layer}: This is where troubleshooting happens if problems are found in the system. Simulations of error identification and correction are carried out by algorithms in this layer. Error and failure diagnosis are supervised by the cognitive layer's capabilities. Involvement from human beings is required for both visualizing the diagnosis and facilitating group decision-making once the final bug has been eliminated.
  \item \textbf{Configuration layer}: We can use CPS to investigate the system's availability and reliability. Output reporting is handled by this layer. The process of setting up the system also evaluates whether or not it can perform a resilience check throughout the process, adjust variances, or optimize faulty processes brought on by random disturbances. This level performs administrative tasks and ensures proper configuration settings throughout the system.
  \item \textbf{Commercial layer}: Regarding the provision of dependable CPS, the commercial level actively participates in the organization's management systems. Once the ultimate result has been acquired, the system's output, usefulness, and level of popularity may all increase. Additionally, this layer initiates development opportunities through coordinated efforts with customers and providers.
\end{enumerate}

\subsubsection{Key FL Concepts and Workflow}
\begin{figure}[htbp]
\centerline{\includegraphics[width=0.95\linewidth]{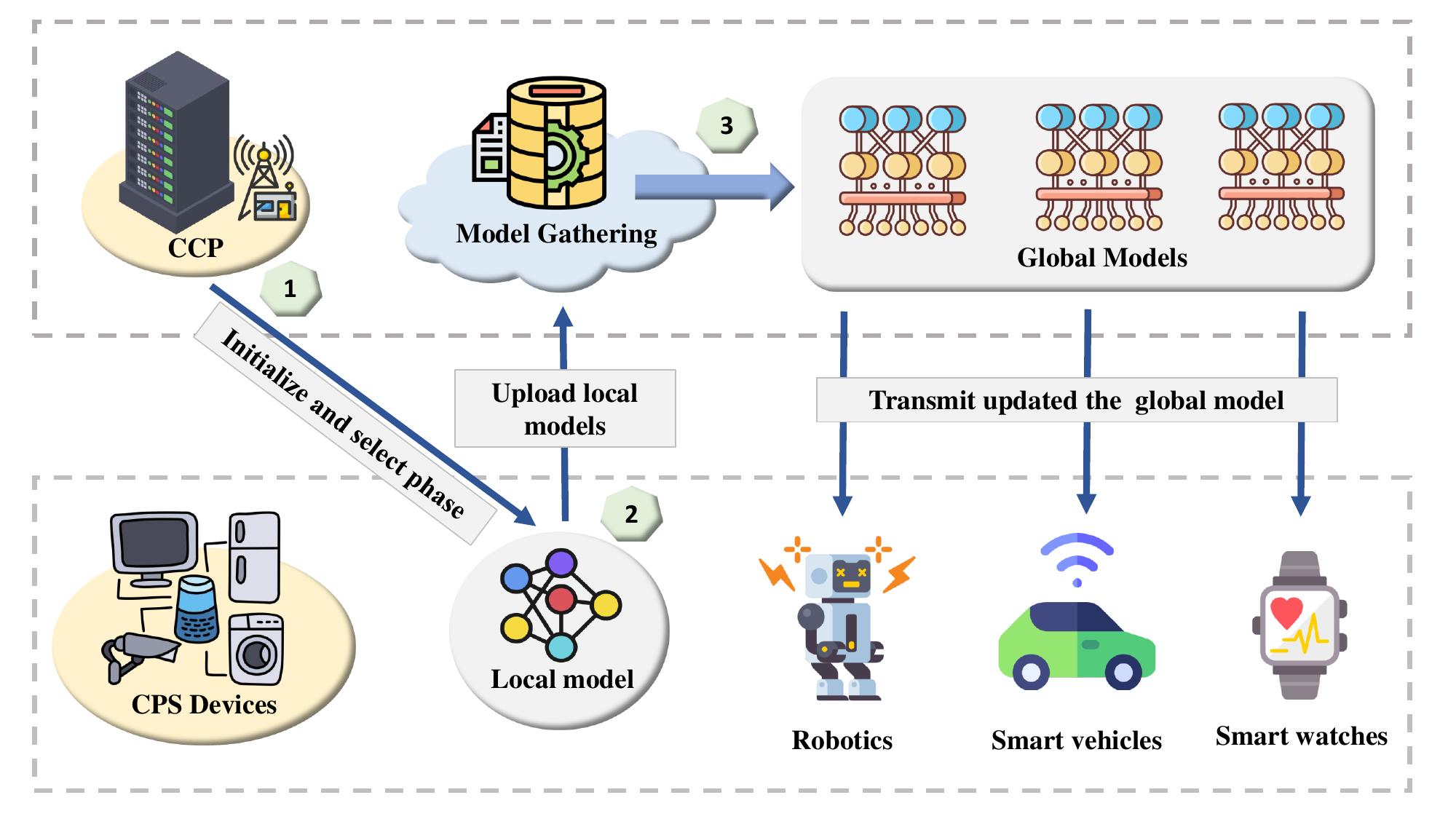}}
\caption{The communication process of FL in CPS.}
\label{fig2}
\end{figure}
As shown in Fig.~\ref{fig2}, FL within CPS involves data clients, like CPS devices, and an aggregator at a centralized point of connection (CCP). In this paradigm, each user, represented by $\mathsf{C_{i}}$, maintains a private dataset $\mathsf{S_{i}}$. These datasets consist of inputs $\mathsf{x_{ik}}$ and corresponding outputs $\mathsf{y_{ik}}$ for the FL algorithm. For instance, in a CPS network involving automobiles, collaboration on a central FL aggregation process enables data collection on surrounding traffic conditions to create a re-routing map, aiming to alleviate congestion. FL is crucial in CPS networks as a CCP cannot aggregate all inputs for ML training, ensuring comprehensive intelligence at the network's edge. Users and the CCP can conduct data analysis and inference within CPS networks. Individual FL models trained on user datasets are called local FL models $\mathsf{W_{local}}$, while central FL models use these local models to create global FL models $\mathsf{W_{global}}$. By leveraging dispersed data training at CPS actuators, the CCP can enhance predictive accuracy without significantly compromising user privacy. Fig.~\ref{fig2} below depicts the primary stages of a centralized FL (CFL), the most widely used FL model in literature that employs a centralized server for model aggregation:
\begin{enumerate}
  \item \textit{A strategy for initializing and selecting actuators}: FL aggregation at the CCP identifies a tasks in CPS, e.g., gesture recognition, and adjusts its learning parameters, e.g., learning speed and communication iterations. In addition, a selection of CPS actuators that will participate in the FL procedure is determined. The circumstances of the channels and the significance of the local updates of each CPS instrument are two of the plentiful selection parameters that might be used for this selection.
  
  \item \textit{Decentralized updates and training from the local users}: Once the training setup is complete, the server will generate an initialization model known as $\mathsf {w_{global}^{0}}$ and deliver it to the CPS users in order to initiate decentralized training. The protected dataset $\mathsf {S_{i}}$ is kept on node $\mathsf {C_{i}}$ of the CPS. Assuming that the NN's loss function is $\mathsf {L(.)}$, at time $\Delta$, $\mathsf {C_{i}}$ coordinates with other nodes to alter its weight based on its own dataset $\mathsf {S_{i}}$, the step size as $\mathsf {\pi}$, and the current weight $\mathsf {w_{i}^{\Delta}}$:
  \begin{equation} \label{eq:FLLocalUpdate}
    \mathsf{w_{i}^{\Delta+1} = w_{i}^{\Delta} - \pi^{\Delta}.\frac{\partial \text{ }L(w_{i}^{\Delta}.S_{i})}{\partial \text{ }w}} .
  \end{equation}
  One or several cycles of this localized update can be performed. Furthermore, different FL techniques might employ diverse L(.) functions \cite{konevcny2016federated}.

  \item \textit{Gathering and distributing models}: All of the endpoints' weights are consolidated in the server. With the support of the gathering function $\mathsf {M(.)}$, the centralized server calculates the average weights for the next iteration. The revised weights computed via redistributing the models at time $\mathsf{\Delta}$ can be written as:
  \begin{equation} \label{eq:FLGlobalUpdate}
    \mathsf{w^{\Delta} = M(w_{1}^{\Delta}, w_{2}^{\Delta},..., w_{n}^{\Delta})}.
  \end{equation}
  When updating the global model, FL aggregators frequently use a mean algorithm to combine all contributed weights into a single value. It is usual practice to make a copy of the model on each edge device that will be used for local prediction. We do not require that all $\mathsf{n}$ computing edges use the same manner of information exchange because edge devices are heterogeneous. This computation will only utilize a subset of the computing edges.
\end{enumerate}

\subsubsection{Common FL Algorithms for CPS Environments}
FL encompasses a suite of algorithms designed to enable collaborative model training while upholding data privacy in decentralized environments. Among these algorithms, Federated Averaging (FedAvg) \cite{konevcny2016federated} stands out as a cornerstone, wherein multiple clients iteratively compute local model parameters by minimizing their respective loss functions, transmit these parameters to a central server, which aggregates them to update the global model. This process ensures that the underlying data remains decentralized and private, making FedAvg widely applicable across domains like image classification and NLP. Personalized federated learning (PersFL) \cite{fallah2020personalized} extends this paradigm by tailoring the global model to individual client characteristics. In PersFL, each client's data is unique, necessitating adaptation of the global model through optimization of personalized loss functions. Consequently, PersFL enhances model performance and user satisfaction while addressing fairness considerations. pFedMe (Personalized Federated Mean Encoding) \cite{t2020personalized} further refines FL by encoding personalized information into the global model using client-specific embeddings. By integrating these embeddings into the global model parameters, pFedMe achieves enhanced model personalization while minimizing communication overhead, a critical consideration in resource-constrained FL settings. Secure Aggregation (SecAgg) \cite{liu2022sash} fortifies FL systems against privacy breaches by aggregating encrypted model updates, thereby safeguarding sensitive information during the aggregation process. Employing cryptographic techniques e.g., homomorphic encryption, SecAgg ensures robust privacy preservation while maintaining computational efficiency. Meta-Learning for FL (FedMeta) \cite{liu2023federated} introduces adaptability to new clients or tasks by learning meta-parameters and updating the global model based on meta-gradients computed from local updates. This facilitates rapid adaptation in dynamic FL environments, albeit requiring careful initialization and regularization to mitigate overfitting. Together, these algorithms cater to diverse FL scenarios, ranging from addressing data heterogeneity and privacy concerns to facilitating dynamic model adaptation, thus underpinning the broader landscape of collaborative and privacy-preserving ML.

\subsubsection{Evaluation Metrics for FL-CPS}
Evaluation metrics play a crucial role in assessing the performance, privacy, and fairness of FL models. Model performance metrics like accuracy, F1-score, precision, recall, and AUC-ROC (Area Under the Receiver Operating Characteristic Curve) quantify the predictive capability and discriminative power of FL models, as mathematically summarized in Table \ref{Table:EvaluationMetrics}. Communication metrics, e.g., the number of communication rounds and total data transmitted, optimize efficiency during client-server interactions. Privacy metrics, notably DP and robustness against attacks, evaluate privacy preservation and resilience to adversarial threats. Fairness metrics, like bias detection and equalized odds, ensure fairness across client groups.
\begin{table*}[h!t]
\centering
\caption{Model performance metrics for FL}
\label{Table:EvaluationMetrics}
\renewcommand{\arraystretch}{2.2}
\begin{tabular}{|c|c|p{6.5cm}|p{5cm}|} 
\hline
\textbf{Metric} & \textbf{Category} & \textbf{Mathematical Formulation} & \textbf{Explanation} \\ \hline
\multirow{5}{*}{Model Performance Metrics} & Accuracy & $\text{Accuracy} = \frac{\text{Number of correctly predicted instances}}{\text{Total number of instances}}$ & Proportion of correct predictions. \\ \cline{2-4} 
& F1-score & $F1 = 2 \times \frac{\text{Precision} \times \text{Recall}}{\text{Precision} + \text{Recall}}$ & Balances precision and recall. \\ \cline{2-4} 
& Precision & $\text{Precision} = \frac{\text{True Positives}}{\text{True Positives} + \text{False Positives}}$ & Proportion of positive predictions that are actually true. \\ \cline{2-4} 
& Recall & $\text{Recall} = \frac{\text{True Positives}}{\text{True Positives} + \text{False Negatives}}$ & Proportion of actual positives correctly identified (Sensitivity). \\ \cline{2-4} 
& AUC-ROC & AUC-ROC = $\int_{0}^{1} \text{Sensitivity}(FPR) \, d(FPR)$ & Overall ability to distinguish between classes, calculated as the integral of True Positive Rate (TPR) with respect to False Positive Rate (FPR). \\ \hline
\end{tabular}
\end{table*}

\subsection{Advances in FL and CPS}
Recent advancements in FL and CPS have laid the groundwork for their synergistic integration. This section synthesizes foundational developments in both domains, highlighting their convergence and remaining challenges.

\begin{figure*}[htbp]
\centerline{\includegraphics[width=0.95\linewidth]{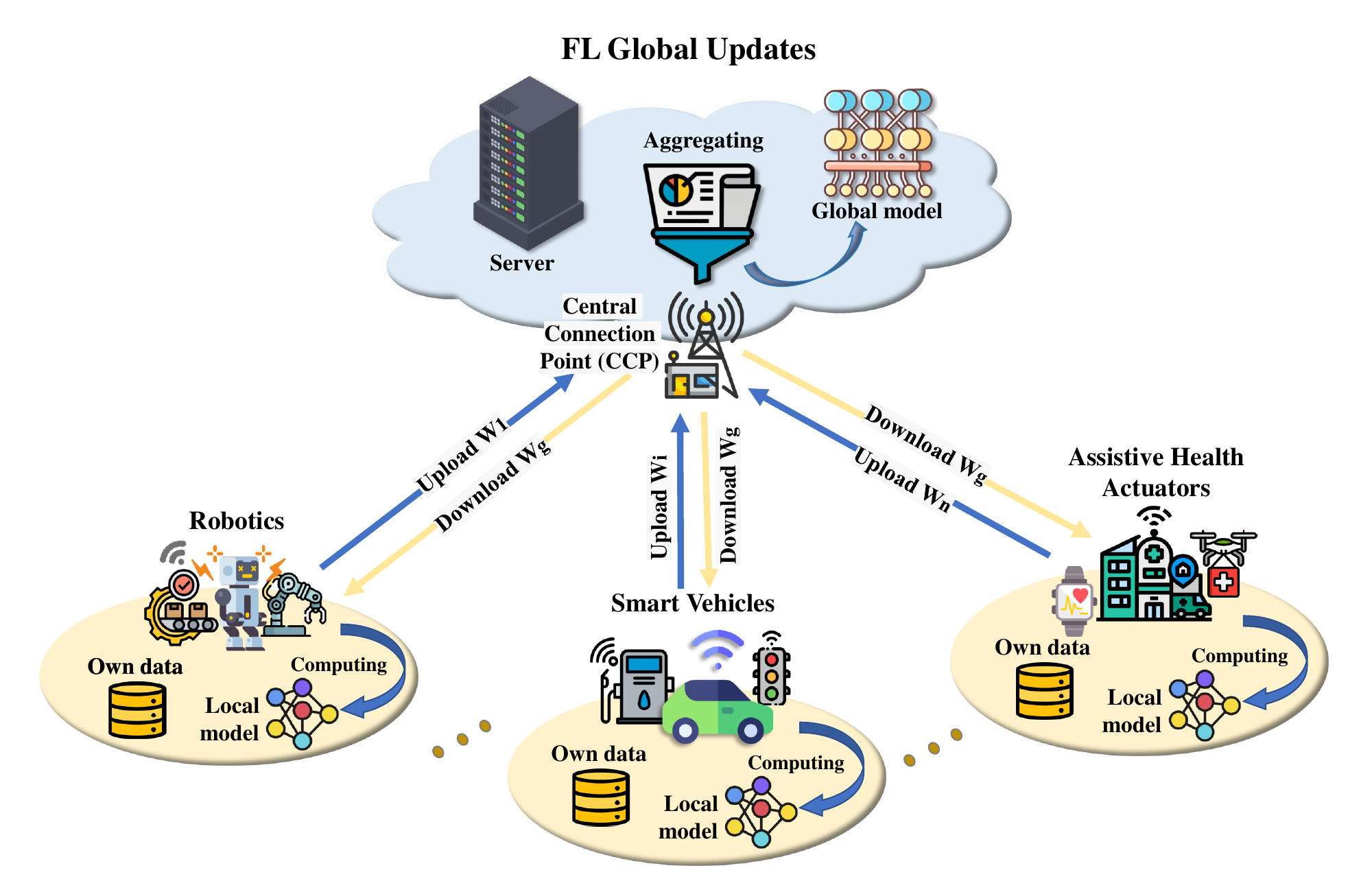}}
\caption{The overall architecture of a typical FL model in CPS.}
\label{fig1}
\end{figure*}

\subsubsection{Recent Advances in FL}
The four most recent variants of cutting-edge FL are FTL, split federated learning (SFL), personalized federated learning (PFL), and decentralized federated learning (DFL). The diversity of their clients, the emphasis on networking structure, and the manner in which training data are dispersed across the sample and feature spaces distinguish these four groups.
\begin{enumerate}
\item \textit{Personalized federated learning (PFL)}: Conventional FL poses significant difficulties for direct deployment in complex CPS due to model, device, and statistical diversity. For this reason, PFL is enabled in an infrastructure for intelligent CPS \cite{huang2022eefed}. PFL utilizes edge computing to perform calculations and communications at the edge \cite{zhou2019edge} in order to address the obstacles given by a multitude of devices with variable capabilities. To resolve the demand for high processing speed and low latency, each CPS device has the option of shifting its own high computational learning activity to the edge.
  \begin{itemize}
    \item \textit{Processes of PFL}: The shared learning process in PFL essentially comprises of the three processes listed below.
    \begin{figure}[htbp]
    \centerline{\includegraphics[width=0.8\linewidth]{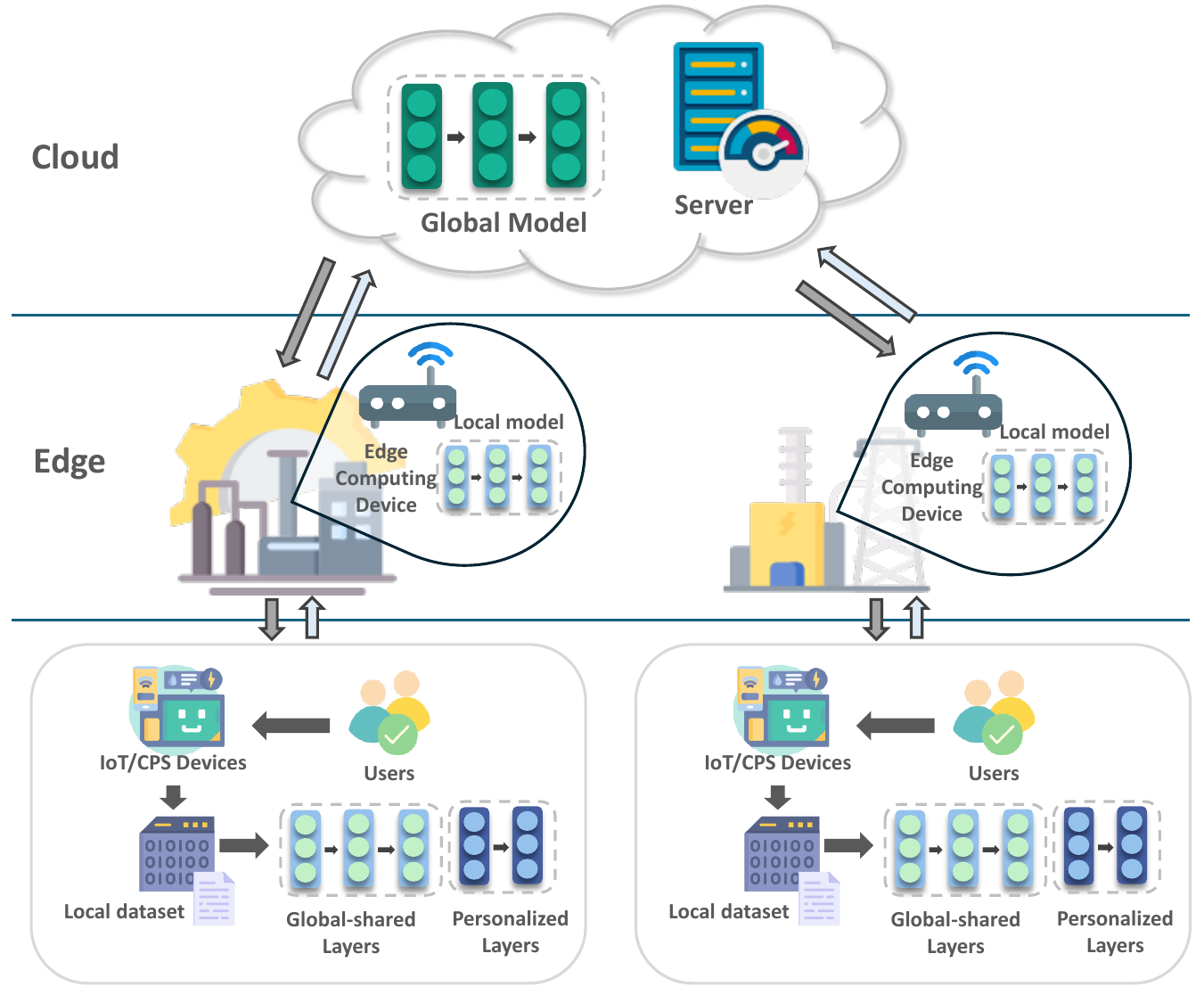}}
    \caption{Personalized FL: The device model integrates both the portion of the model transferred from the cloud server and the personalized layers maintained locally by the users.}
    \label{fig12}
    \end{figure}
    \begin{enumerate}
      \item \textit{Learning process}: Based on personal data samples, the edge devices compute its own model and can deliver this model to the server. Local model information from participating edges is averaged by the server into a common model and re-delivered. Following a predetermined number of iterations, model information exchange converges. Then, a global model can be delivered to the edges for customization.
      \item \textit{Off-loading process}: When the edge is dependable, operators of CPS devices can off-load their whole training model for computing efficiency (for example, a residential or industrial gateway \cite{hosseinalipour2022multi}). The operator will partition the model for device-edge collaborative computing, holding the data samples and input layers at server-side while sending the remaining work to the edge \cite{li2019edge}.
      \item \textit{Customization process}: In order to identify the specifics of each user's characteristics and needs, each device will build its own unique model using both generalized model data and localized data. At this point, the selected PFL method determines the precise learning procedures to be performed.
    \end{enumerate}
    \item \textit{Security and Privacy in PFL}: Integrating edge devices into PFL introduces security and privacy considerations that can be addressed through a multi-layered approach. Secure aggregation at the edge, employing multi-party computation (SMPC) \cite{zhu2020relationship}, prevents any single node from accessing raw updates during the aggregation process.  Edge-enforced differential privacy adds Gaussian noise to gradients, limiting data leakage risk while maintaining acceptable accuracy levels.  Trusted execution environments (TEEs) \cite{mo2021ppfl} isolate sensitive operations within encrypted memory enclaves, preventing adversarial servers from extracting raw data.  Federated authentication mechanisms, using elliptic-curve cryptography (ECC) certificates \cite{gajndran2024elliptic}, ensure only authorized devices participate in PFL rounds. 
    \item \textit{Strategies of PFL}: Intelligent CPS applications can incorporate customized learning approaches with PFL. PFL techniques, include FTL (which is discussed in next section), Federated Multi-task Learning (FML) \cite{smith2017federated}, Federated Meta Learning (FMeL) \cite{lin2020collaborative}, and Federated Distillation (FD) \cite{jeong2018communication}. The aforementioned techniques will decide how the server and edges share model data for collaborative model learning in PFL. Observably, the FTL configurations are only permitted to exchange a subset of model parameters. Using FD techniques, the server can get the output training set of local models trained on many CPS devices.
  \end{itemize}
\begin{figure}[htbp]
\centerline{\includegraphics[width=0.99\linewidth]{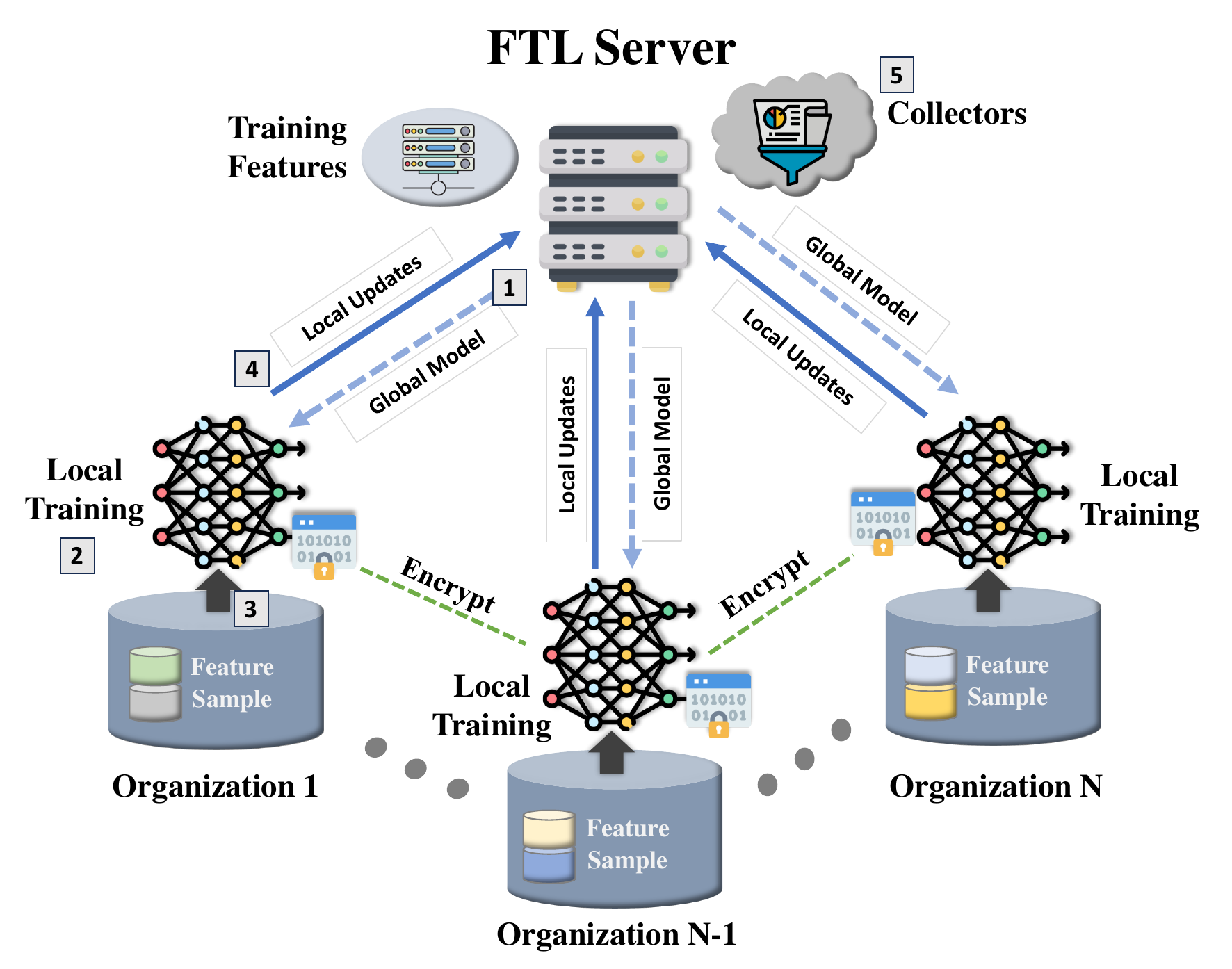}}
\caption{The workflow of FTL.}
\label{fig3}
\end{figure}
  \item \textit{Federated Transfer Learning (FTL)}: In essence, we constantly encounter circumstances in which there are insufficient common characteristics or samples to draw conclusions with confidence. By integrating a FL model with transfer learning (TL) strategies in which knowledge is shared between different entities with the same knowledge domain, this is capable of achieving better results. This combination of FL and TL is referred to as FTL \cite{liu2020secure}.
  \begin{itemize}
    \item \textit{The workflow of FTL}: To the best of our knowledge,  FTL addresses the challenge of data scarcity in federated settings by leveraging transfer learning techniques while preserving data privacy. It enables knowledge transfer from pre-trained models to new or data-limited clients, improving model performance even with insufficient local data. FTL is particularly valuable when there are discrepancies between the data distributions of different clients. In the FTL process, new and existing organizations collaborate, with existing organizations contributing their pre-trained models. A trusted executor facilitates secure model aggregation and transfer, often using cryptographic techniques to protect data privacy during communication. Through using gathered updates from the local learning to minimize a loss function \cite{sharma2019secure}, the executor can gather gradients and do loss convergence tests. As shown in Fig.~\ref{fig3}, new organizations leverage pre-trained models from existing organizations to address data scarcity. A trusted executor facilitates secure model aggregation and transfer. Both new and existing organizations perform local computations on their data, encrypting the results before sending them to the executor. The executor aggregates the encrypted updates, decrypts the results, and uses them to update the global model. Finally, the updated model is shared with all participating organizations, enabling them to further refine their models based on the transferred knowledge. In FTL, both the new and existing organizations compute the gradient and loss using their own data locally and encrypt the results. An encrypted collection of values is provided to the executor. Both new and established organizations use decrypted gradients and losses to refine their models. 
    \item \textit{Current applications of FTL for CPS}: In fact, FTL has a number of CPS applications, including federated healthcare learning \cite{majidi2021privacy}. In the case of disease diagnostics, for example, FTL can facilitate collaboration across multiple institutions with a diverse patient population and pharmacological assays, which are sample space and feature space, respectively. This allows FTL to enhance the accuracy rate of the outcome of the common AI model.
  \end{itemize}
  \begin{figure}[htbp]
  \centerline{\includegraphics[width=0.99\linewidth]{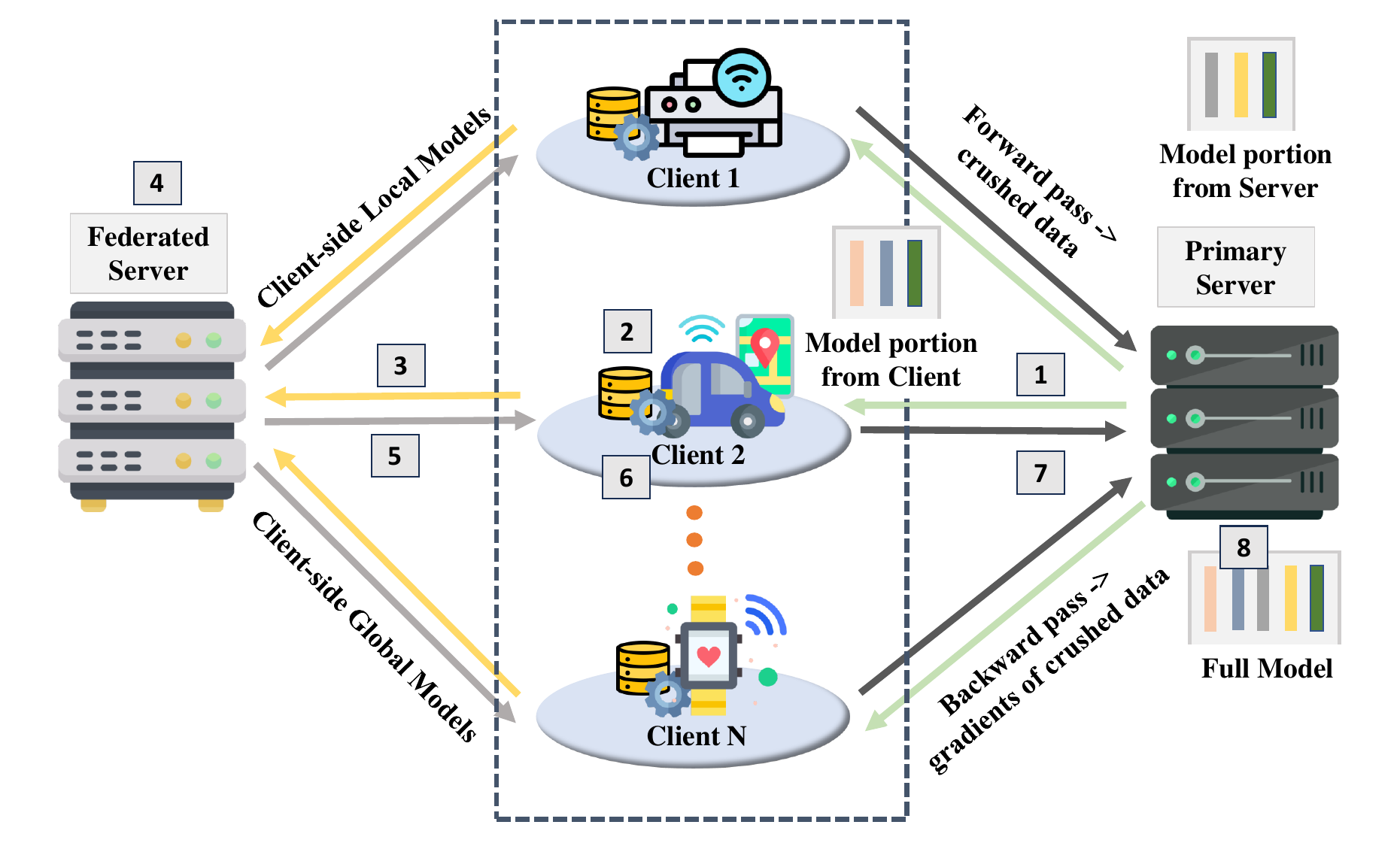}}
  \caption{The workflow of SFL.}
  \label{fig4}
  \end{figure}
  \item \textit{Split Federated Learning} (SFL): By amalgamating methodologies from both FL and split learning (SL), SFL circumvents issues with the demand for extra client-side computation, model privacy, and training time cost (SL) \cite{thapa2022splitfed}. The primary advantages of both FL and SL are combined in SFL, making it possible to train networks simultaneously on both the client and server sides. Fig.~\ref{fig4} represents the basic architecture of the SFL. In Fig. \ref{fig4}, the SFL workflow depicts a collaborative model training process where clients interact with both a primary and a federated server. Clients process their local data and send 'smashed data' to the primary server, which performs forward and backward propagation. Gradients are then sent back to the clients for local model updates. These updates are securely aggregated at the federated server using FedAvg, and the updated global model is shared back with all clients for further training iterations. In contrast to SL, all clients herein engage in simultaneous information exchange with both the primary and the federated server. 
  \begin{itemize}
    \item \textit{Components in SFL}: Whereas the primary server might be a cloud server or an expert with access to high-end computers, the clients might be a hospital or a CPS sensor with fewer computational resources. With the introduction of the federated server, FedAvg can be executed on a client's local updates. Furthermore, the federated server keeps the client-side global model up to date after each round of network training. FedAvg seems to be a very cost-effective service on the federated server \cite{zhou2021communication}. When put into practice, the federated server could very easily be situated toward the regional edge. In addition, if we implement all procedures on the primary server using encrypted data, it can carry out the responsibilities of the federated server (homomorphic encryption-based client-side model aggregation).
    \item \textit{The workflow of SFL}: All clients propagate their client-side model, including its disturbance layer, to the main server and send crushed data. The primary server does forward and back propagation synchronously with each client's crushed data. It sends clients crushed data gradients for backpropagation. The server updates its model by averaging weighted gradients computed during backpropagation on each client's crushed data. Once collecting its crushed data gradients, each client does backpropagation on its local client-side model. These gradients are sent to the federated server using a DP technique \cite{wei2020federated}. FedAvg is performed by the federated server and transmitted to all participating clients.
  \end{itemize}
  
  \item \textit{Decentralized federated learning (DFL)}: DFL is a serverless network design in which individual nodes control training. As an alternative, all participants are interconnected Peer-to-Peer (P2P) to train AI models. In a single communication cycle, participants execute their own training using their own local dataset. Afterwards, via P2P information exchange, participants gather the models that have been updated by their close nodes and reach a compromise on the global update \cite{savazzi2020federated}. There are two well-researched developing approaches, which are blockchain-based federated learning (BFL) and serverless federated learning (Serverless FL), within DFL's edge computing network. The following is a detailed overview of the aforementioned emerging approaches:
  \begin{itemize}
    \item \textit{Blockchain-based federated learning (BFL)}: FL and blockchain technology research has been undertaken at the leading-edge of the computer industry \cite{nguyen2022latency} \cite{wang2021blockchain}. The advantages of merging a blockchain-based approach with standard FL design have been extensively discussed \cite{nguyen2021federated} \cite{zhao2020privacy}. As a result of the distributed and secure nature of BFL, edge computing networks may undergo substantial transformations in the future \cite{majeed2019flchain}. As each BFL node functions as a client, it has complete access to the distributed learning model and may participate in its continuing refinement and consolidation. In the beginning, clients establish their own individual learning settings and data models. After the update has been generated, it is sent to a network of miners as transactions. After a certain period of time, miners compile all client-generated transactions, including any local alterations, into a single block prior to verifying it. Once a new block is formed, it is added to the ledger and the related message is sent to all peers. When each client obtains the updated block, the global model is updated to reflect the modifications made by each client. The procedure is continued until convergence of the global loss function is attained or an acceptable degree of precision is attained.
    
    \item \textit{Serverless federated learning (Serverless FL)}: As seen in \cite{chadha2020towards}, \cite{hegedHus2019gossip}, serverless has been used to characterize P2P FL. In contrast to conventional designs, in which data is gathered in a central data centre, this P2P kind of system gathers models over a wide area network (WAN). In the initial phase, a network is built utilizing current technologies e.g., broadcast/subscription \cite{lv2019iot} overlays and peer discovery \cite{sahal2020big}. After establishing a spanning tree over the P2P setup, updates are routed across the spanning tree and data is gathered at each node. Instead of developing overlays, broadcasting algorithms that rely on gossip are used to update and gather the models in gossip-based learning \cite{hegedHus2019gossip}. Although these methods are scalable and fault tolerant, they have drawbacks: exposing the model to additional entities during routing; (1) utilizing homomorphic encryption (HE) \cite{jayaram2020mystiko}, which can be problematic in terms of key exchange and model complexity expansion; or (2) employing differential privacy (DP) \cite{zhao2019differential}, which degrades model accuracy if hyperparameters are not carefully tuned.
  \end{itemize}
\end{enumerate}

\subsubsection{Recent Advances in CPS}
Computing and AI systems have dramatically evolved over the last two decades due to advancements in computer software and hardware technologies, which are not the only drivers in Information and Communication
Technologies (ICT)\cite{aceto2019survey}. CPS was also described as a tool for monitoring and modifying existing circumstances. The CPS will be indispensable for a vast array of upcoming applications. Strengthening a CPS over time involves ongoing innovation in theory, methodology, algorithm, mining approach, tool, etc. Several technological solutions, including those described below, are important to the CPS design of the present day.
\begin{itemize}
  \item \textit{AI-based CPS}: 
  When we integrate AI-based solutions with CPS in our daily lives, we get numerous benefits \cite{lv2021artificial}. CPS is evolving concurrently with the AI industry, as the success of such systems is largely determined by their data processing capabilities. According to \cite{plakhotnikov2020use}, CPS was implemented using AI. In \cite{khan2021cyber}, numerous CPS are analysed from a smart-city point of view. The DL model architecture presented in \cite{mandapati2022deep} was predicated on semi-supervised classification with feature learning. \cite{abunadi2021blockchain} is responsible for the development of this comprehensive health app. To counteract the propagation of COVID-19, which is detailed in greater depth in \cite{poonia2021cyber}, this application was developed for the healthcare industry to utilize blockchain business and technology management system. The author developed an integrated smart control approach \cite{farivar2019artificial} to rebuild and repair cyberattacks conducted by industrial IoT systems and non-linear CPS via wealth computer communications infrastructure. 
  CPS is regarded as nonlinear when just the forward channel is hackable. Using NN trained on Gaussian radial basis functions, this system analyses and reconstructs real-time cyberattacks from linked devices. The wise evaluator's law of adaptation is founded on the research of Lyapunov. The patient's home, the medical clinic, and the doctor's office were the most prevalent locations where CPS was utilized during medical treatment. New software reduces medical costs and improves quality of life for elderly people whose children work during the day. Using biosensors, a camera, and a local Cloud Hub, a central laboratory may examine an elderly person's daily physical examination data. Private clinics and hospitals can use authenticated wireless or cable data center connections to evaluate medical data, diagnose ailments, and prescribe treatments using AI. If the older were to fall, caregivers would be promptly notified.
  
  \item \textit{Agriculture-based CPS}:
  In agriculture, CPS can assist farmers in monitoring and controlling their crops and livestock in real time, allowing them to make informed decisions that optimize productivity while minimizing waste and environmental effect. The integration of wireless technology with CPS can improve these systems' efficacy and scalability. For instance, the study from \cite{zhang2021challenges} highlighted the uses and direction for future wireless technologies in rural regions, including digital agriculture, transportation, and residential welfare. This study also examined current and developing wireless technologies that potentially enable rural adoption in the context of their future integration with CPS. As another use of CPS in precision agriculture, the work \cite{rad2015smart} provides a CPS architecture model for tracking the vegetative development of a potato crop. This model is used for real-time monitoring of soil moisture, temperature, and nutrient levels. The model's acquired data is then used to maximize the utilization of water, fertilizer, and other resources. In addition, the CPS-based solution for digital agriculture published in \cite{fresco2018enhancing} highlighted the potential of CPS in crop monitoring and production prediction. CPS could be utilized to monitor crop growth, forecast crop yields, optimize farming operations, and boost crop productivity. Similarly, in \cite{an2017agriculture}, it was suggested that an automated irrigation system enabled by an agricultural CPS may be used to reduce water waste while still ensuring that crops get the right amount of water at the right time. Partitioning a field into several parcels and choosing selected parcels for sensor node deployment so that the covered field information is sufficient is the sensor deployment difficulty in agriculture-based CPS.
  
  \item \textit{Blockchain-based CPS}: 
  Owing to the complicates, constraints, and changing patterns of the interplay, centralized solutions for CPS systems are, to the best of our knowledge, incapable of tackling the distinctive difficulties of CPS. These distinctive characteristics need a decentralized strategy to discern the prospects of CPS. CPS solutions based on the blockchain are being considered. As a distributed ledger with the added benefits of anonymity and security, blockchain is supposed to be the solution to the problems plaguing CPS \cite{dedeoglu2020journey}. Blockchain is not widely used in CPS applications due to scalability, throughput, resource consumption, privacy, transaction delays, and trust difficulties. Understanding blockchain architectures and how they work helps us choose the best architecture for a use case \cite{zorzo2018dependable}. Due to its features, CPS are used in many situations \cite{darwish2018cyber}. Each use case has distinct restrictions and requirements, making blockchain-based CPS solutions difficult. All smart manufacturing devices initially detect data. The data from each device is uploaded to the basic CPS and can be transmitted. Each CPS at the fundamental level will share data and transmission logs. The basic CPS then transfers data to the integration CPS. The CPS connects and integrates all uploaded data. It audits and verifies the newly accessed CPS at the fundamental level. Thereafter, uploaded data is saved in the CPS at the level of system integration. Finally, the intelligent contract gets the data and generates a control signal for the intelligent manufacturing system. The smart contract grants users access to and retrieval of data from smart factories.
  
  \item \textit{CPS in 5G/6G wireless network}:
  Numerous prior papers \cite{viswanathan2020communications} \cite{nakamura20205g} have sought to characterize 6G based on the manner in which the digital, physical, and biological domains interact with one another, with an emphasis on the seamless merging of the physical and cyber worlds in real time. Physical space, cyberspace, and connectivity form the triangle's outer edges, while intelligence forms the triangle's foundation.
  \begin{itemize}
    \item \textit{Interplay}: Advanced sensing technologies allow humans to access virtual space and interact with its digital surroundings and other users as if they were physically present. Bulk density films, gesture/speech recognition, six degrees of freedom (6DF) spatial audios, multisensory integration, and haptic perception can aid users in having realistic cyberspace encounters.
    \item \textit{Cyberspace}: Not only is it viable to digitize physical space in the digital age, but also thought abilities. The digital revolution may network virtual worlds internationally. Cyber could be a physical location or an abstract concept. Physical objects have digital twins. Digital twins are ideas' digital counterparts. The Internet can therefore be viewed as either a digital or artificial twin. The Internet best portrays cyberspace. 6G cyberspace can penetrate the space of interest to experience it, unlike 5G Internet, which only offers search terms.
    \item \textit{Connectivity}: Communication links connect the digital and physical universes. Strengthening ties between these two could result in the development of novel scenarios and solutions that could have an impact on future industry and society. To have really immersive experiences, Internet users of the future will require spatial vision and tactile engagement. Latency, reliability, synchronization, and localization demand an unmatched data rate. 6G connectivity demands more stringent and immediate data rate, latency, synchronization, reliability, connection density, and localization standards.
    \item \textit{Intelligence}: Every component of the network, from the core to the air interface, will incorporate AI. As the network develops its offerings, AI will play a crucial part in a portion of them. AI will result in a network that is more autonomous, secure, and flexible, as well as software that is more individualized and private. AI can make it possible for 6G scenarios with rigorous criteria by anticipating and responding to changes in traffic patterns, channels, and user behavior. Consequently, intelligence and connectivity will pervade every facet of daily life.
  \end{itemize}
  The new Internet platform for 6G verticals, e.g., the metaverse or CPS-based applications, will feature a combination of the digital and physical worlds. Future communications technology will link humans, cyberspace, and the actual world. Thanks to the pervasive presence of AI, every component of a 6G devices might possibly be intelligent. 6G will place an emphasis on user interaction. The user will have a palpable and realistic experience. 6G terrestrial mobile systems merge physical space, cyberspace, connectivity, intelligence, and engagement, in contrast to previous generations.
\end{itemize}

\subsubsection{Emerging FL-CPS Convergence}
The fusion of FL and CPS offers a powerful synergy, addressing critical gaps in distributed intelligence and enabling a new era of intelligent and autonomous systems. This integration leverages the strengths of both paradigms, with FL providing privacy-preserving collaborative learning and CPS offering real-time sensing, actuation, and control capabilities. Architecturally, this fusion manifests in various forms. Hierarchical FL-CPS architectures \cite{liu2020client, quan2023hiersfl}, employing tiered aggregation across edge and cloud layers, optimize communication efficiency and scalability, as demonstrated in smart grid applications where communication costs are reduced by 55\% while maintaining high accuracy in load forecasting. Decentralized FL-CPS architectures leverage blockchain technology to achieve consensus and resilience against attacks, exemplified by vehicular networks where collision prediction accuracy remains high even under Byzantine attacks. These architectural synergies unlock new possibilities for distributed intelligence in diverse CPS domains.

The impact of FL-CPS integration is evident in its cross-domain applications. In healthcare, federated analysis of Electronic Health Records (EHR) across multiple hospitals has demonstrated significant improvements in predictive tasks like mortality prediction \cite{brisimi2018federated}. This showcases the potential of FL-CPS to revolutionize healthcare by enabling collaborative learning from sensitive data while adhering to strict privacy regulations. In smart cities, FL-trained traffic models, utilizing peer-to-peer gradient sharing among numerous edge nodes, have demonstrated the ability to optimize traffic flow and reduce congestion, leading to substantial improvements in urban mobility \cite{zuo2024exploring}. These examples highlight the transformative potential of FL-CPS across various sectors, enabling data-driven decision-making and intelligent control in complex and distributed environments.

\subsection{Core FL-CPS Integration Framework}
\subsubsection{Integration Workflow}
Integrating FL with CPS requires careful harmonization of distributed ML workflows with the real-time constraints and safety-critical nature of CPS. This integration involves a canonical workflow comprising five iterative phases, supported by key components that enable FL deployment in heterogeneous CPS environments.

The FL-CPS lifecycle begins with \textbf{Distributed Data Acquisition}, where diverse edge devices, such as sensors and actuators, collect real-time data streams (e.g., 1 kHz vibration in industrial robots, 30 fps thermal imaging). This data often requires preprocessing, including Kalman filtering for noise reduction \cite{badhwar2016noise}, feature extraction to enhance model performance (e.g., MFCC for audio CPS) \cite{cassara2022federated}, and differential privacy mechanisms to protect sensitive information.  Next, \textbf{Federated Model Initialization} involves deploying a baseline model (e.g., Long Short-Term Memory (LSTM) for predictive maintenance, ResNet-50 for visual CPS), with architectural parameters standardized via frameworks like NIST CPS Framework \cite{moller2023nist}:
\begin{equation*}
M_{global}(0) = f_{init}(\Theta_{arch}, D_{meta}),
\end{equation*}
where $\Theta_{arch}$ defines layer configurations. \textbf{Collaborative Model Training} then occurs, with devices performing local Stochastic Gradient Descent (SGD):
\begin{equation*}
w_{t+1}^k = w_t - \eta \nabla L(w_t; D_k).
\end{equation*}
Adaptive synchronization techniques like FedProx \cite{yuan2022convergence} handle stragglers:
\begin{equation*}
L_{prox} = L(w) + \frac{\mu}{2} ||w - w_t||^2.
\end{equation*}
Privacy is enforced through secure multi-party computation or homomorphic encryption.  \textbf{Secure Model Aggregation} follows, often using hierarchical aggregation (edge/cloud) with weighted averaging:
\begin{equation*}
w_{global} = \sum_{k=1}^K \frac{|D_k|}{\sum |D_k|} w_k.
\end{equation*}
Runtime verification techniques, using Linear Temporal Logic (LTL), validate the aggregated model against CPS safety constraints:
\begin{equation*}
C(\phi_{error} \rightarrow \Diamond \psi_{recovery}).
\end{equation*}
Finally, \textbf{Model Deployment \& Feedback} involves over-the-air updates to CPS actuators, with continuous monitoring and analysis via SHAP values \cite{hamilton2023using}:
\begin{equation*}
\phi_i(f,x) = \sum_{S \subseteq F \setminus \{i\}} \frac{|S|! (|F| - |S| - 1)!}{|F|!} [f(x_{S \cup \{i\}}) - f(x_S)].
\end{equation*}
This iterative process allows the FL-CPS system to adapt and improve over time, ensuring both model accuracy and system stability.

\subsubsection{Key Integration Components}
\begin{table}[h!t]
\centering
\caption{Components of the Proposed Framework and their Characteristics}
\label{Table:FrameworkComponents}
\renewcommand{\arraystretch}{1.6}

\begin{tabular}{|p{1.5cm}|p{4.2cm}|p{1.5cm}|} 
\hline
\textbf{Component} & \textbf{Description} & \textbf{Taxonomy Reference} \\ \hline
Data Layer & Time-series streams from IoT/CPS sensors & Section III.B.3 \\ \hline
Model Layer & Adaptive architectures (CNN/LSTM/Transformer) & Section III.B.3 \\ \hline
Aggregation Layer & Privacy-preserving parameter fusion & Section III.B.2 \\ \hline
Control Layer & Real-time model actuation (PLC/ROS integration) & Section III.B.1 \\ \hline
Verification Layer & Formal safety guarantees (UPPAAL/TLA+) & Section III.B.2 \\ \hline
\end{tabular}

\end{table}

Table \ref{Table:FrameworkComponents} details the key components of the integrated FL-CPS framework, outlining their functions and referencing their position within a broader taxonomy.  This framework is designed to address the unique challenges of training models on distributed, privacy-sensitive CPS data.  The Data Layer, the foundation of the system, handles the diverse time-series data streams originating from various IoT and CPS sensors.  Crucially, in an FL setting, this layer remains decentralized, with data residing on the individual devices, respecting data locality and privacy.  The Model Layer employs adaptive architectures like CNNs, LSTMs, or Transformers.  These models are trained locally on each device's data, enabling personalized learning and reducing the need for centralized data collection.  The Aggregation Layer is central to the FL process.  It implements privacy-preserving mechanisms, such as secure aggregation or differential privacy, to fuse model updates from different devices.  This allows for global model improvement without directly sharing the raw data, thereby preserving privacy.  The Control Layer, informed by the aggregated model, enables real-time actions, often integrating with programmable logic controllers (PLCs) or robot Operating system (ROS) for direct interaction with physical systems.  Finally, the Verification Layer offers formal safety guarantees, crucial for CPS deployments.  Using tools like UPPAAL \cite{larsen1997uppaal} and TLA+ \cite{kuppe2019tla+}, it provides mathematical assurance that the FL-trained model adheres to safety constraints, ensuring reliable and predictable system behavior.  These layers work together to create a secure and robust FL framework specifically tailored for the complexities of CPS environments.

\section{FL-CPS Taxonomy Framework}
\begin{figure}[htbp]
\centerline{\includegraphics[width=0.99\linewidth]{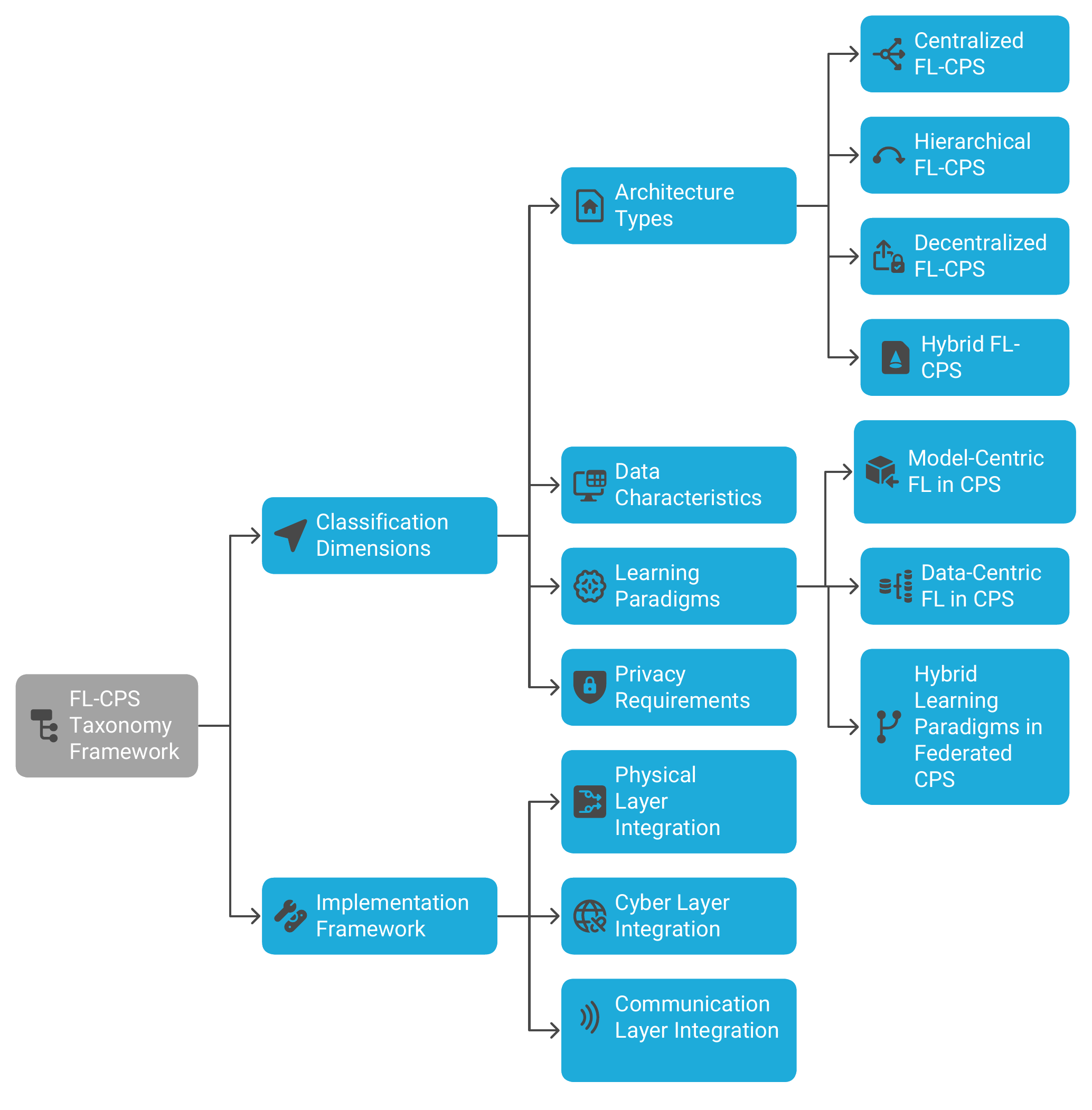}}
\caption{{\color{blue}FL-CPS taxonomy framework structure.}}
\label{fig13}
\end{figure}

The integration of FL with CPS introduces unique challenges in balancing distributed intelligence, real-time responsiveness, and privacy preservation. {\color{blue}As illustrated in Fig. \ref{fig13},} this section establishes a comprehensive taxonomy organized along two axes: classification dimensions (system architectures, data properties, learning paradigms, and privacy requirements), implementation frameworks (physical-cyber-communication integration), and validation metrics. The taxonomy synthesizes insights from NIST’s CPS Framework, empirical FL studies, and industrial CPS use cases to provide a systematic structure for designing and evaluating FL-CPS systems.

\subsection{Classification Dimensions}
\subsubsection{Architecture Types}
FL-CPS architectures are categorized based on computational hierarchy and coordination mechanisms:
\paragraph{Centralized FL-CPS}
Centralized FL-CPS architectures employ a single cloud- or edge-located aggregator to coordinate model training across geographically distributed CPS devices. This paradigm relies on synchronous aggregation mechanisms, such as FedAvg, to merge local model updates from edge nodes, sensors, or actuators into a global model. A centralized trust authority oversees critical functions, including model validation, anomaly detection, and security enforcement, ensuring consistency in federated operations. This architecture is particularly suited for homogeneous CPS networks—systems with uniform hardware specifications, data formats, and operational objectives—such as smart grids employing phasor measurement units (PMUs) for real-time energy monitoring. In such environments, the centralized aggregator mitigates computational asymmetry by standardizing model architectures and training protocols. For instance, in power grid voltage regulation, a cloud server aggregates load forecasting models from substation edge nodes, enabling grid-wide stability without exposing raw sensor data \cite{cobilean2023review}.

The centralized FL-CPS framework excels in applications requiring strict compliance and uniformity, such as critical infrastructure monitoring. In smart grids, PMUs generate time-synchronized measurements at 30–60 samples per second \cite{yang2013power}, which edge nodes preprocess into load forecasting models. A centralized cloud server aggregates these models using FedAvg. However, the architecture’s reliance on homogeneous networks limits its adaptability to systems with diverse sensors or conflicting operational priorities. For example, in industrial CPS combining vibration sensors and thermal cameras, centralized aggregation struggles with feature misalignment, often necessitating manual calibration. To enhance robustness, contemporary designs incorporate lightweight anomaly detection modules at the aggregator layer, flagging divergent updates using statistical divergence metrics like Kullback-Leibler (KL) divergence \cite{zhang2022innovation}. Privacy is enforced through secure multi-party computation (SMPC) \cite{zhou2024secure}, ensuring that neither the aggregator nor participants access raw gradients.

\paragraph{Hierarchical FL-CPS}
As shown in Fig.~\ref{fig1}, Hierarchical FL-CPS architectures employ multi-tier aggregation to balance computational efficiency and privacy preservation in distributed cyber-physical environments. This framework partitions the learning process into two distinct tiers: edge-layer preprocessing (Tier 1) and cloud-based refinement (Tier 2). At Tier 1, edge nodes—deployed at local hospitals, factories, or substations—perform real-time data preprocessing, including Kalman filtering for sensor noise reduction and convolutional feature extraction from raw sensor streams. These nodes train localized models on domain-specific subsets of data, such as MRI sequences from regional patient populations, while preserving data locality. The preprocessed model updates are then transmitted to a global aggregator at Tier 2, which applies homomorphic encryption to merge gradients without decrypting sensitive inputs, ensuring compliance with regulations like HIPAA and GDPR \cite{sohail2023data}. This hierarchical structure reduces communication overhead by 10–20\% compared to flat architectures, as edge layers filter irrelevant features before transmission \cite{wang2021resource}. The architecture’s tiered trust model also mitigates single-point failure risks, as compromised edge nodes cannot directly alter the global model.

A canonical application of hierarchical FL-CPS is medical imaging coordination across federated hospital networks, where privacy and diagnostic accuracy are paramount. In this use case, Tier 1 edge servers at individual hospitals train localized models on proprietary MRI datasets, applying differential privacy to gradient updates during feature extraction. Kalman filters at this tier reduce motion artifact noise \cite{shenoy2019sensor, mert2019design}, improving tumor boundary detection in 7T  magnetic resonance imaging (MRI) scans. The global aggregator at Tier 2 then homomorphically combines these updates using Brakerski-Fan-Vercauteren (BFV) encryption. This two-phase approach reduces data transmission costs, as edge nodes only transmit feature maps rather than raw 4D MRI volumes. However, the hierarchical paradigm introduces latency trade-offs: while edge-layer processing occurs in 12–18 ms per slice, homomorphic aggregation at the cloud tier requires 300–500 ms per global epoch, constrained by BFV’s computational complexity (O(n³) for ciphertext multiplications). To address this, recent implementations leverage hybrid encryption schemes, applying partial homomorphism only to sensitive model weights while aggregating non-sensitive biases in plaintext. Benchmarking on the BraTS dataset shows hierarchical FL-CPS achieves comparable accuracy to centralized training ($\Delta<$0.9\%) while reducing data exposure by 97\% \cite{zhou2024distributed}.

\paragraph{Decentralized FL-CPS}
Decentralized FL-CPS architectures eliminate central aggregators by employing P2P model sharing mechanisms, such as blockchain consensus protocols or epidemic gossip algorithms. With decentralized architectures, CPS devices—autonomous vehicles, drones, or industrial robots—directly exchange model updates with neighbors in a trustless environment, bypassing traditional server-client hierarchies. The architecture leverages blockchain’s immutable ledger to record gradient transactions, ensuring Byzantine fault tolerance against malicious nodes attempting model poisoning or data falsification \cite{gabrielli2023survey}. For latency-critical applications like vehicle-to-everything (V2X) networks, gossip protocols enable ultra-low latency synchronization by propagating updates through localized "infection" rather than global coordination. A key innovation lies in integrating federated learning with vehicular ad-hoc networks (VANETs), where autonomous cars collaboratively train traffic prediction models without roadside infrastructure. For instance, federated gated recurrent unit (GRU) models process LiDAR and radar streams from adjacent vehicles \cite{chellapandi2023federated}.

A prototypical implementation of decentralized FL-CPS is real-time traffic forecasting in autonomous vehicle platoons. Here, each car trains a GRU model on localized LiDAR sequences and shares parameters via IEEE 802.11p DSRC links using a gossip every $k$ seconds (GEKS) protocol \cite{wang2024advanced}. Blockchain smart contracts validate updates through practical Byzantine fault tolerance (pBFT) \cite{wu2023reinforced}, achieving consensus in 120–180 ms intervals across 50-node clusters. This setup demonstrates 94.7\% prediction accuracy for pedestrian trajectories while tolerating $\leq$16 adversarial vehicles in a 48-node fleet. Emerging solutions combine homomorphic hash commitments with zero-knowledge range proofs, enabling privacy-preserving aggregation without centralized oversight. Benchmarking on nuScenes datasets shows decentralized FL-CPS reduces collision false positives by 41\% versus centralized baselines, at the cost of 12–18\% higher energy consumption per vehicle \cite{lan2024bev}.

\paragraph{Hybrid FL-CPS}
Hybrid FL-CPS architectures dynamically adapt their coordination strategies between centralized and decentralized modes, optimizing resource allocation and responsiveness to fluctuating network conditions. This adaptive capability is governed by a supervisory control layer that monitors real-time metrics such as bandwidth availability, node reliability, and task criticality. In edge-centric operational modes, localized subsystems like industrial robotic cells employ lightweight CNNs for real-time anomaly detection. For instance, robotic arms in automotive assembly lines utilize 1D-CNNs \cite{singh2024hybrid}. These edge models operate autonomously during peak network congestion, minimizing dependence on cloud coordination. Conversely, during off-peak periods or stable connectivity, the architecture transitions to cloud-centric modes, enabling global model retraining across distributed CPS nodes.

In smart grid deployments, edge-layer models handle localized load fluctuations via 3-layer perceptrons, while cloud-tier LSTMs refine global demand curves during low-traffic intervals, compressing gradient updates via ternary quantization to save bandwidth \cite{premkumar2021survey}. However, synchronization challenges arise during mode transitions—temporal misalignment between edge and cloud models can induce prediction errors of 4–7\% in the first 30 seconds post-switch \cite{amgothu2024innovative}. To mitigate this, hybrid architectures implement versioned model repositories and rollback protocols, ensuring consistency across distributed nodes. Security remains a critical concern, as adaptive systems present larger attack surfaces. Current solutions integrate homomorphic encryption for policy updates and Byzantine-robust aggregation.

\subsubsection{Data Characteristics}
FL-CPS systems must accommodate heterogeneous CPS data streams (Table \ref{Table:CPSCharacteristics}):
This table summarizes key characteristics of CPS data and their influence on FL performance, along with potential mitigation strategies.  The table highlights four crucial data attributes: distribution, volume, variety, and velocity.  Firstly, the non-independent and identically distributed (Non-IID) nature of sensor data, exemplified by the disparate patterns of ECG and LiDAR sensors, can lead to model divergence in FL.  This issue can be addressed through personalized FL, incorporating device-specific layers into the global model.  Secondly, the high-frequency data volume generated by CPS devices can overload edge devices.  Split learning, offloading computationally intensive tasks to the cloud, offers a solution.  Thirdly, the multi-modal nature of CPS data, such as the combination of images and CAN bus logs, introduces feature misalignment.  Employing cross-modal attention networks can effectively fuse these diverse data streams. Finally, the varying data velocity, ranging from real-time microsecond-level data to hourly batch processing, can result in stale model deployments.  Online FL with incremental Stochastic Gradient Descent (SGD) can address this by enabling continuous model updates.  These characteristics and their corresponding mitigation strategies are critical considerations for deploying effective FL in CPS environments.

\begin{table}[h!t]
\centering
\caption{Characteristics of CPS Data and their Impact on FL}
\label{Table:CPSCharacteristics}
\renewcommand{\arraystretch}{1.2}

\begin{tabular}{|m{1.2cm}|p{1.8cm}|p{2cm}|p{2.2cm}|}
\hline
\makecell[c]{\textbf{Charact-}\\\textbf{eristics}} & \textbf{Definition} & \textbf{FL-CPS Impact} & \textbf{Mitigation Strategy} \\ \hline
Distribution & Non-IID sensor data (e.g., ECG vs. LiDAR patterns) & Model divergence & Personalized FL with device-specific layers \\ \hline
Volume & High-frequency CPS data & Edge device overload & Split learning offloading computations to cloud \\ \hline
Variety & Multi-modal streams (images + CAN bus logs) & Feature misalignment & Cross-modal attention networks \\ \hline
Velocity & Real-time ($\mu$s) vs. batch (hourly) processing & Stale model deployments & Online FL with incremental SGD\\ \hline
\end{tabular}

\end{table}

\subsubsection{Learning Paradigms}
\paragraph{Model-Centric FL in CPS}
Model-centric FL prioritizes global model optimization through parameter aggregation, emphasizing collaborative learning across distributed CPS. This paradigm employs algorithms like FedAvg \cite{zhou2021communication} and FedProx \cite{yuan2022convergence}, which synchronize local updates to refine a shared global model. FedProx addresses straggler nodes in industrial CPS—such as delayed updates from robotic arms with limited compute resources—by introducing a proximal term to the loss function, reducing gradient divergence and accelerating convergence by 15\% compared to FedAvg. Secure aggregation mechanisms, such as AES-256 encryption \cite{rahman2020secure}, ensure confidentiality during model transmission, critical for smart grid deployments where adversarial eavesdropping could compromise load forecasting models. For instance, in power grid CPS, encrypted FedAvg achieves 94.3\% prediction accuracy for day-ahead demand while preventing data reconstruction attacks \cite{shang2024fedpt}. However, model-centric FL struggles under extreme non-IID conditions, such as ECG data from cardiac patients versus LiDAR patterns from autonomous vehicles. These limitations stem from the assumption of data homogeneity across participants, which fails in heterogeneous CPS environments. To mitigate this, recent work integrates personalized layers (e.g., FedPer \cite{xu2022fedper++}) into global models, allowing device-specific adaptations while preserving collaborative learning benefits. 

\paragraph{Data-Centric FL in CPS}
Data-centric FL shifts focus to local data quality enhancement through preprocessing and privacy-preserving transformations, addressing inherent biases and imbalances in CPS datasets. Synthetic Minority Oversampling Technique (SMOTE) \cite{larsen2022synthetic} mitigates class imbalance in medical CPS, generating synthetic samples for rare conditions like pulmonary nodules. Differential privacy further safeguards sensitive data, with $\epsilon$=2.0 Gaussian noise added to X-ray gradient updates \cite{barnawi2024differentially}, achieving 80\% diagnostic accuracy while preventing membership inference attacks. This balance between utility and privacy is critical in healthcare CPS, where HIPAA compliance mandates strict patient confidentiality. However, according to \cite{barnawi2024differentially}, DP introduces noise-induced accuracy degradation—up to 12\% for $\epsilon<$1.0—necessitating adaptive noise scheduling based on data sensitivity. In automotive CPS, data-centric techniques align multi-modal features (e.g., LiDAR and camera streams) via contrastive learning. Despite these benefits, data-centric FL faces scalability challenges: SMOTE’s computational overhead grows cubically with feature dimensions, straining edge devices processing 4D MRI volumes. Hybrid approaches can reduce edge-layer latency. These strategies highlight the paradigm’s versatility in addressing CPS data heterogeneity but underscore the need for lightweight preprocessing frameworks to support resource-constrained devices like wearable sensors.

\paragraph{Hybrid Learning Paradigms in Federated CPS}
Hybrid learning architectures combine model- and data-centric approaches, leveraging split learning and meta-learning to address CPS complexity. Split learning partitions model training across tiers: resource-constrained devices (Tier 1) extract features via lightweight CNNs, while edge servers (Tier 2) train classifiers. For example, trash classification systems deploy MobileNetV2 on IoT bins for feature extraction \cite{yong2023application}, achieving 89\% accuracy on TrashNet with 23 ms inference latency. Meta-learning enhances adaptability through frameworks like Model-Agnostic Meta-Learning (MAML) \cite{finn2017model}, enabling rapid adaptation to new CPS configurations. In smart grids, MAML fine-tunes federated models for novel microgrid topologies in five shots, reducing retraining time from 12 hours to 18 minutes. This capability is vital for dynamic CPS environments, such as reconfigurable industrial lines or disaster-recovery networks. However, hybrid architectures face synchronization challenges, and solutions can include pipelined split learning for parallel tier updates and quantization-aware meta-training to compress models by 4×. 

\subsubsection{Privacy Requirements}
A hierarchical privacy framework is essential to address the diverse threat models inherent in federated learning for cyber-physical systems (FL-CPS), balancing security guarantees with computational efficiency. At the Basic tier, AES-256 encryption combined with secure aggregation protocols safeguards against eavesdropping during model transmission, introducing minimal overhead. This approach is particularly effective in smart grid deployments, where federated XGBoost models employing secure aggregation reduce mean absolute error (MAE) by 15\% in load forecasting \cite{le2021fedxgboost} while preventing adversarial interception of phasor measurement unit (PMU) data. The Enhanced tier integrates $\epsilon$-differential privacy with homomorphic encryption (HE) to counter membership inference attacks, which aim to deduce individual participation in training. While HE preserves privacy during cloud-based aggregation by enabling computations on ciphertexts, its polynomial computational complexity induces medical imaging—a trade-off mitigated through adaptive noise scheduling. For instance, healthcare CPS implement a tiered strategy \cite{plaza2018software} where edge nodes apply $\epsilon$=2.0 DP to MRI gradient updates, while cloud servers utilize HE for final aggregation, achieving HIPAA-compliant tumor detection without raw data exposure. The Maximum tier employs secure multi-party computation (SMPC) and zero-knowledge proofs (ZKPs) \cite{dhokrat2024framework} to thwart data reconstruction attacks, ensuring no single entity accesses complete gradient information. SMPC’s secret-sharing protocols, combined with ZKP-based authentication, enable privacy-preserving collaboration in military drone swarms but incur 5× higher compute costs due to multi-round cryptographic verification. Across all tiers, the framework emphasizes context-aware privacy-utility trade-offs. Therefore, emerging solutions, such as hybrid encryption dividing sensitive/non-sensitive model components.

\subsection{Implementation Framework}
\subsubsection{Physical Layer Integration}
Physical layer integration in FL-CPS ensures reliable coordination between sensors and actuators while adhering to stringent energy constraints inherent to edge devices. Sensor-actuator synchronization begins with NIST-compliant timing protocols, such as IEEE 1588 Precision Time Protocol (PTP) \cite{correll2005design}, which align sensor sampling and actuator responses to $\leq$1 $\mu$s precision in industrial robots. This temporal calibration is critical for closed-loop FL control systems, where delayed LiDAR-IMU data fusion in autonomous drones induces trajectory deviations. Federated Kalman filters \cite{hu2021novel} address this by synchronizing distributed sensor streams across drone swarms, achieving 0.12 m RMSE in 3D pose estimation through consensus-based covariance aggregation. Safety certification via formal methods like UPPAAL \cite{zhou2025comprehensive} verifies actuator responses under FL control loops, model-checking temporal logic properties to guarantee collision avoidance in dynamic environments.

Parallel to temporal coordination, energy-aware deployment strategies optimize FL operations for edge devices constrained by $\leq$10 W power budgets \cite{zhang2024deep}. Dynamic voltage and frequency scaling (DVFS) adapts processor clock rates to computational workloads \cite{liu2021dynamic}, reducing FL training energy by 37\% in smart dust sensor networks. This optimization leverages workload predictability in CPS—such as periodic vibration analysis in predictive maintenance—to scale voltages from 0.8 V (idle) to 1.2 V (peak inference). However, fluctuating energy availability complicates FL convergence, necessitating reinforcement learning-based schedulers that dynamically pause non-critical updates during low-energy states. These physical layer innovations collectively ensure FL-CPS systems meet industrial reliability standards (IEC 61508 SIL-2 \cite{fae2018standard}) while maintaining energy efficiency for deployability in resource-constrained environments.

\subsubsection{Cyber Layer Integration}
Cyber layer integration in FL-CPS orchestrates secure and efficient model aggregation while enforcing formal guarantees for convergence and robustness. The hierarchical averaging mechanism combines edge-layer (Tier 1) and cloud (Tier 2) updates through the weighted aggregation formula:
\begin{equation}
w_{global}(t+1) = \frac{1}{K} \sum_{k=1}^{K} (\alpha w_{edge,k}(t) + (1-\alpha) w_{cloud}(t)),
\end{equation}
where $\alpha = 0.7$ prioritizes edge contributions in healthcare CPS to accommodate non-IID data while retaining cloud-based regularization. For medical CPS, the weighting ensures federated models preserve hospital-specific diagnostic patterns while mitigating overfitting to outlier institutions. 

Hybrid encryption schemes further segment sensitive parameters (e.g., patient biomarkers) encrypted with AES-256 from non-sensitive weights (e.g., feature extractors) transmitted in plaintext \cite{man2019image}. Benchmarking on IEC 61850-compliant smart grids demonstrates this cyber-layer framework maintains 99.4\% model accuracy under 15\% adversarial participation, while runtime verification adds $\leq$8 ms overhead per aggregation round.

\subsubsection{Communication Layer Integration}
Communication layer integration in FL-CPS optimizes data transmission protocols and compression techniques to meet stringent latency, bandwidth, and energy constraints. In industrial CPS, Time-Sensitive Networking (TSN) protocols (IEEE 802.1Qbv) \cite{farkas2018time} guarantee deterministic latency for robotic arm coordination, synchronizing federated model updates across distributed assembly lines. TSN’s time-aware shaper allocates fixed time slots for gradient transmissions. For smart city deployments, bandwidth-efficient protocols like MQTT-SN (Message Queuing Telemetry Transport for Sensor Networks) \cite{bhardwaj2023message} reduce data payloads through 4:1 model pruning, discarding redundant parameters via magnitude-based filtering. In traffic management systems, pruned ResNet-50 models achieve 92\% vehicle detection accuracy while cutting camera-to-cloud bandwidth by 68\%, enabling city-wide FL deployments on existing 4G/LTE infrastructure.

Adaptive compression further enhances efficiency: quantized gradient descent (QGD) \cite{lin2021differentially} represents model updates using 8-bit fixed-point precision (vs. 32-bit floats), reducing transmission costs by 6$\times$ without significant accuracy loss ($\leq$0.5\% drop on MNIST benchmarks). This technique partitions gradients into sensitive and non-sensitive components, applying non-uniform quantization to minimize information loss in critical parameters (e.g., tumor detection layers in medical FL).

\section{FL-CPS Applications}

\subsection{Healthcare cyber physical systems (HCPS)}
HCPS are essential and interconnected networks of healthcare physical equipment potentially saving lives. Increasing numbers of hospitals are installing these smart integrated systems in order to give their patients with qualitative and uninterrupted treatment. Ensuring greater reliability in perspective intelligence, security and privacy, system software, autonomy, physical device stability, and interoperability are only some of the obstacles that have arisen from the necessity of implementing advanced HCPS that are effective and safe \cite{jimenez2020health}. Therefore, various publications on the application of FL approaches to HCPS are published in an effort to address the aforementioned issues. In the following sections, we will explore the various sectors where FL can be employed in HCPS, e.g., Remote Health Monitoring, Electronic Health Records (EHR) management, Medical Data Analytics, and Medical Imaging. Additionally, we will create two tables, Table \ref{Table:FL_HCPS_Applications} and Table \ref{Table:FL_HCPS_Applications_Cont}, which will offer a summary of these four applications. 

\subsubsection{Communication and Networking Architecture in FL-HCPS}
The integration of FL into Healthcare Cyber-Physical Systems (HCPS) necessitates a sophisticated communication and networking framework meticulously tailored to the specific characteristics of CPS and the sensitive nature of healthcare data.  In FL-CPS, healthcare devices function as critical nodes, forming a communication graph that embodies the dynamic interplay of cyber and physical elements within this healthcare context.  The mathematical representation of FL in CPS, characterized by iterative updates across these nodes, is illustrated as:  

\begin{equation}
\theta \leftarrow \theta - \eta \cdot \frac{1}{N} \sum_{i=1}^{N} \nabla f_i(\theta_i),
\end{equation}
where the learning rate (\(\eta\)) and the gradient of the local objective function (\(\nabla f_i(\theta_i)\)) are adapted to the distinctive features of healthcare data within CPS.  This mathematical formulation reflects the adaptive learning process within this healthcare context.  The communication graph, intricately connecting nodes and edges, symbolizes the collaborative nature of FL-CPS, with an associated cost function capturing both communication costs and cyber-physical relationships among devices:   

\begin{equation}
\text{Cost}(G, \theta_i) = \sum_{(i,j) \in G} c_{ij} \cdot \left | \left | \theta_i - \theta_j \right |\right |.
\end{equation}

This cost function provides a nuanced understanding of the CPS ecosystem dynamics. Specialized communication protocols, such as the lightweight MQTT and resource-efficient CoAP, are crucial for effective FL-CPS operation.  MQTT facilitates efficient model update transmission between healthcare devices and the central server, while CoAP ensures minimal bandwidth usage for resource-constrained devices. Security and privacy are paramount in FL-CPS, particularly given the sensitive nature of healthcare data.  Encryption mechanisms fortify communication channels, ensuring secure transmission of model updates and compliance with stringent healthcare privacy regulations.  The distributed nature of FL-CPS is further highlighted by nodes performing local model updates and data preprocessing at the edge.

In essence, FL in HCPS necessitates a tailored communication and networking framework that harmonizes with the unique characteristics of cyber-physical healthcare environments.  This integration optimizes model training across diverse devices and enhances the adaptability and responsiveness essential for effective operation within the complex CPS landscape. 

\subsubsection{Implementation Approaches}
\paragraph{Remote Health Monitoring}
In recent years, there has been a growing interest in innovative solutions for home-based care of critical medical conditions that necessitate immediate attention. FL can enable such home-based health monitoring by constructing a global model using data from geographically distributed home-based medical devices connected to a central server, e.g., a cloud server. FedHome, as presented in \cite{wu2020fedhome}, exemplifies this use case. By keeping user data localized on their devices, FedHome enhances data privacy and reduces the risk of centralized data breaches. In this framework, each home-based medical sensor, e.g., a smartphone or smartwatch, can potentially develop a personalized model using techniques like Convolutional Neural Networks (CNNs) by combining class-balanced datasets with local data. Model parameters are then iteratively refined through synchronized communication between the central server and the distributed devices. The outcome is improved individual predictions and a solution to the issues created by non-IID and imbalanced data. Using a real human behavior dataset, the author undertake thorough measurements that reveal the FL strategy for this situation can achieve a test accuracy of 95.45\%, a significant improvement of 7.49\% over the generic CNN approach with negligible overheads in both the imbalanced and balanced data scenarios.

Moreover, the work \cite{liu2019two} presents a FL-based HCPS designed exclusively for the phenotypic management of obesity and associated disorders. For the sake of clarity, the author proposes a federated two-stage natural language processing (NLP) technique, enabling shared healthcare data training using patient records from several medical institutions without requiring data exchange. At each facility, a patient representation model is developed prior to training a neural network (NN) to extract the healthcare common procedure coding from the records' text. Afterward, a clinical ML model is developed to conduct shared training over numerous locations for the desired attribute for diagnosis of diseases into one of three categories: absence, presence, or uncertainty. Extensive modeling in a FL context with ten hospital facilities indicates greater recall and accuracy than non-FL approaches. Likewise, FedHealth in \cite{chen2020fedhealth} is a new remote healthcare monitoring system. For the purpose of wearable health monitoring, this method has been developed in which smart sensors e.g., smartphones and tablets work collaboratively to build a distributed CNN classifier for gesture detection, while preserving the privacy of personal data. As model distributions across engineered devices and the cloud vary significantly, TL is employed to make the training process more specialized, which aids in the quest of individualization. The results of the constructed system reveal that the FTL method enhances wearables recognition accuracy by an additional 5.3\% over baseline approaches. The potential of this FTL approach can be considerably broadened in numerous captivating healthcare and medical use cases, including but not limited to, disease detection, predicting falls, and monitoring health conditions.

Additionally, the authors in \cite{tan2022tree} develop a FL-based remote healthcare tool for evaluating patients' responses to therapy based on data gathered from a decentralized hospital network. Each medical center develops its own ``custom therapeutic impact estimator". Then, each estimator can be placed in its own cluster, as the personal assessment impact contains the results on patient attributes and the facility indicator is utilized to establish the global clinical efficacy at the collaborating location. In the study referenced as \cite{liu2019confederated}, FL is used to forecast disease utilizing data gathered from global healthcare insurance at 99 medical facilities (including clinical labs and hospitals) in 34 distinct U.S. states. Included are EHRs for illnesses including diabetes, mental illness, and cardiovascular disease. In terms of confidentiality and accuracy, the FL technique is superior than both centralized learning and local training that does not involve any type of collaboration. Another examples of HCPS that can benefit from FL-based portable activities monitoring include assisted living and fall detection \cite{gudur2020federated}. 
The federated neural networks (FNNs) mentioned in  \cite{gudur2020federated} are a specific implementation or application within the broader framework of FL. While FL encompasses various ML algorithms and techniques for training models across decentralized devices, FNNs specifically refer to NN architectures optimized for FL settings. This FNNs for climbing, moving, seating, and even standing are trained utilizing a diverse dataset of behavioral attributes collected from a variety of interconnected computing actuators in the outdoors. In a data environment without IID, activities data are often stored in silos due to differences in data distributions and labeling. While FL demonstrates significant potential in enabling real-time health monitoring and prediction using data from wearable and home-based devices, its application extends to the realm of Electronic Health Records (EHRs), where privacy and data security are paramount. The following section explores the role of FL in addressing these challenges within the context of EHR management.

\paragraph{Electronic Health Records Management}
This section will highlight the critical role of FL in enabling privacy-preserving analytics and ML on sensitive EHR data. In contrast to Section IV.B's focus on real-time health monitoring, the applications discussed here demonstrate how FL can facilitate collaborative research and improved healthcare outcomes while upholding patient privacy and data security. In recent decades, the healthcare industry has made extensive use of deep learning (DL) technologies, specifically FL emerging approaches, to transform digital patient information extracted from EHRs into valuable insights for disease progression and health issues, thereby enhancing the quality of care delivered to patients and the efficacy of medical research. Leakage of sensitive information through the analysis of medical data is also one of the greatest issues that face conventional AI approaches and CPS integration. Indeed, healthcare EHR data are extraordinarily private and sensitive in comparison to other industries. In dynamic clinical situations where different parties, e.g., insurance companies and hospitals, obtain access to shared medical dataset as part of the standard necessities, which include data processing and analysis, the exclusion of extracted metadata from patient records is grossly inadequate to comply with privacy laws.

As a matter of fact, FL provides significantly more reliable solutions for efficient data extraction from EHRs by leveraging AI solutions to facilitate clinical care while adequately securing personal data through the collaboration of various clinical endpoints, e.g., healthcare providers and patients. To illustrate, the work \cite{hao2020privacy} introduces a resource-preserving and privacy-aware interactive FL-based methodology for an EHRs computational processing system including a cloud server and different healthcare facilities. Utilizing cloud computing, each medical center operates its NN model by using its EHRs. To protest the model parameter against without being memorization-exploited during the training process, a lightweight data perturbation approach by adding noise data is explored to ruffle training data. Even though attackers could access altered EHR data, obtaining or recovering the original data is also perplexing. In addition, authors in \cite{brisimi2018federated} propose a FL-based approach for predicting exacerbations for heart-disease patients by utilizing their historical EHRs. In particular, healthcare data from EHRs comprised of patients' portable medical devices and separated clinics are trained independently. Thereafter, to create a global model that fully leverages support vector machine (SVM) to develop an efficient classifier, the server compiles the local model's training parameters. This is intended to analyze and predict exacerbation of heart-disease patients for clinical resource planning without revealing the confidential data. FL is also used to simulate the decentralized EHR data processing across multiple hospital sites in \cite{choudhury2019differential}'s research, thereby protecting patient privacy from attackers. In \cite{wei2020federated}, a DP-based system is implemented in which each regional medical center utilizes artificial noises to mask dynamic changes from the local before uploading them back to the server in order to preserve sensitive data. Using logistic regression, SVM, and perceptron, the authors train FL to demonstrate that DP-based FL techniques achieve remarkable overall accuracy in comparison to conventional FL models. 

Furthermore, in order to enhance the training outcomes of EHRs in FL-based smart health services, the paper \cite{shao2019stochastic} proposes a statistical channel-based FL strategy. To adopt this method, the authors upload a negligible portion of local gradients from participating hospitals, selecting at random the channels with the greatest volatility. For purposes of model aggregation, only neuronal channels that underwent substantial change during the training loop are included. In the FL simulation, mortality prediction is modeled as a binary classification problem and resolved using NNs. The training convergence rate is significantly higher than with standard FL approaches. Another competitive method addressed in \cite{liu2018fadl} is the federated EHR administration. This includes 58 health organizations that collaborate utilizing shared data to predict patient mortality. The authors of \cite{huang2020loadaboost} apply an adaptive boosting strategy known as LoAdaBoost to enhance the efficacy of federated ML in both non-IID and IID distributed processing. They concentrate on reducing complexity in FL applications designed for forecasting mortality in EHRs. This research \cite{boughorbel2019federated} investigates the training of EHRs in FL-based models for healthcare institutions with recurrent neural network (RNN) for forecasting preterm-labor at three months employing organized samples with heterogeneity. In this study, the authors utilize 42 institutions, each of which updates the model refinements that act as localized parameters in the FL technique. Similarly, the federated EHRs frameworks deployed in the works \cite{papadopoulos2021privacy} and \cite{pfohl2019federated} guarantee the dependable and secure operation of current healthcare facilities. In a published research study \cite{pfohl2019federated}, it shows how integrating patients, healthcare providers, and medical facilities potentially lead to a federated mechanism that enhances the efficacy of healthcare data analysis. This is achieved by the establishment of a group of 20 institutions that share EHRs. Specifically, FL is then linked with blockchain for data interactions and model updating to assure training dataset privacy.

\begin{figure}[htbp]
\centerline{\includegraphics[width=0.99\linewidth]{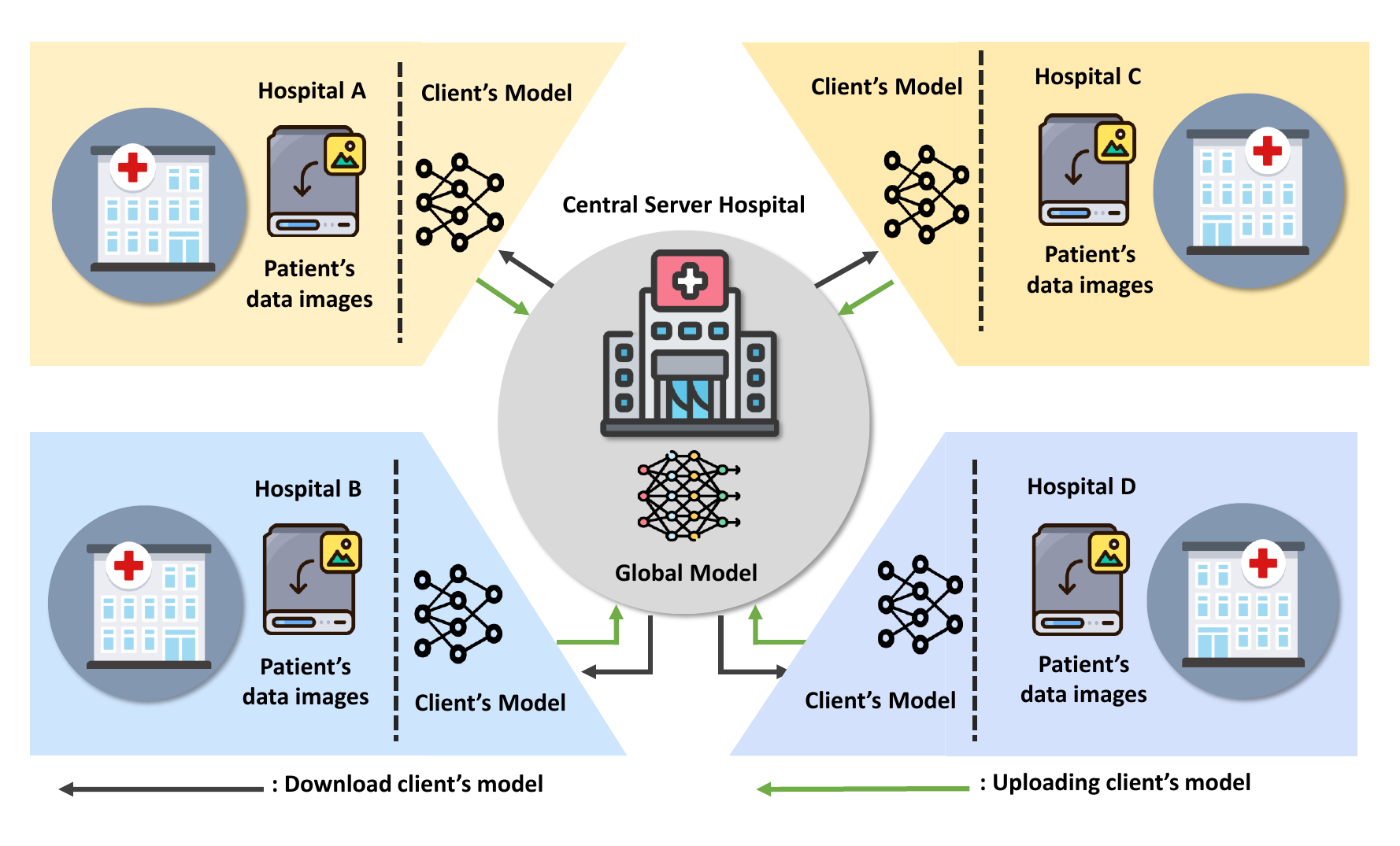}}
\caption{An overview of FL in medical imaging.}
\label{fig5}
\end{figure}

\paragraph{Medical Imaging}
Adopting ML-based Medical Data Imaging (MDI) through the consolidation of multiple medical institutions into a single organization is desirable but perplexing owing to concerns over patients' privacy. As depicted in Fig. \ref{fig5}, the FL diagram in medical imaging analysis is well-described, and FL has been considered a potentially advantageous alternative to existing methods for addressing large healthcare image analysis challenges due to its capacity to learn from the datasets of different physical actuators without direct communication. In \cite{yan2020variation}, a novel FL technique is developed to address discrepancies among clients, e.g., those arising from different clinical institutions. This approach maps all clients' images onto a commonly used image space using a global cloud predictor. Each organization can utilize its own generative adversarial network (GAN) to generate synthetic visual collections and transform pre-processed images to the targeted image space, effectively overcoming privacy challenges associated with client heterogeneity. Constructing a multi-source diffusion-coefficient visual collection has the potential to improve machine classification accuracy, although the observed improvement in their experiments on chronic prostatitis photos was 0.13\%. Additionally, in \cite{liu2022blockchain}, a sophisticated FL framework is proposed to aggregate a shared global model while ensuring data confidentiality within local hospital systems. Moreover, \cite{guo2021towards} presents a meticulously crafted scheme for training disease diagnosis models using distributed medical image datasets, detailing the roles of model providers, servers, and consumers in the FL process. Lastly, in \cite{pennisi2024feder}, the FedER strategy is introduced to enhance model generalization across diverse datasets while maintaining privacy.

In one piece of recent research, the study \cite{srivastava2020intracranial} presents how X-ray scan imaging in conjunction with a FL-based algorithm may be utilized to identify acute neurological symptoms e.g., severe headaches and loss of consciousness. In this study, numerous hospitals in the United States use DenseNet1212 models to train on the Radiological Society of North America's X-ray image dataset. This method promotes feature reuse and propagation while reducing the number of neural parameters. The implementation of dopamine-based DP in FL-based diagnostic imaging is thoroughly described in \cite{malekzadeh2021dopamine}. Only local clinics are entrusted by patients, thus they must be dependable and autonomous. In order for the server to construct an accurate global model with low data leakage rates, these clinics must provide patients access to their private information throughout each local update cycle. These clinics include Gaussian noise into the calculation of privacy loss in an effort to achieve a balance between accuracy loss and privacy cost. This maintains that each update has the same impact on the local learning speed as it does on the global classifier. More than 80\% improvement in classification accuracy was seen in simulations using a CNN-based SqueezeNet model applied to a real-world medical image dataset \cite{ucar2020covidiagnosis}. Collaborative MRI transformation involving students from different universities is described in ``FL-based Magnetic Reconstruction Imaging Reconstruction" (FL-MR) \cite{guo2021multi}. This can be done by first learning the distribution of intermediate latent features from a large number of hospitals, and then matching this distribution to the pattern of sensitive attributes at the target site. When an adversarial domain identifier is utilized, latent variables in the target domain can be more uniformly distributed thanks to the joint training of multiple locally-based reconstructive networks.

In addition, the study \cite{li2019privacy} provides a FL model for brain imaging with the intention of employing deep neural networks (DNN) to aid in brain tumor classification. Each federated provider, similar to an MRI scanner, constructs its own DNN architecture from locally accessible, dependable data and appropriate processing capacity, and then submits its weight updates to a central server, where they are averaged using a model averaging technique. FL ensures user privacy despite dangers e.g., server-side sample reconstruction training. Therefore, the implementation of a DP method during model exchange results in noise that disturbs the training process on all devices performing computations, impedes the transmission of vital information, and distorts updates. Using a preset set of multi-parametric MRI images from 285 patients with brain malignancies, the authors demonstrate that patient anonymity is maintained even when classification results deviate slightly from those obtained using the ideal central technique. To further minimize latency in a federated environment, the work in \cite{sheller2020federated} has additionally implemented an iteration reduction method based on the Alternating Direction Method of Multipliers (ADMM). Furthermore, the authors propose a federated multi-institutional collaboration for brain MRIs. Ten distinct research institutions are actively analysing brain images for suspected abnormalities using NN models. The authors of \cite{silva2019federated} present a novel technique to federated brain imaging using MRI data from many sites. A FL model can recreate a comprehensive infrastructure for confounding factor correction, high-dimensional feature variability measurement, and data processing using a centralized server and medical institutions as partners. The diagnosis, gender, age, and other demographic features of 181 patients with Mild Cognitive Impairment (MCI), 234 individuals with Alzheimer's disease, 232 individuals with Parkinson's disease, 208 individuals with severe MCI, and 455 healthy subjects are studied.

\begin{table*}[h!t]
	\centering
	\caption{Taxonomy of FL-based Solutions for Intelligent HCPS.}
	\color{black}
	\label{Table:FL_HCPS_Applications}
    \renewcommand{\arraystretch}{1.35}
    
		\begin{tabular}{|m{2.2cm}|m{0.5cm}|m{1.7cm}|m{2.2cm}|m{1.8cm}|m{3.2cm}|m{3.2cm}|}
			\hline
			\textbf{Applied Domain}& 
        \textbf{Ref.} &
			\textbf{System Architecture Type} &
			\textbf{Data Characteristics}& 	
			\textbf{Learning Paradigm}&
			\textbf{Key Contributions}&
			\textbf{Limitations}
			\\
			\hline
      \multirow{10}{*}{\hspace{4mm} \begin{tabular}[c]{@{}c@{}}Smart\\ Health \\ Monitoring\end{tabular}} & \centering{\cite{wu2020fedhome}} & Hierarchical (Edge-Cloud) & Non-IID wearable data; High velocity & Hybrid (Personalized + Split Learning) & Edge-cloud FL for remote monitoring (95.45\% accuracy); Privacy via HE & No formal convergence guarantees for non-IID data
			\\ \cline{2-7} & \centering{\cite{liu2019two}} & Centralized & Multi-modal EHR text; Non-IID & Model-Centric (Federated NLP) & Two-stage NLP for obesity analytics across 10 hospitals & Lacks DP guarantees for sensitive health data
			\\ \cline{2-7} & \centering{\cite{chen2020fedhealth}} & Hybrid (Edge-Cloud Split) & Heterogeneous sensor streams; Velocity: Real-time & Hybrid (Transfer Learning) & Adaptive TL for wearable gesture detection (+5.3\% accuracy) & High communication costs from split layers
			\\ \cline{2-7} & \centering{\cite{liu2019confederated}} & Decentralized (P2P) & Non-IID therapy responses; Multi-site & Data-Centric (Cluster Analysis) & Cluster-based efficacy estimation for decentralized networks & Vulnerable to Byzantine attacks (no robust aggregation)
			\\ \cline{2-7} & \centering{\cite{gudur2020federated}} & Centralized & Non-IID activity recognition; High velocity & Hybrid (FNNs + TL) & Federated activity recognition for assisted living (23 ms latency) & Basic AES-128 encryption lacks formal privacy proofs
			\\ \cline{2-7}
			\hline
  
			\multirow{18.5}{*}{\hspace{3mm} \begin{tabular}[c]{@{}c@{}}EHRs\\ Management \end{tabular}} & \centering{\cite{hao2020privacy}} & Centralized & Multi-modal EHRs + lab data; Non-IID & Model-Centric (Perceptron) & Lightweight perturbation against memorization attacks & Limited to $\epsilon$=5.0 DP with 8\% accuracy drop
			\\ \cline{2-7} & \centering{\cite{brisimi2018federated}} & Decentralized (Blockchain) & Multi-institutional EHRs; Non-IID & Hybrid (RL + FedAvg) & Blockchain-FL integration for auditability & 22\% higher energy costs than centralized FL
			\\ \cline{2-7} & \centering{\cite{choudhury2019differential}} & Centralized & Cardiac time-series; Non-IID & Model-Centric (SVM) & Heart disease exacerbation prediction via federated SVM & Requires trusted execution environments (TEEs)
			\\ \cline{2-7} & \centering{\cite{shao2019stochastic}} & Centralized & High-velocity ICU streams; Non-IID & Model-Centric (DNN) & Stochastic channel selection reduces privacy leaks & Unquantified DP budget ($\epsilon$ undefined)			
			\\ \cline{2-7} & \centering{\cite{liu2018fadl}} & Hierarchical (Edge-Cloud) & Multi-site EHRs; Non-IID & Hybrid (Global + Local Layers) & Specialized layers per hospital (92\% cross-site accuracy) & Complex synchronization protocol (300ms latency)
			\\ \cline{2-7} & \centering{\cite{huang2020loadaboost}} & Centralized & Class-imbalanced EHRs & Model-Centric (Adaptive Boosting) & LoAdaBoost handles non-IID data (F1=0.89) & Fails with extreme data skew ($>$50\% class imbalance)
			\\ \cline{2-7} & \centering{\cite{boughorbel2019federated}} & Centralized & Multi-modal pregnancy EHRs; Non-IID & Model-Centric (RNN) & Preterm labor prediction across 42 hospitals & No formal privacy guarantees (raw gradients shared)
			\\ \cline{2-7} & \centering{\cite{pfohl2019federated}} & Hybrid (Edge-Blockchain) & Multi-institutional EHRs; Non-IID & Model-Centric (Logistic Regression) & SMPC + ZKPs for maximum privacy tier & 5× compute overhead for cryptographic operations
         \\ \cline{2-7}
			\hline
	\end{tabular}
\end{table*}

\begin{table*}[h!t]
	\centering
	\caption{Taxonomy of FL-based Solutions for Intelligent HCPS (Cont.).}
	\color{black}
	\label{Table:FL_HCPS_Applications_Cont}
    \renewcommand{\arraystretch}{1.1}
    
		\begin{tabular}{|m{2.1cm}|m{0.8cm}|m{1.7cm}|m{2.2cm}|m{1.8cm}|m{3.2cm}|m{3.2cm}|}
			\hline
			\textbf{Applied Domain}& 
        \textbf{Ref.} &
			\textbf{System Architecture Type} &
			\textbf{Data Characteristics}& 	
			\textbf{Learning Paradigm}&
			\textbf{Key Contributions}&
			\textbf{Limitations}
			\\
			\hline
   \multirow{15.5}{*}{\hspace{5mm} \begin{tabular}[c]{@{}c@{}}Medical\\Imaging \end{tabular}} 
   & \centering{\cite{yan2020variation}} 
   & Decentralized 
   & Non-IID MRI scans; Multi-site 
   & Model-Centric (CNN) 
   & Privacy-preserving FL for cross-client variance reduction in brain MRI analysis 
   & No formal convergence guarantees for non-IID data
			\\ \cline{2-7} 
   & \centering{\cite{srivastava2020intracranial}} 
   & Centralized 
   & High-res CT scans; Non-IID 
   & Model-Centric (Pruned DenseNet) 
   & Architecture pruning enables low-resource deployment (73\% parameter reduction) 
   & Specialty-based data bias in hospital distributions
			\\ \cline{2-7} 
   & \centering{\cite{malekzadeh2021dopamine}} 
   & Centralized 
   & Chest X-rays; Class-imbalanced 
   & Data-Centric (DP-Augmented) 
   & Record-level differential privacy ($\epsilon$=2.1) for COVID-19 detection 
   & 8\% accuracy drop under DP constraints
			\\ \cline{2-7} 
   & \centering{\cite{ucar2020covidiagnosis}} 
   & Hybrid (Edge-Cloud) 
   & Multi-modal (X-rays + vitals) 
   & Hybrid (CNN + Regression) 
   & Embedded FL system for COVID-19 diagnosis (AUC=0.89) 
   & Limited validation on heterogeneous edge devices
      \\ \cline{2-7} 
   & \centering{\cite{guo2021multi}} 
   & Centralized 
   & Multi-contrast MRI sequences 
   & Model-Centric (U-Net) 
   & Federated ADMM optimization reduces reconstruction iterations by 41\% 
   & Tested on limited datasets (n=4)
			\\ \cline{2-7} 
   & \centering{\cite{sheller2020federated}} 
   & Centralized 
   & Brain tumor MRI; Non-IID 
   & Model-Centric (3D U-Net) 
   & BraTS benchmark validation across 10 institutions (Dice=0.85) 
   & No formal privacy guarantees for shared gradients
			\\ \cline{2-7} 
   & \centering{\cite{silva2019federated}} 
   & Decentralized 
   & Genomic imaging data; Non-IID 
   & Data-Centric (Cluster Analysis) 
   & Iteration reduction via ADMM optimization 
   & Limited scalability for large-scale genomics studies					
			\\ \cline{2-7}
			\hline
      \multirow{11}{*}{\hspace{5mm} \begin{tabular}[c]{@{}c@{}}Medical\\ Data \\ Analysis \end{tabular}} 
      & \centering{\cite{unal2021integration}} 
      & Hybrid (Blockchain) 
      & EHRs + sensor streams 
      & Model-Centric (SVM) 
      & Byzantine-robust FL with blockchain validation (92\% attack detection) 
      & 22\% higher energy costs vs centralized FL
      \\ \cline{2-7} 
      & \centering{\cite{dayan2021federated}} 
      & Hierarchical 
      & Multi-modal COVID-19 data 
      & Hybrid (CNN + XGBoost) 
      & Cross-site outcome prediction (AUC=0.92) with 16\% improvement 
      & Requires TEEs for secure aggregation
			\\ \cline{2-7} 
      & \centering{\cite{choudhury2019predicting}} 
      & Centralized 
      & Cardiac time-series EHRs 
      & Model-Centric (LSTM) 
      & Real-time exacerbation prediction (F1=0.87) 
      & Vulnerable to gradient inversion attacks
			\\ \cline{2-7} 
      & \centering{\cite{yi2020net}} 
      & Centralized 
      & Pancreatic CT scans 
      & Model-Centric (U-Net) 
      & Batch normalization improves tumor detection (Dice=0.79) 
      & High communication costs for 3D volumes
			\\ \cline{2-7} 
      & \centering{\cite{sheller2018multi}} 
      & Hybrid (FTL) 
      & Multi-institutional MRI 
      & Hybrid (Transfer Learning) 
      & First privacy-preserving multi-site brain aging study 
      & Lacks formal DP guarantees ($\epsilon$ undefined)
			\\ \cline{2-7} 
      & \centering{\cite{wang2020automated}} 
      & Centralized 
      & Surgical video datasets 
      & Model-Centric (3D CNN) 
      & Automated skill assessment without data sharing 
      & Fails under extreme non-IID instrument distributions 
			\\ \cline{2-7}
			\hline
	\end{tabular}
\end{table*}

\paragraph{Medical Data Analytics}
As part of a privacy-centric review of healthcare data analysis, the authors of \cite{unal2021integration} propose a robust FL-based framework for optimizing predicting processes in the setting of unwanted data injections. The plan lessens the risk of hazardous injections while maintaining anonymity. Model poisoning attacks, also known as targeted and false information injections, are the focus of this research into their detection and elimination procedures. Due to these solutions, data need not leave the premises of the data owner. By recognizing anomalies utilizing blockchain technology and fuzzified one-way hashes \cite{nguyen2020integration}, the research proposes a reliable approach for protecting the dataset's integrity during the training of the model on HCPS. Another noteworthy illustration of the practice of the FL method to medical prediction and data analysis is \cite{dayan2021federated}. This study builds a FL-based EXAM model that can predict the oxygen levels of patients with COVID-19 symptoms using chest X-rays, laboratory data, and vital signs from 20 institutions around the world. Consequently, it enables rapid collaboration in large data research without the need to transmit data and generates a model that determines the medical consequences of COVID-19 using data from a range of unstandardized sources. For predicting outcomes 24 and 72 hours after Emergency Room (ER) diagnosis, the Area Under the Curve (AUC) for this method is greater than 0.92, and it increases AUC by 16\% throughout all collaborating institutions and interpretability by 38\% when particularly in comparison to models trained at a single location with datasets from that location. In \cite{choudhury2019predicting}, the integration of FL-based solutions with electronic medical data to detect Adverse Drug Reactions (ADRs) over HCPS is also discussed. There is a shortage of aggregated data for identifying atypical ADRs, there are resource limitations when merging vast volumes of data from different entities, and there are privacy concerns when developing a database from personally identifiable information. This efficient method has presented two alternative ways of local model agglomeration, centred on class-imbalanced and loss-reduced issues, to maximise the overall improved forecasting skills. Furthermore, Schneble et al \cite{schneble2019attack} introduces an innovative intrusion detection system designed specifically for HCPS, emphasizing the utilization of FL to minimize communication and computation costs while maintaining high security standards. In contrast, Guo et el. from \cite{guo2021towards} proposes a systematic FL framework tailored for training disease diagnosis models using distributed medical image data, highlighting its potential to enhance collaboration and efficiency in healthcare settings. Additionally, Tripathy et al. in \cite{tripathy2024hybrid} presents a hybrid FL approach aimed at improving model accuracy by integrating diverse ML techniques.

Additionally, numerous studies examine into the use of FL in cancer treatment, with researchers alternatively comparing it to more traditional integrated data analysis procedures or offering novel approaches to its hurdles. Using the Kaggle ``Brain MRI Segmentation" dataset, which is divided into 5 ``client" locations, the paper \cite{yi2020net} successfully segmented low-grade gliomas. The network's state-of-the-art performance on the glioma segmentation problem is intact when applied in a FL-based context. For the purpose of brain tumor segmentation, the authors from \cite{sheller2018multi} simulate two distinct FL environments using the BraTS dataset \cite{menze2014multimodal}. When compared to the other two forms of collaborative learning, FL performs better in both contexts. It has about the same accuracy as a model trained on the whole dataset (99.1\%) even without decentralization. In \cite{wang2020automated}, the use of the FL method for data analysis in pancreas ectomy for pancreatic cancer patients is also discussed. Due to the fact that pancreatic cancer cancer is typically diagnosed at a later phase, resulting in relatively high mortality rates, it is of the paramount importance to accurately identify pancreatic cancer utilizing contemporary technology \cite{mcguigan2018pancreatic}. Using two sets of data derived from health facilities in Taiwan and Japan, they reconstruct a FL environment for paired medical actuators and the cloud server. The overall result, backed by FL in data analysis, outperforms trained models solely a single dataset and validates over the other in detecting pancreatic symptoms from both samples. In addition, ClusterFL in \cite{ouyang2022clusterfl} is driven by the FL framework to address the problem with current Human Activities Recognition (HAR) software, which is beneficial for older patients with Parkinson's disease. This FL-based system develops a rather accurate prediction model by evaluating vast amounts of data from both the digital and physical realms. ClusterFL is a modern grouping multi-task FL framework that dynamically captures the underlying clustering link between the dynamic computing nodes and decreases the practical training loss caused by numerous training patterns. ClusterFL is more precise than many current FL algorithms and can save network expenses by more than 50\%.

\subsection{Smart City-centric CPS}
When intelligent transportation, smart buildings, smart grids (SG), and emergency response technology and services are coupled, a smart city is generated. ``Smart cities" and other broad cyber-physical systems rely significantly on real-time offline and online behaviour monitoring sensors \cite{kesswani2017cyber}. It is essential to analyse the effects this has on the agents who can actually enhance cities. Rapid urbanization necessitates the creation of such infrastructure in order to meet the ever-increasing demands of inhabitants, as seen by the rise in housing expenses and average income. To fulﬁll expectations of smart city CPS growth, more and more AI-based solutions are being applied. FL provides enhanced features for distributed smart city services, e.g., high degrees of confidentiality and lower transmission latency. FL can assist with waste management, SG, data management, and smart parking in smart cities. Therefore, the following subsections will provide a scholarly analysis of the application of FL in smart city-centric CPS, and Table \ref{Table:FL_Smart_Cities_Applications} is designed to contrast the application of FL in the context of CPS that revolves around smart cities.

\begin{table*}[h!t]
	\centering
	\caption{Taxonomy of FL-based Solutions for Smart City-centric CPS.}
	\color{black}
	\label{Table:FL_Smart_Cities_Applications}
    \renewcommand{\arraystretch}{1.1}
    
		\begin{tabular}{|m{2.1cm}|m{0.7cm}|m{1.7cm}|m{2.2cm}|m{1.8cm}|m{3.2cm}|m{3.2cm}|}
			\hline
			\textbf{Applied Domain}& 
        \textbf{Ref.} &
			\textbf{System Architecture Type} &
			\textbf{Data Characteristics}& 	
			\textbf{Learning Paradigm}&
			\textbf{Key Contributions}&
			\textbf{Limitations}
			\\
			\hline
      \multirow{4.5}{*}{\hspace{3.5mm} \begin{tabular}[c]{@{}c@{}}Smart\\ Waste \\ Management \end{tabular}} 
      & \centering{\cite{vogiatzis2021dual}} 
      & Centralized 
      & Image data (Non-IID); Batch processing 
      & Model-Centric (CNN) 
      & Evaluated CNN architectures for waste classification (89\% accuracy) 
      & Untested on real-time edge deployments
			\\ \cline{2-7} 
      & \centering{\cite{ahmed2020active}} 
      & Hybrid (Edge-Cloud) 
      & Unlabeled sensor streams; Non-IID 
      & Hybrid (Active Learning) 
      & Reduced annotation costs by 37\% via federated active learning 
      & Unoptimized for intermittent connectivity
			\\ \cline{2-7}
			\hline
  
			\multirow{6}{*}{\hspace{7mm} \begin{tabular}[c]{@{}c@{}}Smart\\ Grids \end{tabular}} 
      & \centering{\cite{taik2020electrical}} 
      & Hierarchical (Edge-Cloud) 
      & Time-series energy data; Non-IID 
      & Model-Centric (LSTM) 
      & Edge-assisted FL for load forecasting (MAE: 12.8 kW) 
      & User-specific model variance persists post-personalization
			\\ \cline{2-7} 
      & \centering{\cite{liu2021federated}} 
      & Centralized 
      & Power usage patterns; Multi-modal 
      & Data-Centric (Cluster Analysis) 
      & Privacy-preserving power trace analysis without raw data exchange 
      & Lacks edge-cloud orchestration optimizations
			\\ \cline{2-7} 
      & \centering{\cite{wen2021feddetect}} 
      & Centralized 
      & Real-time consumption logs; High velocity 
      & Model-Centric (SVM) 
      & Anomaly detection in smart grids (F1=0.91) 
      & No formal convergence guarantees
         \\ \cline{2-7}
			\hline

      \multirow{5.8}{*}{\hspace{3mm} \begin{tabular}[c]{@{}c@{}}Smart\\ City Data \\ Management \end{tabular}} 
      & \centering{\cite{albaseer2020exploiting}} 
      & Hybrid (Edge-Fog) 
      & Unlabeled vehicular telemetry; Non-IID 
      & Data-Centric (Semi-Supervised) 
      & Achieved 84\% labeling accuracy via federated self-training 
      & High computation overhead on resource-constrained vehicles
			\\ \cline{2-7} 
      & \centering{\cite{chiu2020semisupervised}} 
      & Centralized 
      & Multi-modal traffic data; Non-IID 
      & Hybrid (FedSwap) 
      & Addressed non-IID data via model parameter swapping 
      & 18\% latency increase from swap operations
			\\ \cline{2-7} 
      & \centering{\cite{mukhametov2020ubiquitous}} 
      & Hierarchical (Edge-Cloud) 
      & Infrastructure sensor streams; High velocity 
      & Model-Centric (DNN) 
      & Real-time urban monitoring with 92ms inference latency 
      & Vulnerable to gradient inversion attacks
         \\ \cline{2-7}
			\hline

      \multirow{6}{*}{\hspace{6mm} \begin{tabular}[c]{@{}c@{}}Smart\\ Parking \end{tabular}} 
      & \centering{\cite{huang2021fedparking}} 
      & Centralized 
      & Parking occupancy data; Non-IID 
      & Hybrid (LSTM + FL) 
      & Reduced PLO estimation error by 22\% via collaborative learning 
      & Lacks adversarial robustness analysis
			\\ \cline{2-7} 
      & \centering{\cite{lim2020multi}} 
      & Hybrid (Edge-Cloud) 
      & Multi-agent vehicle trajectories; Non-IID 
      & Model-Centric (SVM) 
      & Federated optimization for sparse trajectory alignment 
      & No convergence guarantees for asynchronous updates
			\\ \cline{2-7} 
      & \centering{\cite{pham2021uav}} 
      & Centralized 
      & UAV-captured parking images; Non-IID 
      & Data-Centric (DP-Augmented) 
      & $\epsilon$=1.2 DP for occupancy privacy 
      & 7\% accuracy drop under DP constraints
         \\ \cline{2-7}
			\hline
	\end{tabular}
\end{table*}

\subsubsection{Communication and Networking Architecture Smart City-centric CPS}
The orchestration of communication and networking in FL within Smart City-centric CPS intricately balances mathematical formulations and technological strategies. Within the communication graph \(G = (V, E)\) representing device interactions, the interplay of network topology, graph theory, and communication costs becomes pivotal. This section will delve into a specific cost of communication formula tailored to the dynamics of FL in Smart City CPS. The refined cost function elegantly captures the nuanced aspects of transmitting model updates in this interconnected environment.

\begin{table}[h!t]
	\centering
	\caption{Communication Cost Function for FL in Smart City CPS.} 
	\color{black}
		\label{Table:FL_Smart_Cities_Communication}
    \renewcommand{\arraystretch}{1.1}
		\begin{tabular}{|m{2.5cm}|m{1.8cm}|m{3.2cm}|}
		\hline
		\centering \textbf{Communication Cost Components} & \textbf{Mathematical Focus} & \textbf{Considerations} \\
		\hline
		\centering $\left | \left | \theta_i - \theta_j \right |\right |$ & Distance in Euclidean space & Measures direct dissimilarity between model parameters. Sensitive to magnitude differences. \\
		\hline
		\centering $(\left | \left | \theta_i \right |\right |_2 - \left | \left | \theta_j \right |\right |_2)^2$ & Squared difference of Euclidean norms & Emphasizes the squared magnitude difference, providing a balance between sensitivity and robustness to extreme values. \\
		\hline
		\centering $\text{KL}(\theta_i || \theta_j)$ & Information-theoretic dissimilarity & Measures the information lost when using the model parameters from device $(j)$ to represent the distribution of device $(i)$. Sensitive to differences in distributions. \\
		\hline
		\centering $\text{JSD}(\theta_i, \theta_j)$ & Balanced divergence measure & Provides a symmetric and smoothed version of KL divergence, offering a balanced measure of dissimilarity. Sensitive to both local and global updates. \\
		\hline
	\end{tabular}
\end{table}

The formulas summarized in Table \ref{Table:FL_Smart_Cities_Communication} offer diverse perspectives on dissimilarity, providing a comprehensive toolkit for optimizing communication dynamics. Efforts towards bandwidth optimization, inherent in the FL framework within Smart City CPS, are encapsulated in these refined cost functions. Strategies involving edge computing integration and decentralized model updates contribute to the efficient utilization of resources, addressing the challenges posed by the dynamic urban environment. In essence, this array of communication cost formulas adds a layer of precision to the mathematical discourse of FL in Smart City-centric CPS. They embody the intricacies of collaborative learning, ensuring that the communication and networking framework is not only efficient but also aligned with the unique dynamics of Smart Cities.

\subsubsection{Implementation Approaches}
\paragraph{Waste Management Systems}
Waste management is considered to be a component of CPS-connected smart city environmental control systems because improper waste disposal releases harmful gases that accelerate global warming and deplete water supplies. Any disruption might have catastrophic effects for this essential resource. It is anticipated that Delhi's landfills would become monstrous than the Taj Mahal due to the city's increasing waste generation \cite{ghosh2019assessment}.

As FL can let multiple devices or systems work together to train a model without sharing their data, thus it can be used to increase the accuracy and efficiency of waste classification in CPS, which integrates physical and digital components. Utilizing federated learning for waste management in CPS has various potential benefits. For example, it can get rid of the need to store and process data in one place, which improves privacy and security. FL also facilitates the combination of data from diverse sources, e.g., sensors, cameras, and other devices, potentially leading to more accurate and efficient waste classification. For example, the author of \cite{vogiatzis2021dual} suggests sorting photos of recyclable items using Dual-branch Multi-output CNN to solve the waste sorting problems mentioned above. This CNN is created specifically for recycling and plastic identification. In the suggested architecture, two classifiers, each learned on a varied dataset, retain information that builds upon itself on recyclable materials. In particular, the VGG16, RestNet50, and Densenet121 designs are used, along with improving the data in the Trashnet dataset and changing the way the data looks in the WaDaBa dataset. This approach can identify previously unknown label permutations by mixing data sets. The authors' findings demonstrate that this approach can efficiently separate waste. To address privacy concerns in waste management using the FL architecture, Quan et al. in \cite{quan2024towards} propose a privacy-preserving framework for waste classification on resource-constrained devices. The proposed framework enables accurate classification while protecting user data in smart waste management systems. Consider the work of \cite{ahmed2020active}, in which a FL framework for training models for two unique application domains utilizing unlabeled data samples from each user is offered, to further illustrate how FL may be utilized in the development of CPS-connected waste management systems. Using active learning (AL) techniques, the authors demonstrated the use of FL in their research on natural catastrophes and waste sorting. In particular, sensor-based actuators from different institutions (like schools and hospitals) update local models and make predictions based on real-time photos of how users throw away their trash. Lastly, it would send the updated local model to FL aggregators so that they could change the global model's weights. Similarly, SVM is integrated with Least Square (LS), DL-based segmentation (DLS), Wavelet Theory (WT), and a FL-based solution in \cite{ayeleru2021forecasting} \cite{golbaz2019comparative} to predict hospital and municipal waste creation in CPS-related smart cities. Using CNNs and Artificial Neural Fuzzy Inference Systems (ANFIS) and Long Short-Term Memories (LSTMs) with Genetic Algorithms (GAs), the authors from \cite{coskuner2021application} and \cite{soni2019forecasting} predicted municipal solid waste production.

\paragraph{Smart Grid Integration}
\begin{figure*}[htbp]
\centerline{\includegraphics[width=0.95\linewidth]{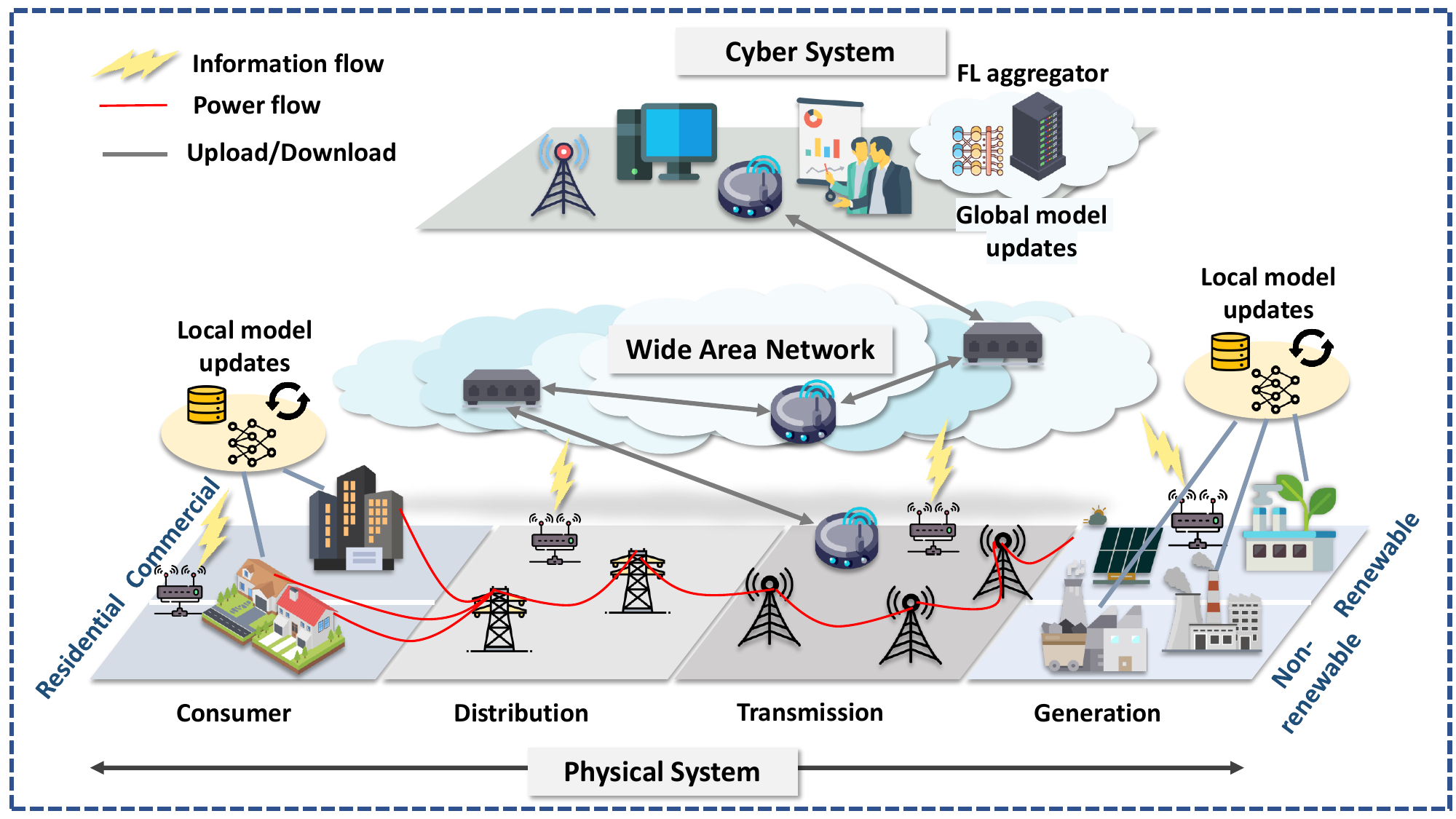}}
\caption{The application of FL for typical SG in CPS.}
\label{fig10}
\end{figure*}
CPS incorporate interaction mechanisms into their physical and digital systems. The cyber system collects data from the physical devices, analyzes it, and then employs a variety of complex algorithms to drive effective real-time control actions. The SG control and monitoring system is an example of a system that employs numerous physical devices and intermediary communication networks. Smart metering, monitoring, control, renewable energy, and other future need are not incorporated into the conventional power grid. SG transforms the grid into a flexible grid by integrating generation, transmission, distribution, consumer, operation, and markets, among other components, to monitor and control its health in real time. Typical cyber-physical system-based SG is illustrated by Fig. \ref{fig10}. Additionally, SG differs from the usual application pyramid based on automation from a CPS perspective \cite{suvarna2021cyber} since SG-CPS emphasizes decentralization, unlike hierarchical automation programs \cite{lyu2019towards}. 

From a pragmatic aspect, little is known about the application of DL in SG, especially FL. In their FL implementations for EV-based energy demand prediction and electric load forecast, both \cite{taik2020electrical} and \cite{saputra2019energy} solely prioritize horizontal data separation and lack a systematic encryption strategy. Power trace leakage will expose system flaws, posing a threat to the local economy and national security, highlighting the significance of safeguarding the privacy of SG data. However, the investigation of \cite{liu2021federated} provides the SG with a FL-based system that may enhance the safety of power lines through the cooperative learning of many parties. This study proposes a hierarchical federated learning sub-framework for multi-area collaborative power consumption forecasting that accounts for communication delays, as well as two decentrailzed federated learning sub-frameworks that allow for collaborative power consumption predictions based on datasets from two parties. Vertically federated XGBoost (or SecureBoost) improves predictions, whereas vertical linear regression safeguards data with a third-party that can be trusted. HFL and VFL sub-frameworks safeguard data and confidentiality.

Furthermore, FedDetect in \cite{wen2021feddetect}, a FL framework for energy theft detection in CPS-related SG that protects consumer privacy, enables consumers to utilize their energy usage data while retaining their privacy. This architecture takes into account a FL system with many physical detection stations, a control center (CC), and a data center (DC). Each embedded sensor has access exclusively to local consumer data, which is protected by a local DP scheme. Through our secure protocol, integrated sensors provide encrypted training parameters to the CC and DC. The CC and DC generate aggregated parameters through homomorphic encryption and return updated model parameters to embedded physical views for model training.
\paragraph{Urban Data Management}
Depending on the available resources (processing speed, power, etc.) and the volume of data that must be managed, CPS has various data management needs. While certain CPS can send computationally complex tasks to data centers and receive responses in near real-time, others require on-board data analysis. CPS data faces four obstacles: cognitive biases, data quality, privacy and security, and the selection of a data management system that closely meets CPS's functional and non-functional criteria.

Therefore, FL has been adopted as a means of providing distributed AI capabilities for smart cities' extensive intelligent data management systems because of its decentralized and privacy-enhancing features. The use of FL in smart cities is exemplified in the introduction of FedSem, a semi-supervised FL approach designed to distribute the processing of unlabeled data, as presented in \cite{albaseer2020exploiting}. The effectiveness of FL is evaluated by utilizing a prototype of a smart car, where each vehicle learned a DNN model relied on the traffic sign image datasets. A centralized server oversees the selection of a number of participants for each learning round in order to coordinate the models. To replicate the FL method, a dataset of German traffic signs containing between one thousand and one hundred ten cars is employed, with thirty cars selected at random for training and updating the model throughout each iteration. Simultaneously, \cite{chiu2020semisupervised} suggests the use of FL to establish a platform for video data management that can be used in smart urban. Buses and other connected mobile physical actuators can capture live imagery of the streets and send it to devices on the network's periphery. Each computing sensor then uses the recovered video segments to run a semi-supervised learning algorithm, which may carry out local video analytics on the gathered data. The authors propose a FedSwap operation to solve the issue of non-IID data, with the predicted outcome of reducing variability within the data and, as proved through simulations, boosting the accuracy of picture categorization by 3.8\%, which represents a significant improvement considering the challenges posed by non-IID data distribution in this smart city application. Moreover, \cite{mukhametov2020ubiquitous} addresses the heterogeneity of data in smart cities across different actuators and streams as well as user privacy concerns by utilizing FL to organize streams of data from pervasive interactive sensing devices that function as FL clients and engage in local training without disclosing the data to outsiders. Existing smart city patterns would be greatly modified by the launch of novel features e.g., social activity monitoring, social economic sharing, global people connection \cite{liu2020fedvision}, and smart urban communication. Intelligent endpoints that are sensor-based and can learn about their environment and apply this knowledge to better serve their consumers \cite{guo2015integration}.
\paragraph{Smart Parking Solutions}
CPS technology is required for the successful implementation of a cutting-edge and comprehensive smart parking system. CPS can be used for a variety of purposes, including traffic management \cite{marwedel2021embedded}. The data is processed by the cyber component once it has been collected by the physical component's sensors and actuators \cite{navickas2017cyber}. Both digital and physical systems are merged under the CPS umbrella. In addtion, CPS applications based on Edge Computing include intelligent, networked autos and parking systems. Autonomous vehicles are outfitted with numerous high-tech features and services, including gesture controls, voice and speech recognition, navigation and safety systems, eye tracking, and more. Smart parking, collision avoidance, and traffic control are examples of collaborative and scalable services enabled by Vehicle-to-Vehicle (V2V) and Vehicle-to-Infrastructure (V2I) connections with Roadside Units (RUs) and adjacent cars \cite{zhao2016study}. 

\begin{figure}[htbp]
\centerline{\includegraphics[width=0.99\linewidth]{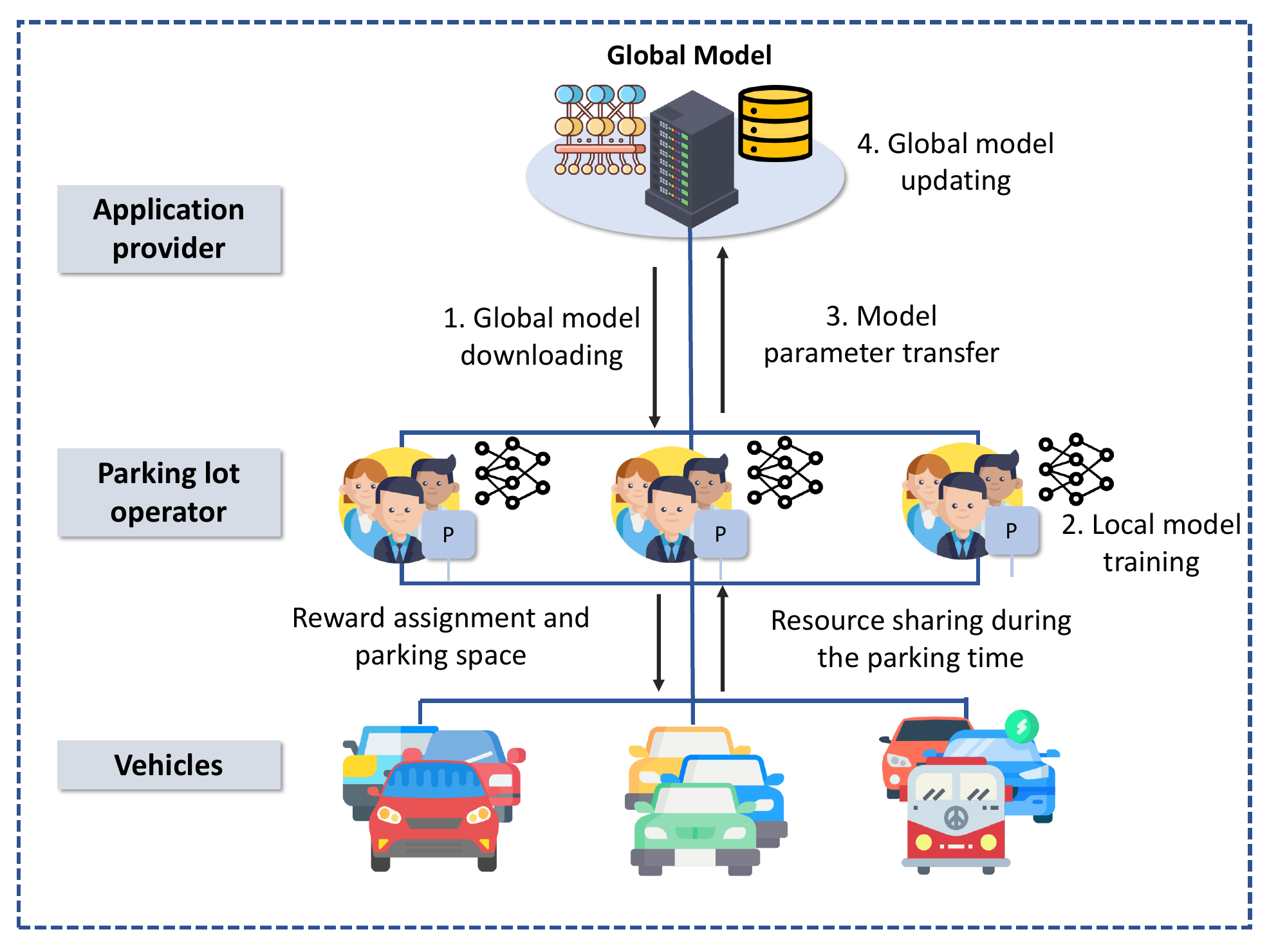}}
\caption{A FL-based parking space estimation scheme.}
\label{fig8}
\end{figure}

Recent research efforts have aimed to improve our understanding of predicting traffic information, e.g., speed \cite{zhang2021fastgnn}, while also ensuring data privacy protection during the prediction process and traffic flow \cite{liu2020privacy}. This work has inspired the extension of FL applications in the way how to manage parking in CPS-related smart cities, leading to the development of FedParking, an FL-based parking space estimation scheme as demonstrated in Fig. \ref{fig8}, as proposed in \cite{huang2021fedparking}. FedParking is created to investigate FL-based parking space estimation using Parked Vehicle-assisted Edge Computing (PVEC). Parking Lot Operators (PLOs) are required to train a single LSTM model for anticipating parking space availability in the absence of sharing raw data. Hence, each PLO is issued a parking space restriction and tasked with devising techniques to motivate automobiles to fill parking spaces, as well as pooling idle computing resources to deliver services in PVEC without exchanging raw data. According to \cite{lim2020multi}, a method is presented in which FL is utilized for UAVs in parking management. The authors propose \cite{pham2021uav} employing UAV-enabled wireless power transfer to ensure the long-term viability of FL-based wireless networks in order to overcome the problem of limited battery life in small physical sensors. FL performance suffers when communication links are broken or nodes are unavailable due to the ongoing rebalancing of model parameters.

\subsection{Vehicular Cyber Physical Systems (VCPS)} 
 In the context of Vehicular Cyber Physical Systems (VCPS), FL could be used to train a model that enhances the functionality of VCPS. A VCPS could, for instance, employ FL to train a model that can forecast traffic patterns and improve the efficiency of vehicle routing, or it may use FL to train a model that can predict maintenance needs and raise the VCPS's reliability. In particular, the sections that follow give a thorough examination of FL's use in intelligent transportation systems (ITS), vehicular networking, vehicular resource management, and vehicular traffic planning. Furthermore, we introduce Table \ref{Table:FL_VCPS_Applications} to present a brief yet comprehensive summary of the principal merits and demerits of incorporating FL in the context of VCPS solutions.

\begin{table*}[h!t]
	\centering
	\caption{Taxonomy of FL-based Solutions for Smart Vehicular CPS.}
	\color{black}
	\label{Table:FL_VCPS_Applications}
    \renewcommand{\arraystretch}{1.1}
    
		\begin{tabular}{|m{2.1cm}|m{0.7cm}|m{1.7cm}|m{2.2cm}|m{1.8cm}|m{3.2cm}|m{3.2cm}|}
			\hline
			\textbf{Applied Domain}& 
        \textbf{Ref.} &
			\textbf{System Architecture Type} &
			\textbf{Data Characteristics}& 	
			\textbf{Learning Paradigm}&
			\textbf{Key Contributions}&
			\textbf{Limitations}
			\\
			\hline
      \multirow{6.4}{*}{\hspace{2.7mm} \begin{tabular}[c]{@{}c@{}}Intelligent\\ Transportation \\ Systems \end{tabular}} 
      & \centering{\cite{cebe2018block4forensic}} 
      & Decentralized (Blockchain) 
      & Non-IID forensic data; Multi-source 
      & Hybrid (Consensus + FL) 
      & VPKI-integrated blockchain for vehicular forensics (92\% detection accuracy) 
      & Lacks penalty mechanisms for malicious nodes
			\\ \cline{2-7} 
      & \centering{\cite{yuan2021fedrd}} 
      & Centralized 
      & Road damage images; High resolution 
      & Model-Centric (Adaptive CNN) 
      & Real-time road damage detection (MAE: 4.2cm) with dynamic model updates 
      & High computation load limits real-time deployment
			\\ \cline{2-7} 
      & \centering{\cite{wilbur2020time}} 
      & Decentralized (Fog) 
      & Real-time traffic patterns; Temporal 
      & Data-Centric (LSTM) 
      & Privacy-preserving fog network routing (23ms latency) 
      & 22\% higher communication costs vs centralized FL
         \\ \cline{2-7}
			\hline
      \multirow{6.8}{*}{\hspace{3mm} \begin{tabular}[c]{@{}c@{}}Vehicular\\ Networking \end{tabular}} 
      & \centering{\cite{pokhrel2020decentralized}} 
      & Decentralized (Blockchain) 
      & Multi-modal V2X data 
      & Hybrid (BFL + RL) 
      & Configurable BFL parameters for VANETs (94.7\% message integrity) 
      & No formal convergence guarantees
			\\ \cline{2-7} 
      & \centering{\cite{aloqaily2021energy}} 
      & Decentralized (P2P) 
      & Energy consumption patterns 
      & Model-Centric (DNN) 
      & Energy-aware BFL for EV networks (18\% energy savings) 
      & Vulnerable to Sybil attacks
			\\ \cline{2-7} 
      & \centering{\cite{barbieri2022decentralized}} 
      & Decentralized (C-FL) 
      & V2X sensor streams; Non-IID 
      & Hybrid (GRU + FedAvg) 
      & 6G-compatible C-FL for vehicular edge computing 
      & Untested with mmWave beamforming
         \\ \cline{2-7}
			\hline
      \multirow{7.5}{*}{\hspace{2.7mm} \begin{tabular}[c]{@{}c@{}}Vehicular\\ Resource \\ Management \end{tabular}} 
      & \centering{\cite{xiao2021vehicle}} 
      & Centralized 
      & Camera/LiDAR streams; High velocity 
      & Data-Centric (Greedy Selection) 
      & On-the-fly vehicle selection for FL tasks (89\% image recognition accuracy) 
      & Exposes model gradients to MITM attacks
			\\ \cline{2-7} 
      & \centering{\cite{prathiba2021federated}} 
      & Centralized 
      & Real-time telemetry data 
      & Model-Centric (SVM) 
      & FL-enabled computation offloading (73\% latency reduction) 
      & Lacks real-world validation
			\\ \cline{2-7} 
      & \centering{\cite{yang2021privacy}} 
      & Centralized 
      & UAV swarm trajectories 
      & Hybrid (AFL + DP) 
      & Privacy-preserved multi-UAV coordination ($\epsilon$=1.5 DP) 
      & 8\% accuracy drop under DP constraints
         \\ \cline{2-7}
			\hline
      \multirow{6.8}{*}{\hspace{4.8mm} \begin{tabular}[c]{@{}c@{}}Vehicular\\ Traffic \\ Planning \end{tabular}} 
      & \centering{\cite{liu2020privacy}} 
      & Centralized 
      & Spatiotemporal traffic data 
      & Model-Centric (GRU) 
      & Encrypted FL for flow prediction (RMSE: 12.4 vehicles/min) 
      & Ignores road network topology
			\\ \cline{2-7} 
      & \centering{\cite{lim2021towards}} 
      & Hybrid (Edge-Cloud) 
      & Multi-modal IoT data 
      & Model-Centric (DNN) 
      & FL-based ITS for smart cities (92.3\% congestion accuracy) 
      & No resource optimization framework
			\\ \cline{2-7} 
      & \centering{\cite{meese2022bfrt}} 
      & Decentralized (Blockchain) 
      & Real-time traffic feeds 
      & Data-Centric (LSTM) 
      & Byzantine-robust traffic prediction (F1=0.89) 
      & High PoW energy consumption (18-22\%)
			\\ \cline{2-7}
			\hline
	\end{tabular}
\end{table*}

\subsubsection{Communication and Networking Architecture in FL-VCPS}
To examine the integration of FL in Vehicular VCPS, this section will closely analyze the mechanics of communication. Our focus extends beyond managing vehicle dynamics; we are fine-tuning several aspects to ensure seamless execution of FL activities. We will explore the mechanisms by which vehicles communicate with one other (V2V), establish connections with infrastructure (V2I), and manage communication across different domains. It involves enhancing communication to meet the specific requirements of FL jobs in the fast-paced environment of VCPS. Prepare yourself for a straightforward and comprehensive exploration of FL in VCPS communication.

\paragraph{Communication Architecture}
For the communication architecture (\(CA\)), consider incorporating FL aspects into the dynamic weighting function (\(\beta(t)\)):
\begin{align*}
    CA(t) &= \beta(t) \cdot CA_{centralized} \\
    &\quad + (1 - \beta(t)) \cdot CA_{decentralized} + FL_{adaptation},
\end{align*}
where \(FL_{adaptation}\) represents an adaptive term based on the progress and requirements of FL tasks. This addition ensures that the communication architecture adapts not only to vehicular dynamics but also to the evolving demands of FL model training.

\paragraph{Communication Protocols}
Enhance the V2V communication model by integrating FL-related parameters. Consider a FL connectivity probability (\(P_{FL}\)) that is influenced by the availability of vehicles participating in FL tasks:
\begin{align*}
    P_{V2V,FL}(d, t) = P_{V2V}(d, t) \cdot P_{FL}(t),
\end{align*}
where \(P_{FL}(t)\) accounts for the probability of vehicles being available for FL collaboration, adding a FL dimension to the connectivity model. In V2I communication, refine the reliability model by considering the impact of FL-related data transmissions:
\begin{align*}
    R_{V2I,FL}(t) = R_{V2I}(t) + \delta \cdot FL_{data}(t).
\end{align*}
This model introduces a term (\(\delta \cdot FL_{data}(t)\)) representing the influence of FL-related data on communication reliability.

\paragraph{Cross-Domain Communication}
Extend the utility function for cross-domain communication to reflect FL-related priorities:

\begin{align*}
    U_{FL}(T_{ij}, \rho_{ij}, FL_{priority}) &= \frac{1}{T_{ij} + \epsilon} \cdot \left(1 + \frac{\rho_{ij}}{\rho_{\text{max}}}\right)^{-\delta} \\
    &\quad + FL_{priority},
\end{align*}
where \(FL_{priority}\) represents the priority assigned to FL-related communication tasks. This augmented utility function balances the optimization objectives between traditional traffic concerns and FL-related data exchanges.

By articulating these FL-specific details into the communication framework, you ensure that the architecture, protocols, and cross-domain communication are not only adaptive to vehicular dynamics but also strategically align with the requirements and nuances of FL in CPS.

\subsubsection{Implementation Approaches}

\paragraph{Vehicular Traffic Planning} \label{VehicularTraffic}
Within the broader context of ITS, vehicular traffic planning plays a crucial role in optimizing the movement and flow of vehicles within transportation networks. This section focuses on the specific applications of FL in addressing challenges related to traffic flow prediction, route optimization, and other traffic management aspects. In the real world, CPS vehicular traffic planning employs computational and physical resources to optimize the movement and flow of automobiles inside transportation networks. Utilizing CPS technology, e.g., sensors, GPS, wireless connection, and ML, it is possible to collect and analyze real-time traffic data in order to improve traffic flow and reduce congestion. In this context, FL can be used to train ML models that optimize traffic flow and enhance transportation efficiency. Multiple stakeholders, e.g., transportation authorities, could collaborate to train a federated ML model for traffic forecasting and network bottleneck detection. The model might then be utilized to optimize the timing of traffic signals and routing decisions in order to enhance traffic flow (e.g., \cite{zeng2021multi}). These models can be deployed on individual vehicles or traffic management systems to make traffic optimization decisions based on the data and goals of each party (e.g., \cite{yang2021brainiot}). FL can also be used to train models that can be deployed on physical edge sensors, e.g., roadside units (RSUs), to improve the efficiency and effectiveness of traffic planning in VCPS (e.g., \cite{niknam2020federated}).

\begin{figure*}[htbp]
\centerline{\includegraphics[width=0.97\linewidth]{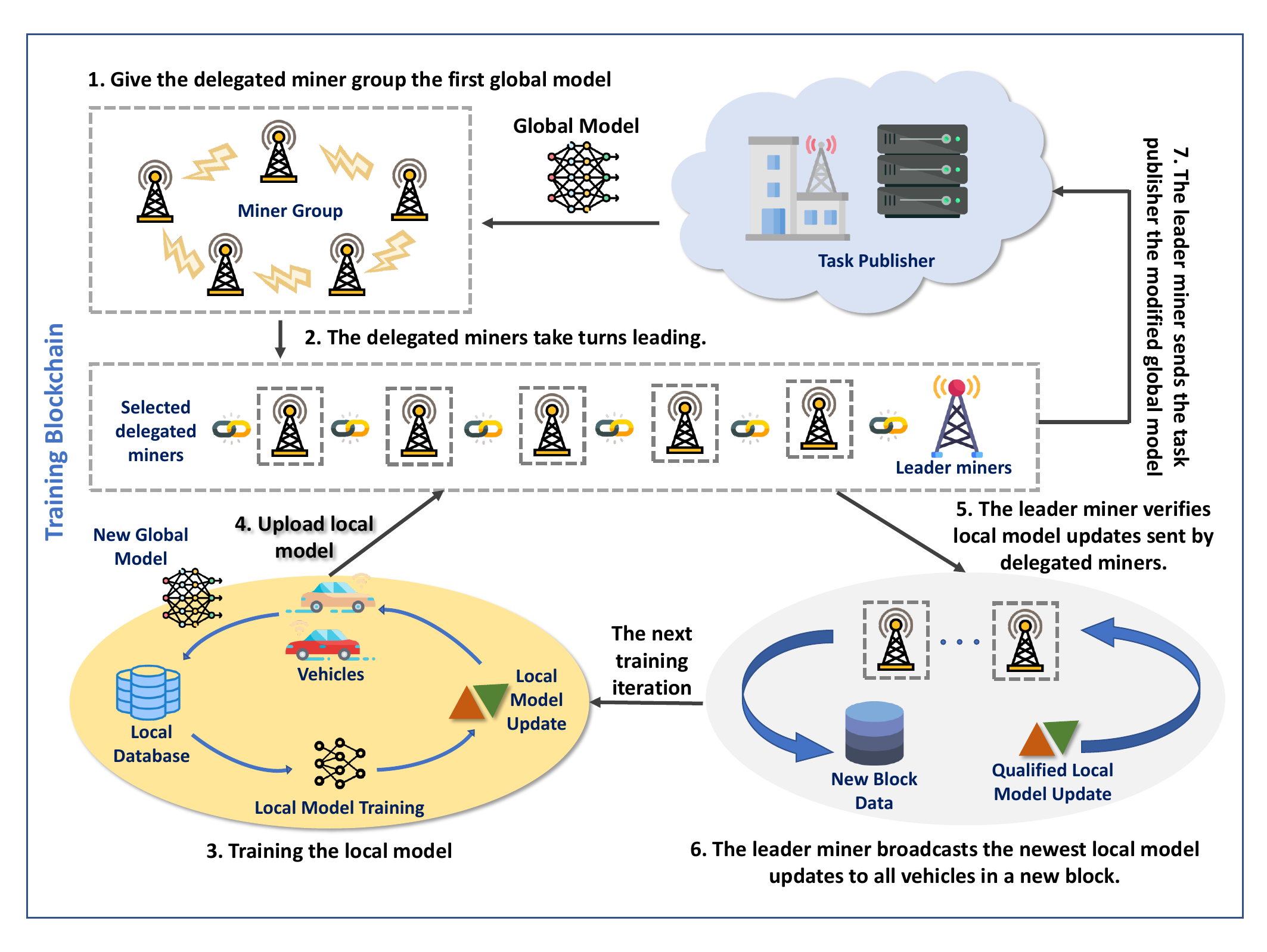}}
\caption{Blockchain-based FL for vehicular CPS.}
\label{fig6}
\end{figure*}

Additionally, the authors of \cite{liu2020privacy} describe FL, a ML approach to Traffic Flow Prediction (TFP) that protects privacy in CPS. DL models have demonstrated potential in TFP, but their reliance on enormous data sets that may contain personally identifiable information creates privacy concerns. This issue can be resolved by FL, which updates universal learning models via secure parameter aggregation rather than raw input. For scaling TFP, a gated recurrent unit NN algorithm (FedGRU) and joint announcement are suggested. Before FedGRU, ensemble clustering is utilized to classify businesses by TFP. Privacy concerns enhance the utilization of FL in the VCPS for data gathering as well as the model training process. The research \cite{lim2021towards} proposes utilizing FL to enable collaborative ML while safeguarding the privacy of independent drones-as-a-service (DaaS) providers in order to develop Internet of Vehicle (IoV) applications. Using a multidimensional contract and the Gale-Shapley algorithm, they link the most cost-effective interacting components with each subregion to alleviate information asymmetries and incentive mismatches between drone owners and model owners. Based on simulation results, this technique is incentive-compatible and efficient, maximizing the model owner's revenues. Furthermore, the authors of \cite{meese2022bfrt} offers a blockchain-based FL method for predicting online traffic flow using real-time data and edge computing to safeguard privacy in the context of VCPS. There is a standard procedure for applying BFL, which consists of the seven phases shown in Fig. \ref{fig6}. Initial global model is delivered to a delegated set of miners. During this time, the intelligent cars train their local models with the global model and send them to the leader miner for validation. The leader miner then verifies the changes transmitted by delegated miners and, if they are genuine, distributes the most recent local model upgrades to all vehicles in a new block. The leader miner then transmits the modified global model to the task publisher, guaranteeing that all CPS vehicles have access to the most current information. In details, implemented on Hyperledger Fabric, the suggested solution employs federated GRU and LSTM models and enables decentralized model training at embedded computing controls equipped in cars. Experiments with shards of dynamically obtained arterial traffic data indicate that the proposed method beats centralized models and is suited for decentralized, real-time traffic flow prediction that protects privacy.

\paragraph{Intelligent Transportation Systems (ITS)}
FL could be utilized as a training strategy for ML models that improve transportation system efficiency and safety in the context of ITS in CPS, e.g., by predicting traffic patterns \ref{VehicularTraffic}, predicting likely accidents \cite{ferdowsi2019deep}, and optimizing routing algorithms \cite{posner2021federated}. Using FL, ML models could be built from data collected on several devices without requiring centralizing that information. This ensures the anonymity of individuals whose data is being examined for inclusion in the model and facilitates the development of more reliable predictive models by leveraging the collective expertise of all system users. 

In reality, without the sensors embedded into modern automobiles, self-driving vehicles would not exist. The data from these vehicles is utilized by a wide range of individuals, but forensic investigation of traffic crashes is an essential application. The Block4Forensic architecture in \cite{cebe2018block4forensic} provides a blockchain-based FL framework with vehicular public key infrastructure and a fragmented ledger for keeping and analyzing vehicle-related data for trustless, traceable, and private post-accident analysis. This framework may also be beneficial when investigating and evaluating car accidents. Road degradation detection and warning systems are crucial for safety. Cloud-based data processing slows down user alerts. Computing close to the network's edge reduces latency but limits communication range and may compromise privacy. FedRD in \cite{yuan2021fedrd} diagnoses road deterioration through edge-cloud computing and FL. Through hierarchical feature fusion, adaptive FL, and pixelization, it safeguards privacy. Simulations demonstrate that FedRD can detect potentially dangerous road deterioration and provide timely, accurate warnings without compromising privacy, even when some road edges have little data.

Furthermore, cloud computing has enabled the deployment of data-intensive smart city applications, e.g., Google Maps, that use shared memory models and can only be deployed in data centers. Therefore, the authors of \cite{wilbur2020time} present a decentralized route planning technique for private fog networks that leverages FL and CPS to improve service speed and reliability by enabling real-time data processing and decision-making at the network's edge as opposed to central data centers. A simulation of a community of average size in the United States is used to test the technique. Another example for optimizing routing algorithms in ITS, the research of \cite{dinh2022network} presents a decentralized edge network architecture for FL that cuts training delays and cloud traffic and processing overhead by a significant amount. The in-network computation framework (INC) consists of a user scheduling system, an in-network aggregation procedure, and a network routing algorithm capable of reducing FL training latency by up to 5.6 times. Instead of depending on centralized data centers, the INC architecture can be connected with CPS to enable real-time data processing and decision-making at the network's edge. This can enhance the performance and dependability of the FL service and simplify the creation of smart city applications and other data-intensive systems. While FL has demonstrated its potential in various ITS applications, its specific role in optimizing vehicular traffic planning warrants further exploration. The Section \ref{VehicularTraffic} will delve into the application of FL techniques to enhance traffic flow, reduce congestion, and improve overall transportation efficiency.

\paragraph{Vehicular Networking }
Vehicular networking in CPS is a critical enabler of new transportation services and technologies, having the ability to dramatically enhance the performance and efficiency of transportation systems, whereas FL has the potential to significantly enhance the performance and efficiency of vehicular networking in CPS. Vehicular networking encompasses communication between vehicles, V2I, and vehicles and other devices (vehicle-to-everything communication, or V2X), and FL can be used with these protocols to give vehicles the ability to make smart decisions about how to optimize different parts of the communication network. This makes the network safer, more efficient, and uses less energy.

Autonomous BFL designs for efficient and privacy-aware vehicle communication networks, for example, are proposed by \cite{pokhrel2020decentralized}. This demonstrates that blockchain consensus can allow decentralized interchange and verification of local on-vehicle ML model upgrades in the context of VCPS. A mathematical method for assessing network and BFL parameter effects on system performance is also offered. They calculate the end-to-end latency with BFL and use this knowledge to determine the best block arrival rate, taking communication and consensus delays into account, by studying the dynamics of the on-vehicle ML system. Since blockchain technology is used to decentralize the provisioning and control components, and to maintain authenticity and integrity, the BFL approach is also used at the edge to guarantee accurate and up-to-date service provisioning depending on the surrounding environment and network restrictions in \cite{aloqaily2021energy}. Consensus-driven FL (C-FL) is a comparable approach that is built and studied on a realistic V2X network in \cite{barbieri2022decentralized}. Cars outfitted with C-FL may coordinate their activities by sharing NN parameters. To facilitate the building of V2X connections, route information is gathered using a basic traffic simulator. The paper examines the influence of V2X connectivity on the FL's consensus process, convergence time, and loss/accuracy tradeoff is then examined, and the FL's performance under various traffic densities is evaluated. Additionally, the authors of \cite{pham2021uav} present drone-based wireless device recharging as a technique of assuring the long-term viability of the application FL in vehicular networking. By applying the iterative UAV-SFL approach and making modifications to the transmission time, power, bandwidth, and position, they are able to improve the UAV's power efficiency. The simulation results demonstrate that the designed UAV-powered FL wireless network is not only better but also more sustainable.

\paragraph{Vehicular Resource Management}
Due to urbanization and rising prosperity, the number of vehicles on the road has increased. With the advancement of technology, the prevalence of CPS-based autonomous vehicles is increasing. Vehicular Edge Computing (VEC) is developed as a solution to the resource constraints that cars typically face when trying cognitive tasks. However, the nature of automotive activity may periodically impede the delivery of information. FL can help reduce resource latency by reducing trip times and fuel consumption, hence helping to the enhancement of transportation system effectiveness. However, the capabilities and data quality of each vehicle also vary, which can impact the performance of model training. 

Thus, the authors in \cite{xiao2021vehicle} present a min-max optimization problem to optimize on-board compute capabilities, transmission power, and local model precision in order to minimize FL costs in the worst-case scenario. In addition, they present a greedy algorithm for recognizing vehicles with good image quality and a heuristic search technique for optimization. The simulation results demonstrate that the given algorithms are efficient and strike a balance between fairness and cost. Optimization of nonlinear programming consists of two subproblems. The Lagrangian dual problem, subgradient projection technique, and adaptive harmony algorithm for heuristic search handle the challenges of resource allocation and local model precision, respectively. FLOR (FL enabled computation offloading and resource management) is proposed in \cite{prathiba2021federated} for low-latency offloading and resource management in CPS. The FLOR framework efficiently manages CPS resources across heterogeneous networks with a focus on V2X, DSRC (dedicated short-range communications), 5G mmWave, and 6G V2V. The stochastic network calculus computes the maximum delay for heterogeneous communication and the probability of offloading computation workloads to the optimal network. To enhance CPS resource allocation and usage and reduce resource cost increases, FLOR offers a federated Q-learning framework, called e-RRM (effective Radio Resource Management). Simulations indicate that FLOR offloads computation between heterogeneous networks with efficient resource allocation 20.69 percent better than CPS alternatives.

In addition, CPS sense, process, and act in the physical world using computing, communication, and physical components. CPS physical components are integrated in devices or systems that interact with the environment, e.g., UAVs or autonomous cars. The CPS uses computational and communication components to process and send sensor readings and control signals and make decisions. The paper \cite{yang2021privacy} researches multi-UAV networks, a sort of CPS, to increase data exchange reliability and efficiency. Asynchronous federated learning (AFL) trains local models without sending sensitive data to UAV server sites, solving privacy issues. Simulations show that the A3C-based architecture and algorithm for device selection and resource management improves federated convergence speed and precision. Another VCPS example, authors from \cite{lu2019differentially} suggests using differentially private AFL (DPAFL) for safe and private resource sharing in automobile networks. Local DP, FL, and a random distributed update approach secure updated local models from a centralized curator, while update verification and weighted aggregation increase convergence. Real-world datasets demonstrate DPAFL's accuracy, efficiency, and data secrecy.

\subsection{Core Cybersecurity Services}
\subsubsection{Attack Detection}
The integration of sophisticated networking and computing technologies, e.g., 5G/6G and AI, with traditional industrial infrastructures has greatly increased the cyberattack susceptibility of industrial CPS. Due to the lack of detailed instances of assaults, it has been challenging to protect these huge, complex, and diversified CPS from these risks. FL-based CPS is also susceptible to the emerging threats common to beyong-5G wireless systems \cite{ruzomberka2023challenges}. Thus, FL can be used to train a model for use in the field of CPS attack detection to assist spot abnormalities or malicious system behavior. The model is trained utilizing data from several devices or systems inside the CPS without needing the data to be centralized or shared. To avoid the compromise of sensitive data, it is vital to protect the privacy of individual devices and systems.

In real-world cases, CPS are subject to False Data Injection Attacks (FDIAs) because they create enormous amounts of diverse data. In dynamic and dispersed scenarios, e.g., CPS-based energy systems, centralized FDIA detection algorithms risk compromising data privacy and becoming useless. The paper \cite{tahir2021experience} proposes a distributed FDIA detection method with deep FL and careful aggregation as a solution. All system nodes can simultaneously and precisely identify stealthy FDIAs, and system clients can train a shared model in secret. In a distributed setting, the suggested approach surpasses the state-of-the-art in terms of detection precision, computational complexity, and data privacy. 
FL has the potential to enhance the security of CPS data by mitigating adversarial attacks, e.g., data poisoning attempts that may lead to CPS malfunctions. The choice of hyperparameters is equally important in constructing a successful FL model that can endure adversarial assaults. In response to adversarial attacks, the authors of \cite{yamany2021oqfl} describe the optimized Quantum-based federated learning (QFL) framework, which employs quantum-behaved optimization of particle swarms to fine-tune the parameters of autonomous cars. This new technique can combat malicious attacks when integrating FL in CPS in a cyber security architecture.

Additionally, the study \cite{abdel2022privacy} offers a blockchain-based edge intelligence (BoEI) architecture with Fed-Trust decentralized FL to detect cyberattacks in CPS. Fed-Trust performs semi-supervised data learning using semi-labeled data using a temporal convolutional generating network. BoEI uses a reputation-based blockchain to record and verify transactions in order to protect data privacy. By transferring block mining to actuators, fog computing enhances communication and Fed-processing trust. Simulations utilizing two public datasets demonstrate Fed-superiority Trust's over current attack tactics. It is crucial to safeguard CPS communication services from infiltration. Current methods for safeguarding Points Of Interest (POI) microservices typically include anonymity and DP technologies despite their drawbacks. The study \cite{guo2022deep} presents a FL-based strategy for securing POI microservices in CPS. The system design employs a training mechanism between the cloud data center and the user data on edge nodes. Edge nodes supply users with DL models, while the cloud server processes FL modifications to these models. The isolation and connectedness between the cloud's perimeter and its center enable and enhance FL and data security, respectively.

\subsubsection{Authentication Systems}
FL is an effective method for validating the identity of users or devices in CPS. In CPS, there may be a large number of devices or systems that need to be authenticated, and centralizing data storage or sharing may not be feasible due to concerns about privacy or security. FL allows for the development of a global authentication model that is trained using data from multiple devices or systems, while still protecting their privacy and independence. This can enhance the accuracy and strength of the authentication process in CPS, as well as making it more flexible and able to adapt to evolving needs or requirements.

The increasing prevalence of drones in CPS, for example, inspired the authors of \cite{yazdinejad2021federated} to evaluate the significance of establishing flawless identification systems for these devices. In order to comprehend the distinctive characteristics of the signals emitted by each of these drones, the authors develop an authentication model for FL drones. This method enables the system to validate the authenticity of the multiple drones transmitting RF signals. HE and safe aggregation are incorporated into this model, which is optimized and executed locally on the drones using SGD. The experimental findings demonstrate that the suggested approach is more effective than previous techniques for authenticating drones using ML. Another application of FL for authentication in CPS is discussed in \cite{zhang2021federated}. This work presents a FL-based approach for identifying problems with rolling bearings, a common component of rotating machinery. There are several reasons why these bearings may be troublesome. The proposed method achieves its objective of making the global model more resistant to faulty inputs by combining a dynamic authentication mechanism with an unsupervised learning strategy. After examining how well the models perform on a validation set, only models with a high validation performance are aggregated using the dynamic validation approach employed by the FL framework. On two distinct datasets, the proposed method is evaluated and shown to be effective as a methodology for private decentralized learning. Additionally, a cooperative verification system that trains its model utilizing distributed edge devices, e.g., linked autos and RSUs, is presented in \cite{liu2021blockchain}. The federated design of this system reduces the amount of resources that are required by the central server while retaining both security and privacy. The aggregation approach benefits from an increased level of safety as a result of the use of blockchain technology, which is employed to store and distribute the training models. It has been demonstrated that the proposed system is capable of protecting the collective privacy of vehicles, reducing the costs of communication and computing, and withstanding common risks.

\subsubsection{Intrusion Detection}
Anomaly intrusion in CPS refers to the occurrence of abnormalities or deviations from normal CPS activity, which may be indicative of an intrusion or cyber assault. These abnormalities might be the result of a variety of circumstances, e.g., malfunctioning equipment or software, human mistake, or hostile actions. In certain instances, anomalies may be subtle or difficult to identify, and they may only manifest over time or in combination with further abnormalities. Deployment of AI-based intrusion detection systems capable of discovering and recognizing these abnormalities in a timely and precise manner is required. In light of this, recent research have focused on the efficacy of FL-based strategies, which enable model training utilizing local data without exposing it to the risk of illegal access. This can assist in mitigating or limiting the effects of cyber intrusions on the CPS and ensuring the system's continuing operation and dependability.

DeepFed is one of the best-known strategies for defending CPS against anomaly intrustion. This federated DL technique is created by the authors of \cite{li2020deepfed} in order to analyze cyber hazards in industrial CPS. DeepFed utilizes a CNN and a gated recurrent unit to create an intrusion detection model, and a FL architecture enables several CPS to assemble a complete model while maintaining user anonymity. While DeepFed trains a model, sensitive parameters are transmitted via a Paillier-based secured communication protocol. When applied to a real-world CPS dataset, it outperforms cutting-edge approaches. The study \cite{huang2022eefed} is an additional unique FL research for investigating CPS intrusion detection. This paper presents a FL technique for intrusion detection in CPS that enables several devices or systems to collaborate to increase the effectiveness of a single local model against cyber threats. The proposed technique generates a broad global detection model and allows the local detection model to be adapted to individual characteristics. This is achieved by a Federated Execution \& Evaluation Dual (EEFED) architecture. A customized updating technique and an efficient backtracking parameters replacement mechanism reduce the detrimental impact of FL on data asymmetry and non-iid distribution. Extensive testing on a network dataset reveals that the proposed strategy improves model stability and outperforms single models and state-of-the-art solutions in a variety of cyber situations.

In addition, the method for uncovering intrusive attacks in connected and FL systems is presented in \cite{preuveneers2018chained}. The solution is to implement a permissioned blockchain-based FL approach, wherein updates to an anomaly detection ML model are connected on a distributed ledger. By doing so, ML models may be audited without having to consolidate their training data, and with just a small (5-15\%) hit to FL's efficiency. The method may be used with a wider variety of NN architectures and practical contexts. For smart manufacturing firms, the authors of \cite{verma2022fldid} offer a Federated Learning-enabled Deep Intrusion Detection (FLDID) architecture for identifying cyber threats. Paillier-based encryption is utilized to protect the confidentiality of model gradients transferred between edge devices and the server in the FLDID architecture, which enables many sectors to construct a shared model for risk detection. The proposed approach, which employs a DL-based hybrid model for intrusion detection, has been demonstrated to be successful in identifying cyber threats in smart sectors by testing on a publically accessible dataset, in comparison to state-of-the-art techniques.

\begin{table*}[h!t]
	\centering
	\caption{Taxonomy of FL-based Solutions for Cybersecurity and Privacy in CPS.}
	\color{black}
	\label{Table:FL_Cybersecurity_Applications}
    \renewcommand{\arraystretch}{1.2}
    
		\begin{tabular}{|m{2.1cm}|m{0.7cm}|m{1.7cm}|m{2.2cm}|m{1.8cm}|m{3.2cm}|m{3.2cm}|}
			\hline
			\textbf{Applied Domain}& 
        \textbf{Ref.} &
			\textbf{System Architecture Type} &
			\textbf{Data Characteristics}& 	
			\textbf{Learning Paradigm}&
			\textbf{Key Contributions}&
			\textbf{Limitations}
			\\
			\hline
      \multirow{3.85}{*}{\hspace{4.8mm} \begin{tabular}[c]{@{}c@{}}Attack\\ Detection \end{tabular}} 
      & \centering{\cite{tahir2021experience}} 
      & Decentralized (P2P) 
      & Non-IID sensor data; High velocity 
      & Model-Centric (CNN) 
      & Real-time FDI attack detection (F1=0.93) with P2P validation 
      & Requires TEEs for secure aggregation
			\\ \cline{2-7} 
      & \centering{\cite{yamany2021oqfl}} 
      & Centralized 
      & Real-time vehicular CAN logs 
      & Hybrid (Adaptive Q-Learning) 
      & Dynamic hyperparameter tuning reduces false positives by 22\% 
      & 15\% accuracy drop under adversarial training
			\\ \cline{2-7}
			\hline
  
			\multirow{5}{*}{\hspace{1.8mm} \begin{tabular}[c]{@{}c@{}}Authentication \end{tabular}} 
      & \centering{\cite{yazdinejad2021federated}} 
      & Hybrid (Edge-Cloud) 
      & Multi-modal drone telemetry 
      & Model-Centric (DNN) 
      & Privacy-preserved UAV authentication (99.4\% accuracy) 
      & 18\% latency increase for encrypted gradients
			\\ \cline{2-7} 
      & \centering{\cite{zhang2021federated}} 
      & Centralized 
      & Vibration time-series; Non-IID 
      & Data-Centric (SMOTE) 
      & Anomaly detection in rotating machinery (AUC=0.89) 
      & Vulnerable to gradient inversion attacks
			\\ \cline{2-7} 
      & \centering{\cite{liu2021blockchain}} 
      & Decentralized (Blockchain) 
      & Multi-modal V2X data 
      & Hybrid (PoW Consensus) 
      & Trust-aware block verification (94.7\% integrity) 
      & 22\% higher energy consumption vs CFL
         \\ \cline{2-7}
			\hline

      \multirow{3}{*}{\hspace{5mm} \begin{tabular}[c]{@{}c@{}}Intrusion \\Detection \end{tabular}} 
      & \centering{\cite{li2020deepfed}} 
      & Centralized 
      & ICS network logs; High velocity 
      & Model-Centric (CNN-GRU) 
      & Detects 92\% of APTs in industrial CPS 
      & Requires 300ms inference latency
			\\ \cline{2-7} 
      & \centering{\cite{huang2022eefed}} 
      & Hybrid (Personalized) 
      & Heterogeneous threat vectors 
      & Hybrid (TL + FedAvg) 
      & Customized detection models (F1=0.91) 
      & 8\% accuracy drop under DP ($\epsilon$=1.5)
         \\ \cline{2-7}
			\hline
	\end{tabular}
\end{table*}

\subsection{Lessons Learned}
FL-CPS applications demonstrate significant potential across diverse domains, but also highlight key challenges and limitations, summarized in Table \ref{Table:FL_HCPS_Applications}, \ref{Table:FL_HCPS_Applications_Cont}, \ref{Table:FL_Smart_Cities_Applications}, \ref{Table:FL_VCPS_Applications}, \ref{Table:FL_Cybersecurity_Applications}. In healthcare, FL enables remote monitoring with high accuracy, improved gesture detection, and reduced latency for activity recognition.  However, challenges include ensuring convergence with non-IID data, high communication costs in hybrid architectures, vulnerability to attacks in decentralized settings, and the need for stronger privacy proofs.  FL also facilitates privacy-preserving EHRs management with techniques like perturbation, blockchain integration, and specialized layers, but limitations include accuracy drops under DP constraints, higher energy costs in blockchain-based settings, and complex synchronization.  In medical imaging, FL shows promise in cross-client variance reduction, low-resource deployment, and COVID-19 diagnosis, but faces challenges with data bias, accuracy degradation under DP, and limited validation on heterogeneous devices.

Smart city applications leverage FL for waste classification, load forecasting in smart grids, urban data management, and smart parking solutions.  However, limitations include the need for real-time edge deployments, optimization for intermittent connectivity, user-specific model variance, and vulnerability to attacks.  In vehicular applications, FL enhances intelligent transportation systems, vehicular networking, resource management, and traffic planning, but faces challenges with security, high computational load, and communication costs.  Cybersecurity and privacy applications utilize FL for attack detection, authentication, and intrusion detection, but must address limitations related to TEEs, accuracy drops, and energy consumption. Overall, FL-CPS demonstrates promising contributions in various domains, but also reveals key limitations related to data heterogeneity, privacy, security, communication costs, and computational overhead.  In details, research challenges and future research will be delved into next section.

\section{Research Challenges and Future Directions}
As indicated in earlier sections of this study, FL is receiving more and more attention in the application and implementation of CPS. FL provides the decentralized training of ML models across a network of devices, allowing CPS to use local data when executing challenging tasks. Despite FL's significant promise, the survey found a number of substantial research hurdles that must be overcome before FL can be deployed effectively in CPS. Security problems, resource management, standard specifications, heterogeneity, and fairness are hurdles for CPS networks. In addition, this section explains potential research avenues to address these issues.

\subsection{Security Issues in FL-based CPS}
While FL has the potential to increase security in CPS by allowing encrypted data to be transmitted between actuators from multiple distributed devices, thereby decreasing the likelihood of data breaches and attacks on centralized systems, several security concerns must be addressed when introducing FL into a CPS environment. FL, for example, has been contaminated in several ways \cite{fang2020local}. These attacks are classified according to their objective and the attacker's skill. The target determines whether an assault is targeted or not. Untargeted poisoning assaults decrease the global model's accuracy for all test inputs \cite{baruch2019little} \cite{guerraoui2018hidden}. Targeted poisoning attacks reduce test input precision while maintaining high precision \cite{bhagoji2019analyzing}. Backdoor assaults \cite{bagdasaryan2020backdoor} discreetly activate the attack's inputs. Untargeted attacks pose a greater threat to FL since they can destroy the global model. Indeed, the skill of the attacker is a critical factor in model and data poisoning attacks. Model poisoning attacks \cite{xie2020fall} modify gradients on infected devices before to their transmission to the server. Data poisoning \cite{jagielski2018manipulating} is the manipulation of device training data in order to modify gradients. Model poisoning attacks against FL have an instantaneous impact on gradients. Poisoning training data cannot duplicate the damaging gradients of model poisoning attacks because data poisoning attacks cannot generate arbitrary gradients. To assess the likelihood that FL will be poisoned, we concentrate on the more severe untargeted model poisoning attempts. Besides, given that data is saved locally on each client in a FL configuration, it makes sense for an attacker to employ sybils to maximize the efficacy of an attack \cite{fung2020limitations}. As a result, the global update may expose more personal data from local training data, compromising user privacy. In information processing domains like FL, CPS networks must store sensitive user data like preferences and home addresses. FL's privacy and security risks to intelligent CPS prevent data training in collaboration without a protection.

Generally, the integration of cyber and physical components in CPS presents new security concerns, and the security of FL systems, which are employed in these systems, is of the utmost significance. Several approaches are given as potential solutions to these problems, but further research and development are required to assure the security and dependability of FL systems in CPS. Specifically, the authors of \cite{wu2020mitigating} suggest a technique known as federated neuron pruning to counter the risk of FL backdoor assaults in CPS. This strategy decreases the model topologies of FL systems while preserving performance on preset tasks in order to make it more challenging for an attacker to encode a harmful pattern in the model. In addition, they examine the consequences of various attack settings, including the kind of backdoor design, the distribution of data across clients, and the aim of the assault. For this reason, the FL server might not be capable of rapidly update client adjustments because of the absence of training data and a validation dataset. In addition, model modifications are stochastic and unpredictable processes, which can result in considerable differences in simpler models. FoolsGold is another method presented in \cite{fung2018mitigating} for guarding against FL assaults in CPS. This technique aims to identify and fight against Sybil attacks in FL by regulating client learning rates depending on the similarity of their input to the global model. The researchers in \cite{zhao2022detecting} developed another way to detect and prevent label-flipping poisoning attacks on FL systems embedded in CPS. They employ GANs to audit participant models on the FL server as a countermeasure. The training set removes participant model parameters when their accuracy falls below a threshold. Eliminating malicious players' model parameters reduces label-flipping poisoning assaults.

\subsection{Resource Management Issues in FL-based CPS}
 Although it is previously recognized that FL-based solutions are feasible for various uses in CPS, the inherent difficulties of FL in ensuring timely model training with various computing-capable devices are evident in research fields. FL's objective in the context of CPS is to enable the maximum number of CPS clients to submit their local models prior to the global iteration deadline \cite{nishio2019client}. These clients, equipped with sensors, actuators, and embedded systems, can collect physical measurements and take actions based on the analysis of the data. However, selecting more clients can lead to an increase in energy consumption \cite{wang2020federated}. An alternative approach is to use fewer clients in the initial rounds and more in later rounds, which can can improve both training loss and model accuracy \cite{xu2020client}. The FL server must balance maximizing the number of selected clients while minimizing energy consumption, which favors fewer clients. In a cyber-physical system, finding the perfect balance between two objectives is tricky. The total energy consumption per global iteration across all connected is a function of the number of selected devices and the per-device energy consumption. This makes it hard to achieve the desired balance. Resource management plays a crucial role in the energy consumption of a device, for example, using fewer computational resources and reducing transmission power during data uploads decreases energy consumption \cite{wang2021federated}, while the opposite actions will lead to an increase in energy consumption.

Client selection has been the topic of various scholarly inquiries as a matter of significant concern. The objective of \cite{nishio2019client} is to create a system to assist the timely delivery of local models by means of a representative sample. Due to the fact that this approach does not account for client waiting time, executing a global iteration may result in an excessive latency. In order to reduce training latency, the creators of \cite{shi2020device} worked to optimize client selection and bandwidth distribution. The total number of global iterations was determined based on the sample size of the selected clients. The technique proposed in the publication \cite{yang2020federated} aims to improve the number of clients whose mean-squared-error criterion may be satisfied during each global iteration by combining beamforming with client selection. The study \cite{xu2020client} proposed a dynamic sampling of customers with a changeable maximum client count. This technique selects fewer consumers during early global iterations, when the number of customers has minimal impact on learning performance, and more consumers during later global iterations, when the number of customers can have a significant impact on learning efficiency. Dynamic client sampling minimizes energy consumption while enhancing training loss and model precision. The method in \cite{yao2020enhancing} shows how to optimize computing and wireless transmission power resources to tradeoff energy consumption and learning rate. Additionally, \cite{tran2019federated} also considers local model accuracy of the selected clients.

\subsection{Standard Specifications for Federated CPS Deployment}
Recent academic study has focused on the intricate interplay between the digital and physical worlds. Cyber-Physical Systems were developed for the purpose of defining these interdependencies.  By integrating actuators and sensors with communication, computers, and the physical substrate, CPS enables the development of open and adaptable system-of-system architectures. Despite the fact that multiple studies have demonstrated the potential benefits of FL in CPS services, there is no standard approach for comparing the efficacy of various treatments for the same disease. In order to reduce reliance on a centralized server and ensure the integrity of local device updating, several blockchain-based frameworks have been proposed for FL systems \cite{ngo2022blockchained}. With that said, they were all developed for distinct CPS use cases and tested with distinct network setups and data, making direct comparison challenging. Moreover, the absence of a uniform set of communication protocols, device hardware, deployment situations, and aggregation methodologies presents considerable obstacles. 

The latest IEEE Std 3652.1-2020 \cite{qiang2021white} standard defines FL architectural and design requirements. This document also addresses crucial issues for FL systems, including privacy, security, performance efficiency, economic viability, assessment schemes, and performance measurements. Besides, various elements of federated CPS have been the subject of numerous papers, including data aggregation and disaggregation \cite{jain2015towards}, compliance \cite{grangel2016towards}, process integration and standardization, and quality standards. When building an Industrial Automation System, it is common practice to adhere to the internationally recognized standard IEC 61131 \cite{grangel2016towards}. IEC 61131 was revised to version 2.0 to address problems expressed by Complex Industrial Automation systems, and IEC 61499 was produced as a successor standard. Before this new technique can be widely implemented, however, it must be standardized \cite{dafflon2021challenges}.

\subsection{Systems and Data Heterogeneity}
In the context of FL-CPS, system heterogeneity refers to client computing and communication capabilities, e.g., hardware, software, and network circumstances. Limiting the scalability of the FL approach, clients with restricted processing and communication capabilities may delay global model convergence. Clients with sluggish computations, known as stragglers, can slow down the FL algorithm. However, data heterogeneity refers to the distribution and properties of consumer data. This may result in an imbalance of data and a lack of control, making it difficult to train models applicable to the entire population. Data heterogeneity increases FL models' susceptibility to backdoor attacks, which leverage discrepancies in data distribution to add harmful behavior.

 Researchers have proposed a number of methods to enhance FL's functionality in the face of data and system heterogeneity. Numerous studies e.g., \cite{pillutla2019robust}\cite{ang2020robust}\cite{grama2020robust}, for example, proposed approaches for robust aggregation in the learning process of FL. The suggested methods employs a geometric median-based robust aggregation oracle that builds a robust aggregate by repeatedly using a non-robust averaging oracle. This robust aggregation oracle protects privacy similarly to the non-robust secure average oracle on which it is based. Additionally, the research \cite{zhang2020federated} suggested ACFL as a solution for FL communication problems resulting from dynamic bandwidth variety and unstable networks. Based on network conditions, the Cecilia cloud-edge-clients FL design of ACFL may change shared data compression. Similar to \cite{wang2019adaptive}, which focuses on a large class of FL models trained by gradient-descent techniques. This technique has handled the problem caused by crowdsourcing's generation of massive amounts of data at CPS.

\subsection{Communication Costs in FL-based CPS}
In FL, each client trains the model on its local data and communicates model changes (e.g., gradients) to a central server, which aggregates the updates and computes a new version of the global model, which is then returned to the clients. The global model is then updated using the aggregated client model updates, and the procedure is repeated until convergence. The clients and central server communicate over a broadcast channel with feedback. This channel carries both client-side model updates to the server and the server-side global model version to the clients. The channel's feedback function allows clients to report transmission issues or delays to the central server. The quality of the broadcast channel with feedback can have a significant effect on the efficiency of the FL algorithm. If the channel has limited capacity or excessive latency, it might result in model update delays and inaccuracies, slowing the convergence of the algorithm and leading to poor model performance.

With feedback, a variety of tactics can be used to increase the utilization of the broadcast channel. They include compressing model updates, prioritizing clients depending on their contribution to the model, and structuring communication rounds adaptively based on the channel's input.  The study \cite{murin2014ozarow} proposes an alternative, more promising method for the two-user Gaussian broadcast channel with noiseless feedback: linear-feedback schemes, which employ estimation techniques with memory in place of the memory-less estimation techniques utilized in the original work \cite{ozarow1984achievable}. This work provides a recursive description of the mean square errors attained by the suggested forecasting models and a proof of the existence of a fixed point to characterize the conceivable rates of the extended scheme. In addition, the author of \cite{ahmad2015concatenated} introduced a novel linear feedback framework for the symmetric additive white Gaussian noise broadcast channel with noisy feedback that was employed as an interior code in the concatenated coding scheme that was optimized to accomplish sum-rates greater than the sum-capacity without feedback. The FL algorithm provides faster convergence and improved model precision by employing the aforementioned feedback techniques to improve the efficacy and reliability of the broadcast channel.

\section{Conclusion}
In this comprehensive survey, we explored the integration of FL and CPS, offering a detailed analysis of recent advancements, key concepts, and the vast potential of FL-CPS applications across diverse domains such as healthcare, smart cities, and intelligent transportation. We delved into the intricacies of FL algorithms, CPS architectures, and communication protocols, highlighting their collaborative potential and addressing challenges related to privacy, security, and data heterogeneity. Our work also emphasized the importance of a generalized FL-CPS integration framework, providing a structured approach for researchers and practitioners. By shedding light on significant research gaps and outlining future directions, we aimed to catalyze further innovation and exploration in this rapidly evolving field, ultimately contributing to the development of more intelligent, efficient, and secure CPS solutions.

\balance
\bibliography{abbreviation, references}
\bibliographystyle{IEEEtran}

\end{document}